\documentclass[10pt]{article}

\usepackage{amssymb,amsmath,amsthm,mathrsfs} 
\usepackage{mathtools}
\usepackage{graphicx}
\usepackage{lineno}
\usepackage[font=small,labelfont=bf]{caption} 
\usepackage[authoryear,sort&compress]{natbib}
\usepackage{url}
\usepackage{enumerate}
\usepackage{colordvi,color,pspicture}
\usepackage{psfrag}
\usepackage{rotating}
\usepackage{epstopdf}
\usepackage{algorithm}       
\usepackage{algorithmic}
\usepackage{multirow}        
\usepackage{tablefootnote}
\usepackage{booktabs}
\usepackage{array}

\newcommand{\Z}{\mathcal{Z}}
\newcommand{\Y}{\mathcal{Y}}
\newcommand{\X}{\mathcal{X}}

\newcommand{\M}{\mathcal{M}}
\newcommand{\g}{\gamma}

\newcommand{\V}{\mathcal{V}}
\newcommand{\K}{\mathcal{K}}
\newcommand{\A}{\mathcal{A}}
\newcommand{\W}{\mathcal{W}}
\newcommand{\R}{\mathbb{R}}

\newcommand{\E}{\mathcal{E}}

\DeclareMathOperator{\argmax}{argmax}

\DeclareMathOperator{\vol}{Vol}

\theoremstyle{plain}
\newtheorem{theorem}{Theorem}

\begin{document}

\title{Technical Report \#4: \\ Parameter Free Clustering with Cluster Catch Digraphs}
\author{Art\"{u}r Manukyan\thanks{Brigham and Women's Hospital, Harvard Medical School \& Broad Institute of MIT and Harvard, Boston, MA, USA}  \,  \&
             Elvan Ceyhan\thanks{Auburn University, College of Sciences and Mathematics, Department of Mathematics and Statistics, Auburn, AL 36849 USA}}
\date{\today}
\maketitle

\pagenumbering{arabic} \setcounter{page}{1}

\begin{abstract}
\noindent
We propose clustering algorithms based on a recently developed geometric digraph family called cluster catch digraphs (CCDs). These digraphs are used to devise clustering methods that are hybrids of density-based and graph-based clustering methods. CCDs are appealing digraphs for clustering, since they estimate the number of clusters; however, CCDs (and density-based methods in general) require some information on a parameter representing the \emph{intensity} of assumed clusters in the data set. We propose algorithms that are parameter free versions of the CCD algorithm and does not require a specification of the intensity parameter whose choice is often critical in finding an optimal partitioning of the data set. We estimate the number of convex clusters by borrowing a tool from spatial data analysis, namely Ripley's $K$ function. We call our new digraphs utilizing the $K$ function as RK-CCDs. We show that the minimum dominating sets of RK-CCDs estimate and distinguish the clusters from noise clusters in a data set, and hence allow the estimation of the correct number of clusters. Our robust clustering algorithms are comprised of methods that estimate both the number of clusters and the intensity parameter, making them completely parameter free. We conduct Monte Carlo simulations and use real life data sets to compare RK-CCDs with some commonly used density-based and prototype-based clustering methods.

\end{abstract}

\noindent
{\small {\it Keywords:}  Class cover problem, Digraph, Domination, Prototype-based clustering, Support estimation, Arbitrarily shaped clusters, Noisy clustering

\vspace{.25 in}




\section{Introduction}
\label{sec:intro}

Clustering is one of the most challenging tasks in machine learning and pattern recognition, and perhaps,
discovering the exact number of clusters of an unlabelled data set is the leading one. Many clustering methods find the clusters (or hidden classes) and the number of these clusters simultaneously \citep{frey2007,sajana2016}.
Although there exist methods for validating and comparing the quality of a partitioning of a data set, algorithms that provide the (estimated) number of clusters without any input parameter are still appealing. However, such methods or algorithms rely on other parameters viewed as the intensity, i.e. expected number of objects in a unit area. The value of the intensity parameter works as a threshold, and if the local intensity of the data set exceeds the threshold, it may indicate the existence of a possible cluster. However, the choice of such parameters is often a difficult task since different values of such parameters may drastically change the result of the algorithm.
We use unsupervised adaptations of a family of vertex random digraphs, namely class cover catch digraphs (CCCDs), that showed relatively good performance in statistical pattern classification \citep{priebe:2003b,manukyan2016}. Unsupervised versions of CCCDs are called \emph{cluster catch digraphs} (CCDs) \citep{devinney2003,marchette2004}. Primarily, CCDs use statistics that require an intensity parameter to be specified or estimated. We approach the clustering of a data set by incorporating spatial data analysis tools that estimate the spatial intensity of regions in the domain.
We propose algorithms combining Ripley's $K$ function with CCDs to establish density-based clustering methods that find the optimal partitioning of unlabelled data sets, without the specification of a priori parameters.


We approach the problem of clustering data sets with methods that solve the class cover problem (CCP),
where the goal is to find a region that encapsulates all members of a class of interest. This particular region can be viewed as a \emph{cover}; hence the name \emph{class cover} \citep{cannon2004}. This problem is closely related to another problem in statistics, \emph{support estimation}: estimating the support of a particular random variable defined in a measurable space \citep{scholkopf2001}. Here, each cover can be viewed as an estimate of its associated class support. We explore the data set and reveal hidden classes by estimating their supports. We use the estimates of the supports to model the data set as a digraph, and then, we find the (approximate) \emph{minimum dominating set(s)} of these digraphs which presumably is the set of cluster centers we are looking for. CCDs are closely related to the class cover catch digraphs (CCCDs) of \citet{priebe2001} who introduced CCCDs to find graph theoretic solutions to the CCP, and provided some results on the minimum dominating sets and the distribution of the domination number of such digraphs for one dimensional data sets. \citet{priebe:2003b} applied CCCDs to classification and showed that approximate minimum dominating set(s) of CCCDs (which were obtained by a greedy algorithm) can be used to establish efficient prototype-based classifiers. \citet{devinney2002} defined random walk CCCDs (RW-CCCDs) where balls of class covers establish classifiers less prone to overfitting. CCCDs have been applied in face detection and latent class discovery in gene expression data sets \citep{socolinsky2002,priebe2003dna}. CCCDs are also shown to be robust to the class imbalance problem \citep{manukyan2016}. There are several other approaches in the literature for solving the class cover problem, including covering the classes with a set of boxes \citep{bereg2012} or a set of convex hulls \citep{takigawa2009}.

Clustering algorithms based on CCDs are very similar to density-based clustering methods which can find the exact number of arbitrarily shaped clusters in data sets \citep{sajana2016}. Jarvis-Patrick method and DBSCAN are some of the first density-based clustering methods \citep{jarvis1973,ester1996}. Both of these algorithms require prior domain knowledge on parameter values that determine the density of a subset of the data set in order to be considered as a member of a cluster, and this threshold can also be related to the \emph{intensity} of the data set. Both radius and MinPTS parameters of DBSCAN represent how dense a subset of the data is,
hence they are intensity parameters. CCDs estimate the number of clusters, but need the value of an intensity parameter to be specified as well. We propose modifications to the existing CCD method for dropping the necessity of knowing the intensity. We estimate the second-order moments of a set of points in a domain, and tell whether the points are clustered or not. \citet{ripley1977} introduced the $K$ function to estimate the second-order moments and used it to test clustering of spatial data sets. We combine Ripley's $K$ function with CCDs to find clusters with no prior assumption on the intensity. Tests based on  Ripley's $K$ function determine the size of the covering balls associated with CCDs, hence the name RK-CCDs. \cite{streib2011} use Ripley's $K$ function for a similar purpose: to calculate the scaling (bandwidth) parameter of Gaussian kernels used for similarity matrices. Therefore, we believe that the bandwidth parameter of kernel-based clustering algorithms fall into the category of intensity parameters, since \cite{streib2011} use tests based on Ripley's $K$ function in order to estimate the bandwidth. Thus, the bandwidth parameters of kernel-based algorithms like Support Vector Clustering (SVC) and pdfCluster can also be viewed as intensity parameters \citep{ben2001,azzalini2007,ester1996}. In this work, we use the $K$ function as an auxiliary tool specifically for our clustering algorithms, and we show that our adaptations of CCDs do not require any domain knowledge (i.e. the number of assumed clusters or the spatial intensity), hence are able to detect clusters without any prespecified value of a parameter.

We use the approximate minimum dominating sets (MDSs) of the CCDs to locate possible clusters. Our adaptations of CCDs successfully separate noise from the true clusters; however, we use cluster validation methods to detect true clusters. Specifically, the cluster validity methods help choosing a subset of the approximate MDS in such a way that the CCDs become efficient for data sets with unimodal convex clusters. Hence, the method works considerably well for data sets composed of either uniformly or normally distributed clusters. We also introduce other graph invariants of RK-CCDs for clustering data sets with arbitrarily shaped clusters. We find the maximal connected components of CCDs to locate arbitrarily shaped clusters in data sets. The support of the data set is estimated via a union of covering balls which allows us to explore an optimum partitioning of data sets with different types of clusters.  We use Monte Carlo simulations, as well as real data experiments,
to show that our modified CCDs improve over the existing clustering methods based on CCDs, and to demonstrate that the new CCDs perform comparable to other existing methods.

The article is organized as follows: in Section~\ref{sec:cluster}, we review the existing parameterized and parameter-free clustering algorithms for partitioning data sets (as well as the methods for validation) with convex and arbitrarily shaped clusters. In Section~\ref{sec:digraph}, we review cluster catch digraphs, and some auxiliary tools for these digraphs that are used to establish density-based clustering algorithms. In Section~\ref{sec:ccdclust}, we review the existing clustering methods based on CCDs, and also introduce some modifications to increase the clustering performance. In Section~\ref{sec:ccdripley}, we introduce a novel CCD based clustering algorithm, namely RK-CCDs. In Section~\ref{sec:montecarlo}, we assess the performance of CCD clustering methods and compare them with existing methods (such as fuzzy $c$-means and pdfCluster) on simulated data sets with three different simulation settings. Finally, in Section~\ref{sec:realdata}, we assess our clustering methods on real data sets with either convex or arbitrarily shaped clusters, and in Section~\ref{sec:conc}, we present discussion and conclusions as well as future research directions.

\section{Data Clustering} \label{sec:cluster}

Prototype-based, graph-based and density-based clustering algorithms are some of the most well known clustering algorithms in literature \citep{gan2007}. In particular, $k$-means algorithm, which falls into the prototype-based category, finds an optimal partitioning of the data set assuming the number of clusters are exactly $k$. Density-based algorithms, on the other hand, group the objects to similar clusters of a data set,
if either they are close to some high density regions of the domain or if they satisfy some similarity/dissimilarity relation. These methods mostly rely on some parameters that determine the minimum number of points such that a subset of the data set could be considered as a cluster. This parameter can be summarized as the assumed \emph{intensity} of a cluster. The clustering methods we discuss in this work are hybrids of graph-based and density-based algorithms. We determine the regions of high density in a data set by choosing the size and the location of the covering balls that are associated with the potential cluster centers. Once the size (or the radius) of each ball is determined, we establish geometric digraphs to find the clusters and the number of clusters simultaneously. We highlight some benchmark parameterized clustering algorithms, and investigate some of the recently developed parameter-free algorithms which are mostly adapted from their parameterized counterparts. Later, we briefly describe some validity indices and methods for assessing the performance of clustering algorithms.

\subsection{Algorithms for Convex and Arbitrarily Shaped Clusters}

Let $(\Omega,\M)$ be a measurable space, and let $\X=\{X_1,X_2,\ldots,X_n\} \subset \Omega$ be a set of $\Omega$-valued random variables with distribution $F$ and support $s(F)$. Here, we assume $X_i$ are drawn from a finite mixture of distributions $\{F_1,F_2,\ldots,F_M\}$. Thus, let $X_i \sim F:=\sum_{m=1}^{\K} \pi_m  F_m$ for $\K$ being the number of components of $F$ (which presumably correspond to the number of clusters) and $\pi_m \in (0,1)$ such that $\sum_{m} \pi_m=1$.
The goal of a partitional clustering algorithm potentially is to divide the data set $\X$ into $\widehat{\K}$, the estimate of $\K$, number of disjoint subsets $\mathcal{P}=\{P_1,P_2,\ldots,P_{\widehat{\K}}\}$ where $\X=\cup_{m=1}^{\widehat{\K}} P_m$, and hence to minimize (or maximize) some objective function $h(\mathcal{P})$ that gives the ``best" or ``optimal" partitioning. Here, $\widehat{\K}$ is either unknown or specified prior to the partitioning. In reality, there is no best partitioning (and thus no best function $h$) of a data set $\X$, since the objective function $h$ depends on the distribution $F$, and most importantly, depends on the problem at hand. We review some of the most frequently used partitional clustering algorithms for data sets with both convex and arbitrarily shaped clusters.
We first examine those algorithms that require a priori parameters, and then we move on to algorithms that are parameter-free.

\subsubsection{Parameterized Algorithms}

The $k$-means algorithm is perhaps the most known and commonly used clustering algorithm, even though it is simple and prone to errors for some common types of data sets \citep{ball1965}. Let the (estimated and presumed) number of clusters $k=\widehat{\K}$ be given. The objective function ---to be minimized--- $h_k$ of the $k$-means algorithm is defined as
\begin{eqnarray*}
h_k(\mathcal{P})=\sum_{i=1}^n \sum_{m=1}^{\widehat{\K}} d^2(X_i,\mu(P_m)).
\end{eqnarray*}
Here, $d(\cdot,\cdot)$ is a dissimilarity measure (e.g. Euclidean distance), and $\mu(P_m) \in \R^d$ is the center of the cluster $P_m$ which is randomly chosen prior to the execution of the algorithm. At each step, $\mu(P_m)$ is updated as the current center of mass of all points in $P_m$, and then the members of $P_m$ are reconfigured given the collection of all points closest to $\mu(P_m)$. This is repeated until no considerable change is observed in $\mu(P_m)$ for all $m=1,\ldots,\widehat{\K}$. There are many extensions of $k$-means algorithm each mitigating the effects of some difficulties observed in real life data sets. One such algorithm is the ISODATA where the number of clusters is initialized before the execution of the algorithm; however, the clusters are merged or split during the execution based on some criteria (such as minimum number of points in a cluster or the minimum distance between two clusters) \citep{ball1965}. But ISODATA is sensitive to weakly separated compact clusters (i.e. clusters that are close to each other), and may merge or split existing clusters, hence diverging from the optimal solution.  \cite{dunn1973} introduced two adaptations of $k$-means, known as fuzzy $c$-means, where the first is a fuzzy version of ISODATA, and the latter is a completely novel fuzzy partitioning algorithm with the objective function
\begin{eqnarray*}
h_c(\mathcal{P})=\sum_{i=1}^n \sum_{m=1}^{\widehat{\K}} u^q_{ij} d^2(X_i,\mu(P_m)).
\end{eqnarray*}
The objective function $h_c$ is slightly different than the function $h_k$ where $u_{im}$ denotes the magnitude of membership of $X_i$ in cluster $P_m$, and $q>1$ is some parameter to adjust fuzziness. Clustering methods with fuzzy membership functions are more robust to weakly separated clusters; that is, they tend to converge to the optimal clustering solution even when clusters are substantially close to each other. We primarily investigate the fuzzy $c$-means algorithm and compare it with our CCD methods.

Clustering methods such as $k$-means perform relatively well if the clusters in a data set are convex and compact. However, all such algorithms require to be initialized with a prior knowledge on the number of clusters which may not be estimated or guessed easily before the execution of the algorithm. In addition, these algorithms satisfy a partitioning criteria which may not be suitable for data sets with arbitrarily shaped clusters. These special types of data sets do not necessarily have centers. Density-based methods are based on capturing the density around points, and hence they estimate the true number of clusters. The Jarvis-Patrick algorithm is one of the first methods of its kind for capturing arbitrarily shaped and non-arbitrarily shaped clusters \citep{jarvis1973}. The DBSCAN algorithm of \cite{ester1996} and the OPTICS algorithm of \cite{ankerst1999} work in a similar fashion. Given two parameters, the radius $\epsilon$ and the minimum number of points \emph{MinPts}  in the neighborhood of a point, the data set is divided into several types of points: \emph{core}, \emph{border} and \emph{noise} points. The union of these neighborhoods (i.e. the points within the radius $\epsilon$) of core points constitutes the clusters. The parameters $\epsilon$ and \emph{MinPts} may change the result of the algorithm drastically and both of these parameters represent the \emph{intensity} of the spatial distribution of clusters in a data set.

Kernel-based clustering algorithms are often employed for estimating the true number of convex clusters or locating arbitrarily shaped clusters in a data set. The intensity of a spatial distribution is provided by the bandwidth parameter of the Gaussian kernel. One such algorithm is the \emph{pdfCluster} where clustering depends on the density estimation of the data sets via Delaunay triangulation (or Delaunay tessellation in $\R^d$ for $d>2$). Vertices of the triangulation, whose estimated density is below a certain cut value are removed, and the remaining connected components of the triangulation are set to be the clusters. A culmination of results from several choices of cut values is analysed to decide the final partitioning. The density around each point is estimated with a kernel function $\Phi$ (usually a Gaussian kernel with bandwidth $h$) as follows: for $y \in \R^d$,
\begin{eqnarray*}
f(y) = \sum_{i=1}^n \frac{1}{n h_{i1} \cdots h_{id}} \sum_{j=1}^d \Phi\left(\frac{y-X_{ij}}{h_{ij}}\right).
\end{eqnarray*}
where $X_{ij}$ and $h_{ij}$ are the value and the kernel width, respectively, of $j$'th coordinate of the random variable associated with the $i$'th observation.

\subsubsection{Parameter-Free Algorithms} \label{sec:param-free}

Many algorithms have been proposed to find clusters in a data set without either introducing the assumed number of true clusters or the intensity parameter. Among these methods, Affinity Propagation (AP) algorithm is perhaps the most popular one which attracted considerable attention in the bioinformatics community \citep{frey2007,bodenhofer2011}. AP algorithm considers each observation as a potential cluster center, referred to as \emph{exemplars}. The measures of \emph{responsibility} and \emph{availability} of each pair of observations are iteratively updated until a considerable convergence is achieved. Given the similarity measure $s(i,j)$ between $i$'th and $j$'th points, the responsibility $r(i,k)$ reflects the evidence for how well the $k$'th point serves the role of being an exemplar for $i$'th point:
\begin{equation}
r(i,k) := s(i,k) - \underset{k'\neq k}{\max}\{a(i,k') + s(i,k')\}.
\end{equation}
Here, the availability $a(i,j)$ measures how appropriate for $i$'th point to choose $k$'th point as its exemplar:
\begin{equation}
a(i,k) := \min\{0, r(k,k) + \underset{i' \not\in\{i,k\}}{\sum} \max\{0,r(i',k)\}\}.
\end{equation}
Thus, the availability and responsibility measures are updated, back and forth, in a message-passing fashion to find the best exemplars out of all data points. Throughout the execution of the algorithm, each observation competes for the ownership of other points. Those points whose sums of availability and responsibility measures are nonnegative are chosen as exemplars, or cluster centers. However, initial responsibility measures of the observations, referred to as \emph{self-responsibility}, is set to an initial value which may be based on the similarity measures between pairs of points. The higher the self-responsibility, the higher the number of clusters. \cite{bodenhofer2011} suggested a parameter which sets the self-responsibility as the $q$'th quantile of the set of similarities. In fact, AP algorithm is not entirely parameter-free but it achieves high performance when the minimum similarity value (i.e. the minimum input preference where $q=0$) is chosen as the self-responsibility. Therefore, the resulting partitioning of the data set heavily depends on the initial self-responsibility (or $q$) parameter.

DBCLASSD algorithm is a density-based clustering algorithm that does not require an intensity parameter or assumed number of clusters to detect arbitrary or non-arbitrary shaped clusters \citep{xu1998}. The algorithm, however, only assumes that the class conditional distributions of clusters are bivariate uniform. DBCLASSD follows an incremental approach, adding the next point to a cluster, if the new point does not significantly change the nearest neighbor distance distribution of the cluster which depends on an approximately calculated volume of the data set in the domain.
Here, the set of nearest neighbor distances of a set of points is assumed to have $\chi^2$ distribution.
In fact, DBCLASSD is viewed as an adaptive version of DBSCAN; that is, both assumptions on the values of parameters $\epsilon$ and \emph{MinPts} are dropped.

There are several other methods which mainly are variations of AP or DBSCAN algorithm, or both. APSCAN is a version of the AP algorithm adapted for data sets with arbitrarily shaped clusters \citep{chen2011}. Similarly, Dsets-DBSCAN combines the Dominant Sets clustering and DBSCAN algorithms \citep{hou2016}. A subset of the vertex set is said to be a \emph{dominant set}, if it satisfies \emph{internal homogeneity} and \emph{external inhomogeneity} conditions. These sets are found by solving Standard Quadratic Program (StQP) which is based on a weighted adjacency (or similarity) matrix. \cite{hou2016} showed that initial values of the $\epsilon$ and \emph{MinPts} parameters do not drastically change the result of the algorithm. PFClust is another parameter-free algorithm that partitions the data into an initial random number of clusters, and iteratively refines (or merges) the clusters until the partition of the data set converges to a good solution. PFClust successfully clusters three dimensional structures of protein data sets.

The CCDs are graph representations of the data sets, and we locate the (approximately) minimum dominating sets of such directed graphs. The existence of a directed edge (arc) indicates a similarity between a pair of nodes where the similarity is established by methods similar to density-based algorithms.
Hence, \emph{our methods are considered to be hybrids of graph-based and density-based clustering methods}.
One advantage of our adaptations of CCDs is that our algorithms are able to find the correct number of clusters
even if the clusters are convex or arbitrarily shaped. We also show that, for some convex clusters, such as normally distributed ones, our algorithms are able to find the true cluster centers by separating them from the noisy clusters.

\subsection{Validation of the Clustering Algorithms}

A partitioning $\mathcal{P}$ can be validated by a variety of indices.
These indices are often divided into three major types based on the criteria employed
which are referred to as \emph{internal}, \emph{external} and \emph{relative} criteria \citep{gan2007}.
Some existing indices can be modified indices of either internal or external criteria. These are Hubert's $\Gamma$ statistic and Goodman-Kruskal's $\gamma$ statistic. Moreover, Rand statistic, Jaccard statistic and Folkes-Mallows index are used as indices of external; and cophenetic correlation coefficient is used as an index of internal criteria. We focus on the relative criteria since our algorithms are based on this family of indices.


Indices of relative criteria are only meaningful when several clustering algorithms are to be compared.
However, we use these indices to locate an appropriate subset of our approximate minimum dominating sets where members of this subset correspond to the true clusters. On the other hand, the more noisy a data set becomes, the more we find spurious clusters. A collection of noise in the data set may be misinterpreted as a set of members from a separate cluster, or two or more clusters may appear as a single cluster. \cite{ben2014} state that a clustering algorithm may become robust to noise; that is, the \emph{noise clusters} (subsets of noise in the data set) could be separated from the true clusters by increasing a priori defined number of clusters in prototype-based clustering methods like $k$-means before execution. Hence, it is as if the algorithm includes a ``garbage collector", a term to refer to a collection of noise points in the data set. Our goal is to employ validity indices to separate noise clusters and the true clusters in a similar fashion.

For RK-CCD, we will show that relative indices are particularly appealing for finding a set of distant and compact clusters in which both the true and noise clusters exists. Many indices of relative criteria indicate or measure some relationship between \emph{intra-cluster} distances (the distances between the members of the same cluster) and \emph{inter-cluster} distances (the distances between the (members of) different clusters). A simple index based on this criteria is the \emph{silhouette} index \citep{gan2007}. Given a partitioning $\mathcal{P}=\{P_1,P_2,\ldots,P_{\widehat{\K}}\}$ of the data set $\X=\{X_1, X_2, \ldots, X_n\}$,
the silhouette index measures how well a data set is clustered
(or how well a clustering algorithm performed in labeling the points correctly).
Let the point $X_i$ be a member of cluster $P_m$.
Let $a(i)$ be the average distance from $X_i$ to all other members of $P_m$, and let $b(i)$ be the distance from the point $X_i$ to the closest cluster $P_j$ for $j \neq m$ where the distance between $X_i$ and $P_j$ is given by the average distance from $X_i$ to all members of $P_j$. Hence, we have $$a(i):=\frac{1}{|P_m|-1}\sum_{Y \in P_m} d(X_i,Y)$$ and $$b(i):=\min_{j \neq m} \frac{1}{|P_j|}\sum_{Y \in P_j} d(X_i,Y).$$
Thus, the silhouette $sil(i)$ of the random variable $X_i$ is denoted as
$$sil(i):=\frac{b(i)-a(i)}{\max\{a(i),b(i)\}}.$$
Here, $sil(i)$ indicates how well the point $X_i$ is clustered in the cluster $P_m$ in a given partitioning $\mathcal{P}$. An overall assessment of how well the entire data set is clustered can be defined as the average of silhouette of the data set, i.e. $sil(\mathcal{P})=\sum_{i} sil(i)/n$.

The maximum average silhouette is often employed to choose the best $k$ (or $\widehat{\K}$) in $k$-means clustering method. The algorithm is executed for a set of different number of clusters $\widehat{\K}$, and the partitioning which produces the maximum average silhouette is chosen to be the best clustering of the data set. Unlike methods such as $k$-means where the number of clusters should be specified before execution, we will incorporate this measure to choose a subset of the minimum dominating sets of CCDs that maximizes the average silhouette, thus revealing the true clusters.


\section{Digraphs and Auxiliary Tools} \label{sec:digraph}

We propose clustering methods that are hybrids of density-based and graph-based clustering methods. CCDs are digraphs constructed by some similarity relation between pairs of points in a data set. The size of each covering ball is determined by the rejection of a spatial pattern test, and only a subset of these covering balls is chosen as indicators of the true clusters where their centers are the members of the minimum dominating sets. However, unlike the recently introduced CCDs \citep{devinney2003,marchette2004}, our unsupervised learning methods select the members of MDSs, or cluster centers, in a different fashion. Specifically, we first find a minimum dominating set of the data set, then we select a subset of this set which is the final MDS (nearly) equivalent to the set of centers of the clusters we are looking for. The first set is based on catch digraphs whereas the second is based on intersection graphs. Moreover, we use tests based on Ripley's $K$ function to find the optimal covering balls for all the points in the data set whereas the existing CCD methods use Kolmogorov-Smirnov based statistics that rely on some spatial intensity parameter.


\subsection{Class Cover Catch Digraphs} \label{sec:cccds}

Class Cover Catch Digraphs (CCCDs) are graph theoretic representations of the class cover problem (CCP) \citep{cannon2004,priebe2001,priebe:2003b}. The goal in CCP is to find a region that encapsulates all the members of the class of interest where this particular region can be viewed as a \emph{cover}. Let $\X_0=\{x_1,x_2,...,x_{n_0}\} \subset \R^d$ and $\X_1=\{y_1,y_2,...,y_{n_1}\}\subset \R^d$ be sets of observations from two classes of data sets. We refer to the class that we aim to find the cover of as \emph{target class}, and the remaining class(es), or the class(es) that are not of interest, as \emph{non-target class(es)}. Without loss of generality, assume that the target class (i.e. the class of interest) is $\X_0$. In a CCCD, for $u,v \in \X_0$, let $B(u,r)$ be the hypersphere centered at $u$ with radius $r=r(u)$. A CCCD $D_0$ is a digraph $D_0=(\V_0,\A_0)$ with vertex set $\V_0=\X_0$ and the arc set $\A_0$ where $(u,v) \in \A_0$ iff $v \in B(u,r)$;
then it is said that $u$ catches (or covers) $v$. One particular family of CCCDs are called pure-CCCDs (P-CCCDs) wherein the \emph{covering} ball (or covering hypersphere) $B(u,r)$ is the ball with $u$ at its center, expanding until it hits the closest non-target class point. Therefore, $r=r(u):=\min_{v \in \X_1} d(u,v)$ \citep{marchette2010}. Here, $d(.,.)$ can be any dissimilarity measure, but we use the Euclidean distance henceforth. For all $u \in \X_0$, the radius $r$ is defined in such a way that no point of $\X_1$ is in the covering ball $B(u,r)$, i.e. $\X_1 \cap B(u,r) = \emptyset$
for any $u \in \X_0$. The digraph $D_1$ is constructed by interchanging the roles of target class being $\X_1$ and non-target class being $\X_0$.

For $j=0,1$, the cover (or estimated support) of class $\X_j$ is the union of all covering balls $Q_j:=\cup_{x \in \X_j} B(x,r(x))$.
However, we seek a lower complexity class cover which roughly gives the same estimation as $Q_j$.
We define the lower complexity cover as $$C_j:= \cup_{x \in S_j} B(x,r(x))$$ where $S_j \subseteq \X_j$ is a subset of the data set, often referred to as the \emph{prototype set}, equivalent to the minimum dominating sets of CCCDs. \cite{priebe:2003b} used the prototype sets to establish appealing class covers that have relatively good performance when used as semi-parametric classifiers. \cite{manukyan2016} showed that these prototype sets also balance the number of observations from each class so as to mitigate the effects of the class imbalance problem.

\subsection{Cluster Catch Digraphs}

Cluster Catch Digraphs (CCDs) are unsupervised adaptations of CCCDs \citep{marchette2004}. The definition of a CCD is similar to CCCDs; that is, a CCD is a digraph $D=(\V,\A) $ such that $\V=\X$ and, for $u,v \in \V$, we have $(u,v) \in \A$ iff $v \in B(u,r)$. In CCCDs, however, each point $X \in \X$ is the center of covering ball $B(X,r(X))$ where $r(X)$ is a function of both the target and non-target classes. In CCDs, we do not have (or know the existance of) a  non-target class to determine the radius. Instead, the objective is to locate, or \emph{catch}, existing points around the point $X$
which potentially constitute a cluster, hence the name \emph{cluster catch digraph}. The radius $r(X)$ is determined by maximizing a Kolmogorov-Smirnov (K-S) based statistic \citep{devinney2003}.
Thus, we have
\begin{equation} \label{k-s}
	 r(X) := \underset{r \in \{d(X,Z): Z \in \X\}}{\argmax} RW(X,r)-F_{0}(X,r).
\end{equation}
Here, $RW(X,r)$ is the random walk function proportional to the percentage of points lying inside the covering ball $B(X,r)$; i.e.
\begin{eqnarray}
RW(X,r)=\sum_{Z \in \X} I(d(X,Z) < r),
\end{eqnarray}
were, $I(\cdot)$ is the indicator functional,
and $F_{0}(X,r)$ is a function based on the null distribution
and is proportional to the expected number of points in $B(X,r)$ under the null distribution. A maximum value of
$RW(X,r)-F_{0}(X,r)$ on the radius $r$ indicates that the points inside $B(X,r)$ more likely constitute a cluster than a random collection of points. We take $F_{0}(X,r)=\delta r^d$ for $\X \subset \R^d$ but other choices of the form of this function are also possible \citep{marchette2004}.
Here, the null hypothesis assumes that the points randomly uniformly fall in the covering ball, then the density is proportional to the volume of the ball. However, the function $F_{0}(X,r)$ depends on the parameter $\delta$ which represents the intensity of the tested homogeneous Poisson process. After we determine $r(X)$ for all $X \in \X$, we construct the CCD, $D$, and find the approximate minimum dominating set $S_{MD}$ of $D$.
We provide an illustration of the digraph $D$, the set of covering balls and the approximate MDS, $S_{MD}$, in Figure~\ref{fig:digraph}.

\begin{figure}[h]
\centering
\begin{tabular}{cc}
\includegraphics[scale=0.4]{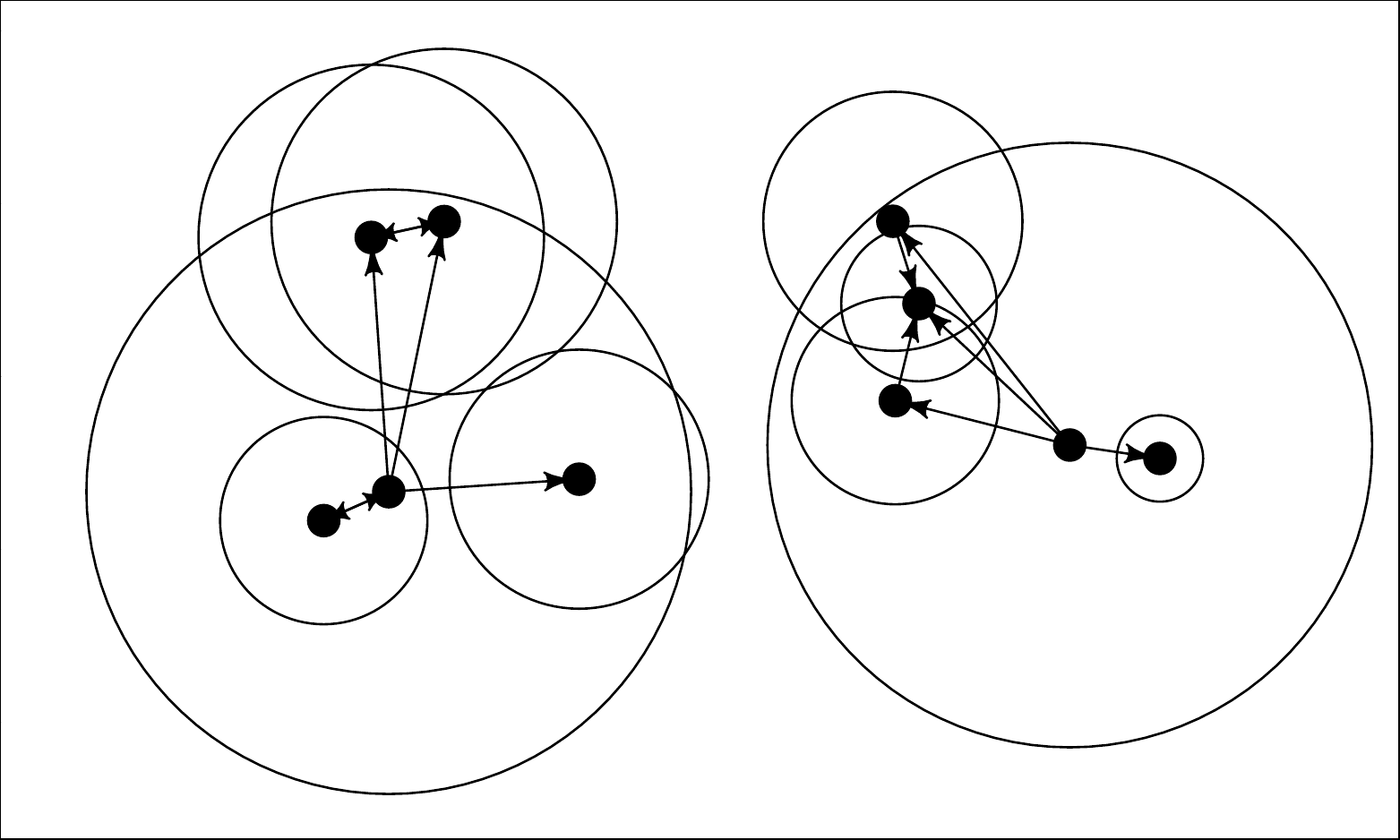} & \includegraphics[scale=0.4]{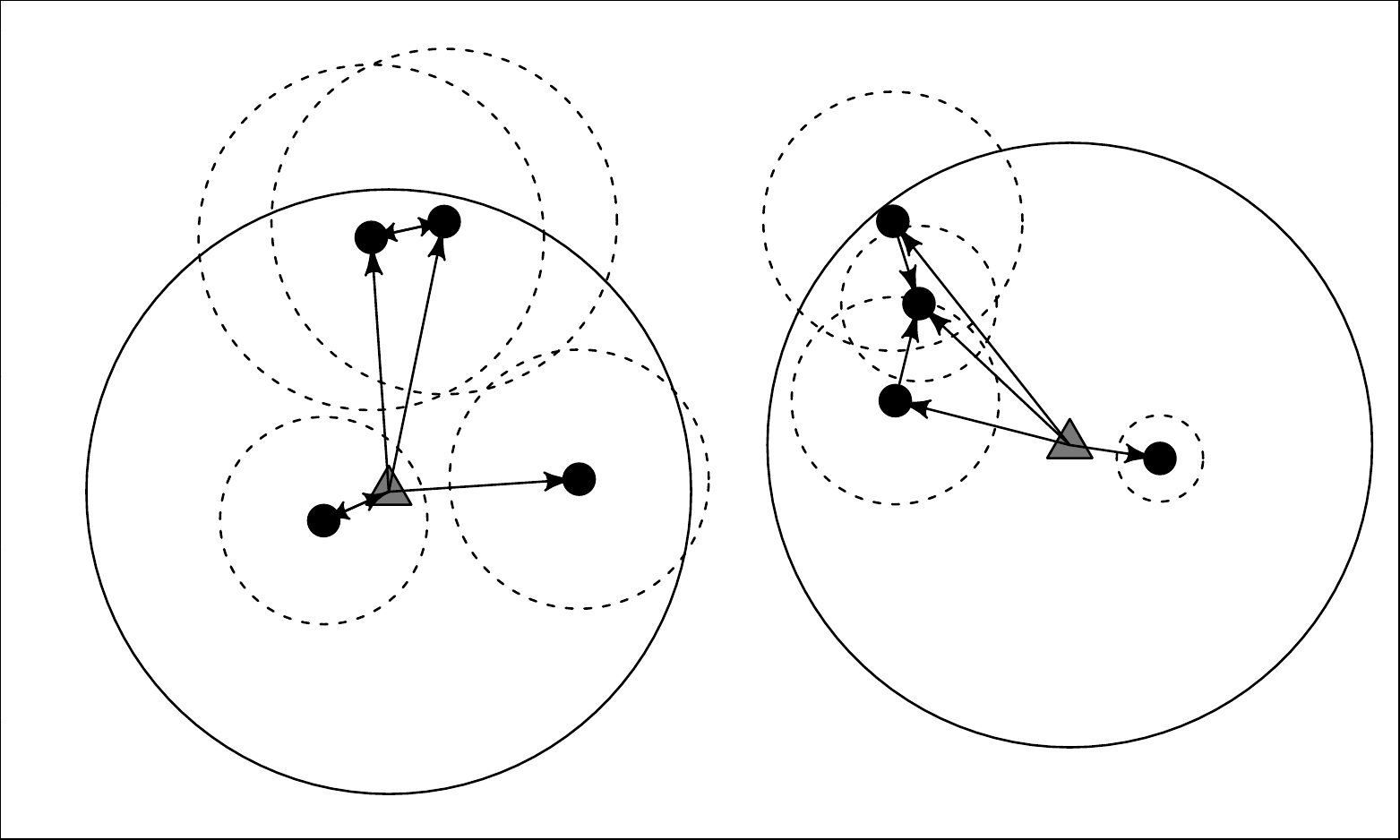} \\
(a) & (b) \\
\\
\multicolumn{2}{c}{\includegraphics[scale=0.4]{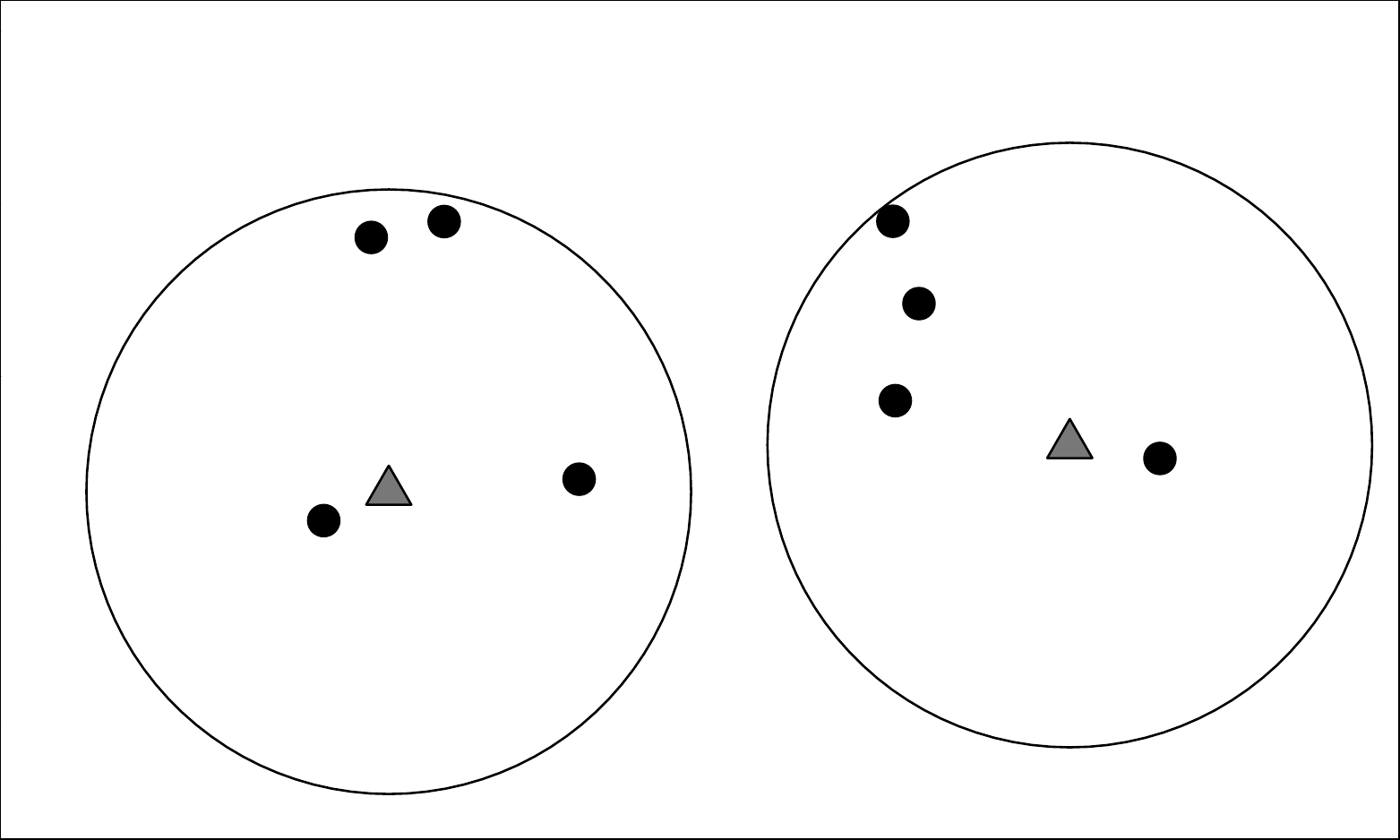}} \\
\multicolumn{2}{c}{(c)} \\
\end{tabular}
\caption{An illustration of the CCDs of a data set $\X \subset \R^2$ that constitute two clusters.
(a) All covering balls and the digraph $D=(\V,\A)$.
(b) The balls that constitute a cover for the data points and are centered at points which are the elements of the dominating set $S_{MD} \subseteq \V(D)$.
The covering balls of the elements of the set $S_{MD}$ are given with solid lines (and the elements as grey triangles) and the covering balls of the elements of $\V(D) \setminus S_{MD}$ are given in dashed lines.
(c) The covering balls associated with two clusters.}
\label{fig:digraph}
\end{figure}

\subsection{Domination in Graphs and Digraphs}

For a given set of covering balls $B(X,r)$ for $X \in \X$, our estimate for the support of the data set with several clusters is $Q:=\cup_{X \in \X} B(X,r)$.  Nevertheless, the support of the data set $\X$ can be estimated by a cover with lower complexity (fewer number of covering balls). For this purpose, we wish to reduce the complexity by selecting an appropriate subset of covering balls that still approximately gives a good estimate; that is, let the cover be defined as $C:=\cup_{X \in S} B(X,r)$, where $S$ is a prototype set of points $\X$ such that $\X \subset C$. A reasonable choice of prototype set is a {\em minimum dominating set} (MDS), whose elements are often more ``central" than the arbitrary sets of the same size for CCDs. We use various algorithms to find (approximate) minimum dominating sets of digraphs constructed by the covers for both
providing lower complexity estimates of the support, and for locating the cluster centers.

In general, a digraph $D=(\V,\A)$ of order $n=|\V|$, a vertex $v$ {\em dominates} itself and all vertices of the form $\{u:\,(v,u) \in \A\}$. A {\em dominating set}, $S_D$, for the digraph $D$ is a subset of $\V$ such that each vertex $v \in \V$ is either dominated by a vertex in $S_D$ or a member of $S_D$. A {\em minimum dominating set} (MDS), $S^{*}_{MD}$, is a dominating set of minimum cardinality, and the {\em domination number}, $\g(D)$, is defined as $\g(D):=|S^{*}_{MD}|$, where $|\cdot|$ stands for the cardinality of a set. If an MDS is of size one, we call it a {\em dominating point}. Finding an exact (or optimum) MDS is, in general, an NP-Hard optimization problem \citep{karr1992,arora1996}. However, an approximate MDS can be obtained in $O(n^2)$ time using a well-known greedy algorithm provided in Algorithm~\ref{alg:dom_greedy} \citep{chvatal:1979,parekh:1991}.

\begin{algorithm}
\begin{algorithmic}
 \REQUIRE A digraph $D=(\V,\A)$
 \ENSURE An approximate dominating set, $S_{MD}$
 \STATE  \textbf{set} $H=\V$ and $S_{MD} = \emptyset$
 \WHILE{$H \neq \emptyset$}
  \STATE $v^{*} \leftarrow \argmax_{v \in \V(D)} |\{u \in \V(D):(v,u) \in \A(D)\}|$
  \STATE $S_{MD} \leftarrow S_{MD} \cup \{v^{*}\}$
  \STATE $\bar{N}(v^{*}) \leftarrow \{u \in \V(D): (v^{*},u) \in \A(D)\} \cup \{v^{*}\}$
  \STATE $H \leftarrow \V(D) \setminus \bar{N}(v^{*})$
  \STATE $D \leftarrow D[H]$
 \ENDWHILE
\end{algorithmic}
\caption{The greedy algorithm for finding an approximate MDS, $S_{MD}$, of a digraph $D$.
Here, $\V(D)$ and $\A(D)$ are the vertex set and the arc set of the digraph $D$, respectively, and $D[H]$ is the digraph induced by the set of vertices $H \subseteq \V$ \citep[see][]{west2000}.}
\label{alg:dom_greedy}
\end{algorithm}

Algorithm~\ref{alg:dom_greedy} can be modified in a few ways that satisfies some other properties. We will see that it is more desirable for CCD algorithms to choose covering balls closer to the cluster centers.
These balls are bigger in size and cover more points than those located around edges of clusters. Hence, we change Algorithm~\ref{alg:dom_greedy} in such a way that the members of MDS have relatively high outdegrees
(hence balls tend to have bigger radii), i.e. $d_{out}(v):=|\{u:(v,u) \in \A(D)\}|$, compared to the MDSs found by Algorithm~\ref{alg:dom_greedy}. At each iteration, Algorithm~\ref{alg:dom_greedy_outdegree} chooses the vertex having maximum outdegree and having arcs to those vertices not yet covered. This approach more likely solves the approximate MDS problem with members of the dominating set being closer to the cluster centers.

\begin{algorithm}
\begin{algorithmic}
 \REQUIRE A digraph $D=(\V,\A)$
 \ENSURE An approximate dominating set, $S_{MD}$
 \STATE  \textbf{set} $S_{MD} = \emptyset$
 \WHILE{$\V(D) \neq \emptyset$}
  \STATE $v^{*} \leftarrow \argmax_{v \in \V} |\{u \in \V(D): (v,u) \in \A \}|$
  \STATE $S_{MD} \leftarrow S_{MD} \cup \{v^{*}\}$
  \STATE $\V \leftarrow \V \setminus v^*$
  \STATE $H \leftarrow \V(D) \setminus \bar{N}(v^{*})$ where $\bar{N}(v^{*})$ is as in Algorithm \ref{alg:dom_greedy}.
  \STATE $D \leftarrow D[H]$
 \ENDWHILE
\end{algorithmic}
\caption{The greedy algorithm for finding an approximate MDS, $S_{MD}$, of a digraph $D=(\V(D),\A(D))$ exploiting outdegrees
where $\V(D)$ and $\A(D)$ and $D[H]$ are as in Algorithm \ref{alg:dom_greedy}.
}
\label{alg:dom_greedy_outdegree}
\end{algorithm}

Algorithms~\ref{alg:dom_greedy} and~\ref{alg:dom_greedy_outdegree} select the vertices with the highest outdegree;
however, we may use other attributes of the vertices to select members of the (approximately) minimum dominating sets; that is, let $sc: \V \rightarrow \R$ be a
(score) function of the vertices of a digraph, and let each vertex $v$ is associated with a score $sc(v)$. Algorithm~\ref{alg:dom_greedy_score} adds a vertex $v \in \V$ to the set of dominating points $S_{MD}$, if it maximizes the score $sc(v)$.
The algorithm terminates when $H$, the set of remaining vertices, becomes empty.


\begin{algorithm}
\begin{algorithmic}
 \REQUIRE A digraph $D=(\V,\A)$ and a score set $\{sc(v):v \in \V\}$
 \ENSURE An approximate dominating set, $S_{MD}$
 \STATE  \textbf{set} $H=\V$ and $S_{MD} = \emptyset$
 \WHILE{$H \neq \emptyset$}
  \STATE $v^{*} \leftarrow \argmax_{v \in \V(D)} sc(v)$
  \STATE $S_{MD} \leftarrow S_{MD} \cup \{v^{*}\}$
  \STATE $H \leftarrow \V(D) \setminus \bar{N}(v^{*})$ where $\bar{N}(v^{*})$ is as in Algorithm \ref{alg:dom_greedy}.
  \STATE $D \leftarrow D[H]$
 \ENDWHILE
\end{algorithmic}
\caption{The greedy algorithm exploiting the scoring function $sc:\V \rightarrow \R$ for finding an approximate MDS of a digraph $D =(\V(D),\A(D))$
where $\V(D)$ and $\A(D)$ and $D[H]$ are as in Algorithm \ref{alg:dom_greedy}.
}
\label{alg:dom_greedy_score}
\end{algorithm}

The simplest of scoring functions for digraphs is the outdegree $d_{out}(v)$ of a vertex $v \in \V(D)$.
In Algorithm~\ref{alg:dom_greedy}, the outdegrees of vertices change each time a dominating point is added to $S_{MD}$. However, when $sc(v):=d_{out}(v)$ is fixed, the Algorithm~\ref{alg:dom_greedy_score} takes the original outdegrees into account. This algorithm will successfully locate points of high domain influence which potentially correspond to the centers of some (hidden) clusters.
All of these algorithms can easily be modified for use on undirected graphs (where arcs are replaced by edges).

\subsection{Spatial Data Analysis and Ripley's $K$ function}

We use spatial data analysis tools to test the existence of clusters in the domain. One such test, based on the Ripley's $K$ function, exploits distances between points to determine the second-order moments of point processes in a window $\W \subset \Omega$. Human eye is proficient in detecting the second-order properties of spatial objects in 2 dimensional (or in 3 dimensional) domains \citep{julesz1975}. Moreover, \cite{ripley1976} states that the first and second-order moments of stationary point processes are sufficiently expressed in terms of the intensity $\lambda$ and a function $K$, respectively, which can be used to devise tests for analyzing spatial data sets. We offer algorithms that employ both CCDs and Ripley's $K$ function to estimate the supports of individual hidden classes in a data set. Here, we review the $K$ function and the associated tests for spatial point patterns.

Methods based on quadrats and nearest neighbor distances are some of the most popular methods \citep{ripley2005}
in analyzing mapped point patterns in literature.
In addition, \cite{ripley1977} showed that the second-order moments of homogeneous spatial processes could be reduced to a function $K(t)$ on distance $t \in \R_+$ where $K(t)$ of a data set $\X$ is defined as the expected number of pairs in $\X$ that are at most $t$ distance apart. Hence, the distances between points are exploited to test whether a set of random events fit a certain spatial distribution. Now, let $\lambda$ be the expected number of events in a region $A \subset \Omega$ of unit area (or volume). Hence, $\lambda^2 K(t)$ is the expected number of ordered pairs at most $t$ distance apart.

Let $\Z=\{Z_1,Z_2, \ldots, Z_n\}$ be a set of $\R^d$-valued random events in a domain, or \emph{window}, $\W \subset \R^d$. We use the estimate $\widehat{K}(t)$ of the expected number of pairs that are at most $t$ distance apart in $\W$ \citep{ripley1976}:
\begin{equation} \label{K_biased}
\widehat{K}(t):=\frac{\vol(A)}{n(n-1)} \sum_{z \in \Z}\sum_{z' \in \Z \setminus \{z\}} I(d(z,z') < t)\vartheta(z,z').
\end{equation}
Here, $\vartheta(z,z')$ is the weighting function to correct the bias due to the edge or boundary effects
of the study region $\W$.
Specifically, the neighbourhood of $z$ with radius $t$ is a ball $B(z,t)$ which may be sufficiently close to the edge of the window $\W$ such that $B(z,t) \setminus \W \neq \emptyset$. Since $\Z \subset \W$ and some subset of $B(z,t)$ would be outside of $\W$,
and these subsets of $B(z,t)$ would be contain no data points;
thus $\widehat{K}(t)$ in Equation (\ref{K_biased}) will be a biased estimator of $K(t)$
(underestimating $K(t)$),
if $\vartheta(z,z')=1$ for all $z,z' \in \W$. Hence, we choose a correction coefficient $\vartheta(z,z')$ for any $z \in \Z$ that is proportional to the volume of $B(z,t) \cap \W$. We use the translation correction method throughout this work \citep{baddeley2015}. Although $\widehat{K}(t)$ can be computed for any value of $t>0$, an appropriate $t_{max} \geq t$ value is in order,
since high values of $t$ may increase the variance of $\widehat{K}(t)$ drastically.
Hence, we restrict $t_{max}$ depending on the window geometry; for example, a $t_{max}$ value equal to a quarter of the smaller side length of a rectangular window is usually recommended \citep{baddeley2015}.

We test whether the realizations of these random events are drawn from a homogeneous Poisson process
(or from events of any other spatial distribution; but we assume complete spatial randomness (CSR) as the benchmark here).
However, the distribution of $\widehat{K}(t)$ is unknown.
Under the assumption of CSR, it is known that $K(t)=\pi t^2$ in $\R^2$, i.e. the area/volume of a ball with radius $t$. To test the null hypothesis, we use a set of simulated data sets to build confidence bands for $\widehat{K}(t)$ values.
Each simulated data set is a random sample of uniformly distributed points inside $\W$.
We record minimum and maximum $\widehat{K}(t)$ for each value of $t$. These values are equivalent to $\alpha$'th and $(1-\alpha)$'th quantiles, respectively,
which establish 95\% confidence bands with $N=19$ Monte Carlo replications and 99\% confidence bands with $N=99$ replications. Hence, the more the number of replications $N$, the less stringent the test is. We test if the $\widehat{K}(t)$ of the data set is equal or greater (less)
than the $N(1-\alpha/2)$'th ($N(\alpha/2)$'th) value of all $\widehat{K}(t)$ values
(from both simulated data sets and the real data set).
If this is the case for any value of $t$, the null hypothesis is rejected.
A variant of the $K$ function,
$L(t)-t$, of $t$ provides better visualization of the envelope which is defined as follows: for $\Z \subset \R^2$ \citep{ripley1979},
\begin{equation}
L(t):=\sqrt{\frac{K(t)}{\pi}}.
\end{equation}
The estimate $\widehat{L}(t)$ is defined similarly. It is easy to see that, for $K(t)=\pi t^2$, we have $L(t)-t=0$. Hence, the envelope of $L(t)-t$ provides confidence bands around $0$ which is visually more convenient. In Figure~\ref{fig:K_t_pois}, we illustrate three example sets of point patterns in the unit square window $\W$ in $\R^2$ where one is drawn from a homogeneous point pattern, and others are clustered processes (where some regions in the window either have different local density or zero Lebesgue measure), and we also provide the $\widehat{L}(t)-t$ curves.

In Figure~\ref{fig:K_t_pois}, upper and lower dashed curves for the envelopes represent the maximum and minimum possible values of $\widehat{L}(t)-t$ corresponding to each value of $t$. If the $\widehat{L}(t)-t$ curve of the data set is not (entirely) inside the  envelope, it is concluded that there is sufficient evidence that the data set deviates significantly from a homogeneous Poisson process (the data set does not exhibit CSR). As illustrated with a window of one or two clusters in Figure~\ref{fig:K_t_pois}, the curve of the data set is outside the envelopes which indicates the rejection of the null hypothesis. We use the function $\widehat{K}(t)$ to find regions that encapsulate subsets of the data sets which follow CSR inside the covering ball. We establish a collection of subsets given by these regions (viewed as a window $\W$) such that the union of such collection of regions is used to estimate the supports of hidden classes.

\begin{figure}[t]
\centering
\begin{tabular}{ccc}
\includegraphics[scale=0.33]{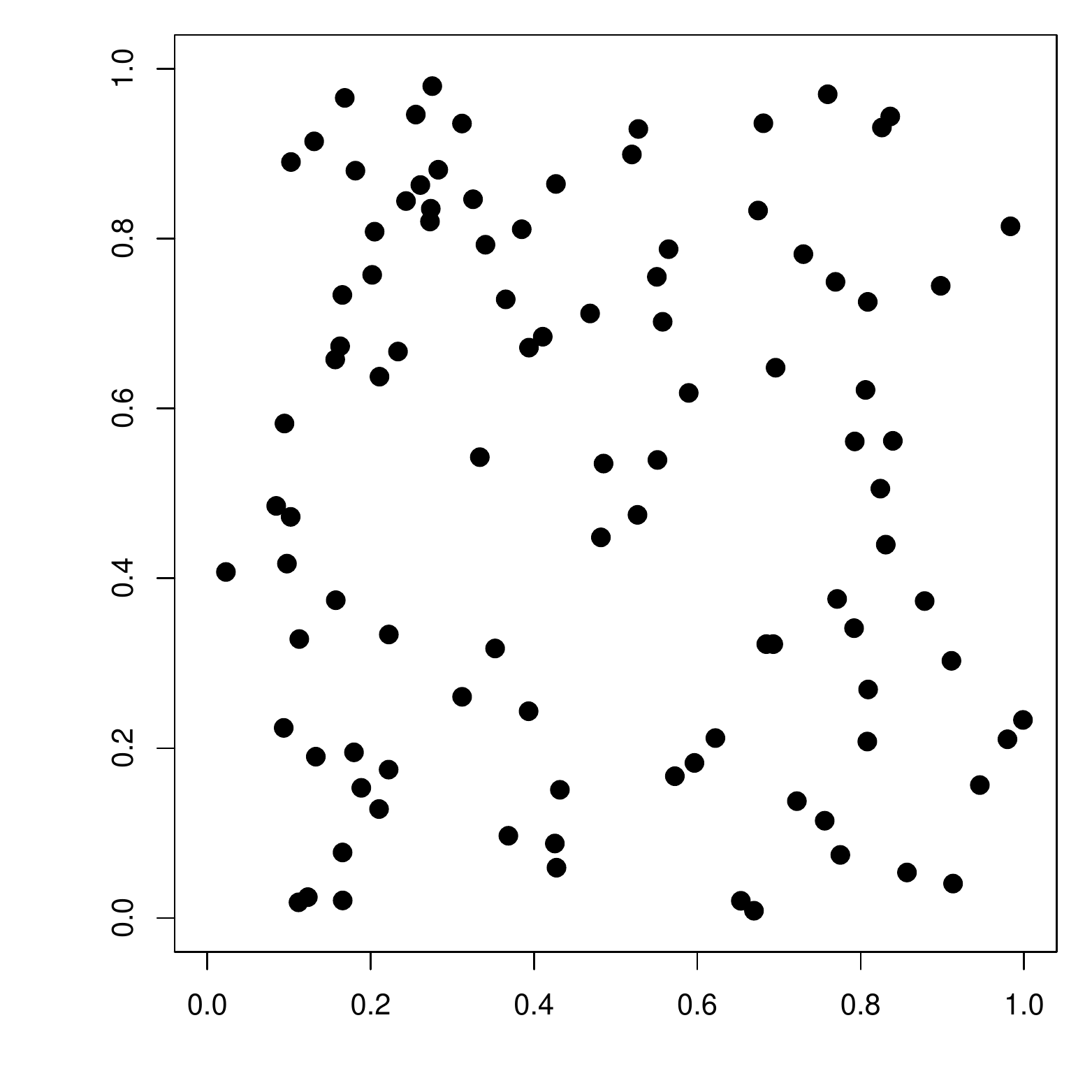} & \includegraphics[scale=0.33]{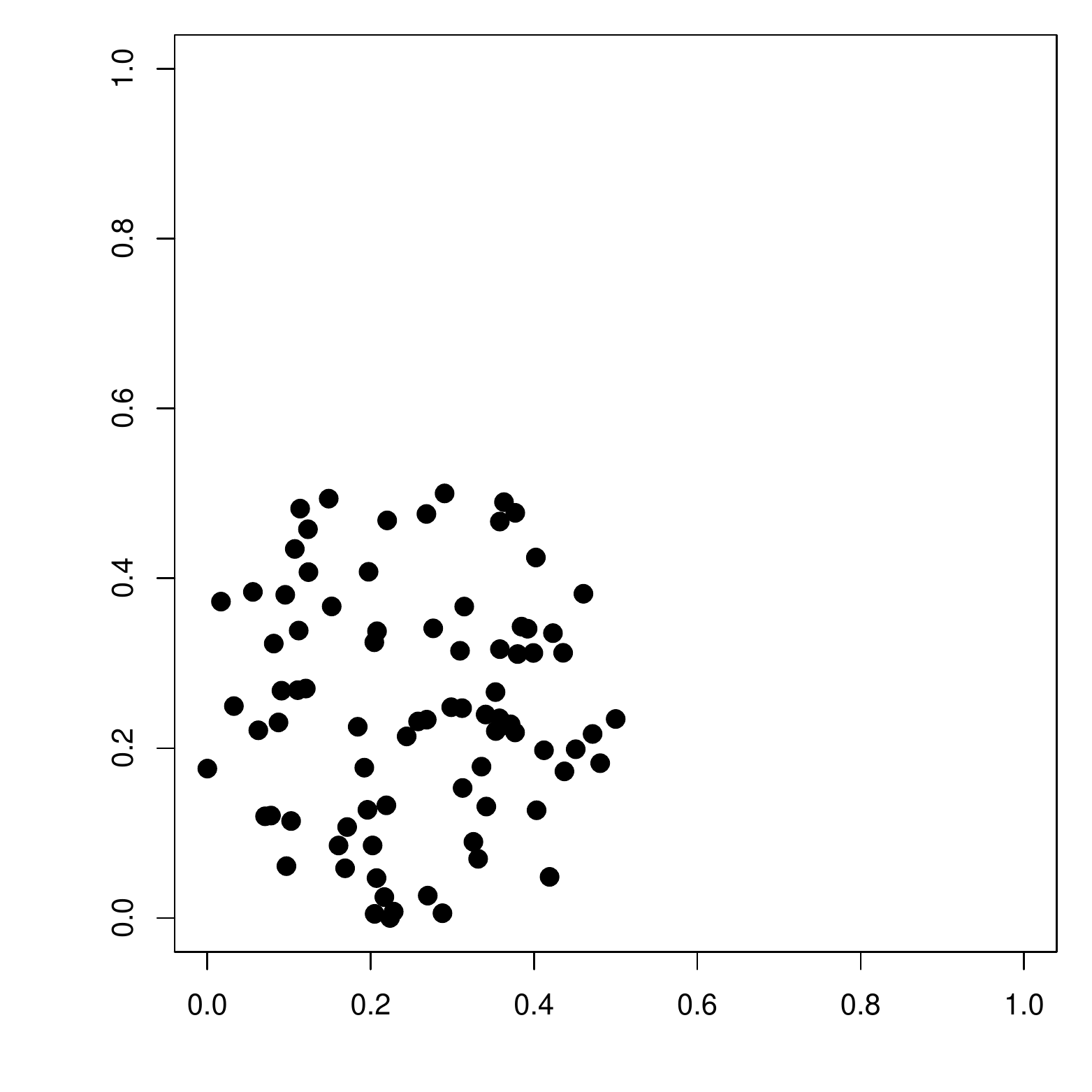} & \includegraphics[scale=0.33]{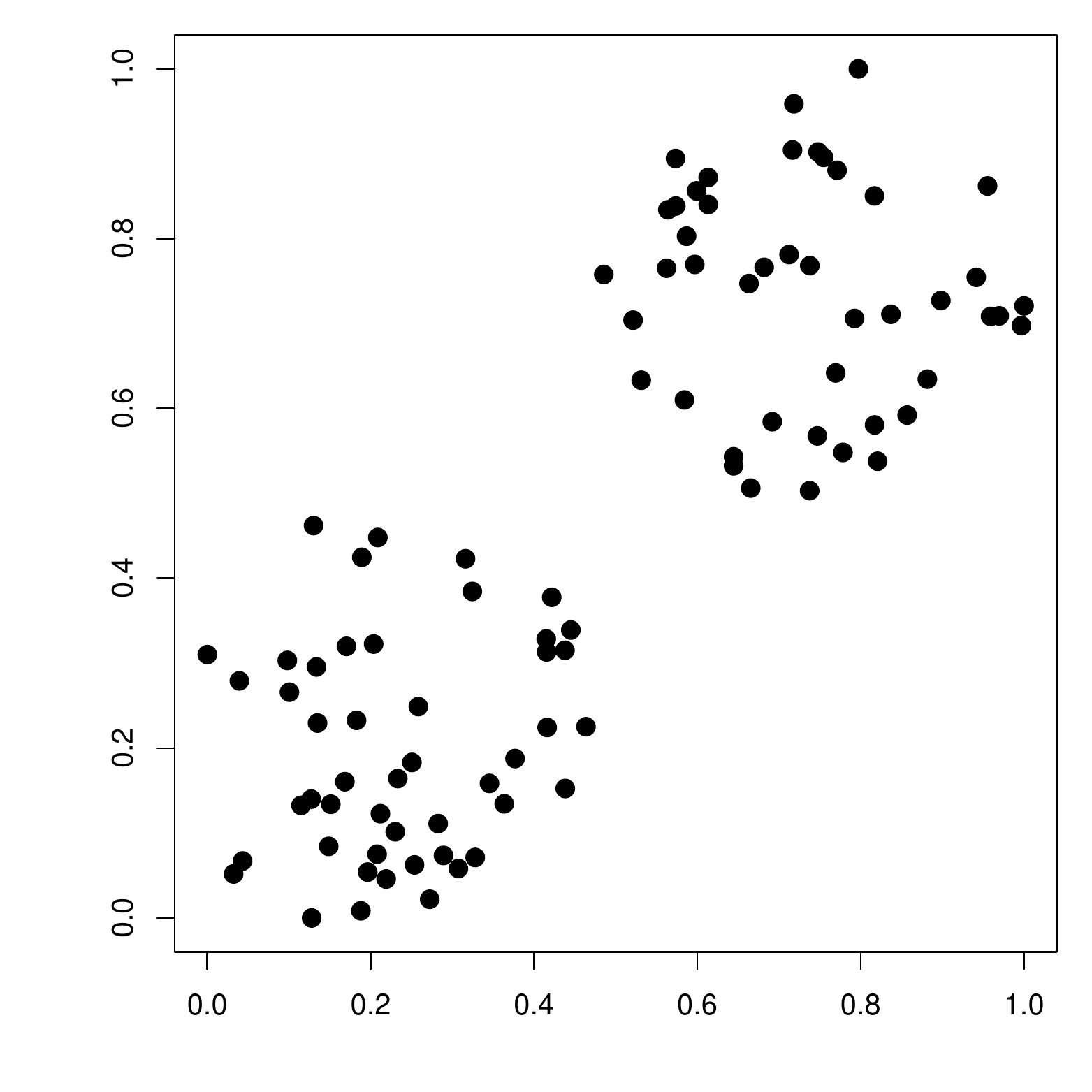} \\
\includegraphics[scale=0.33]{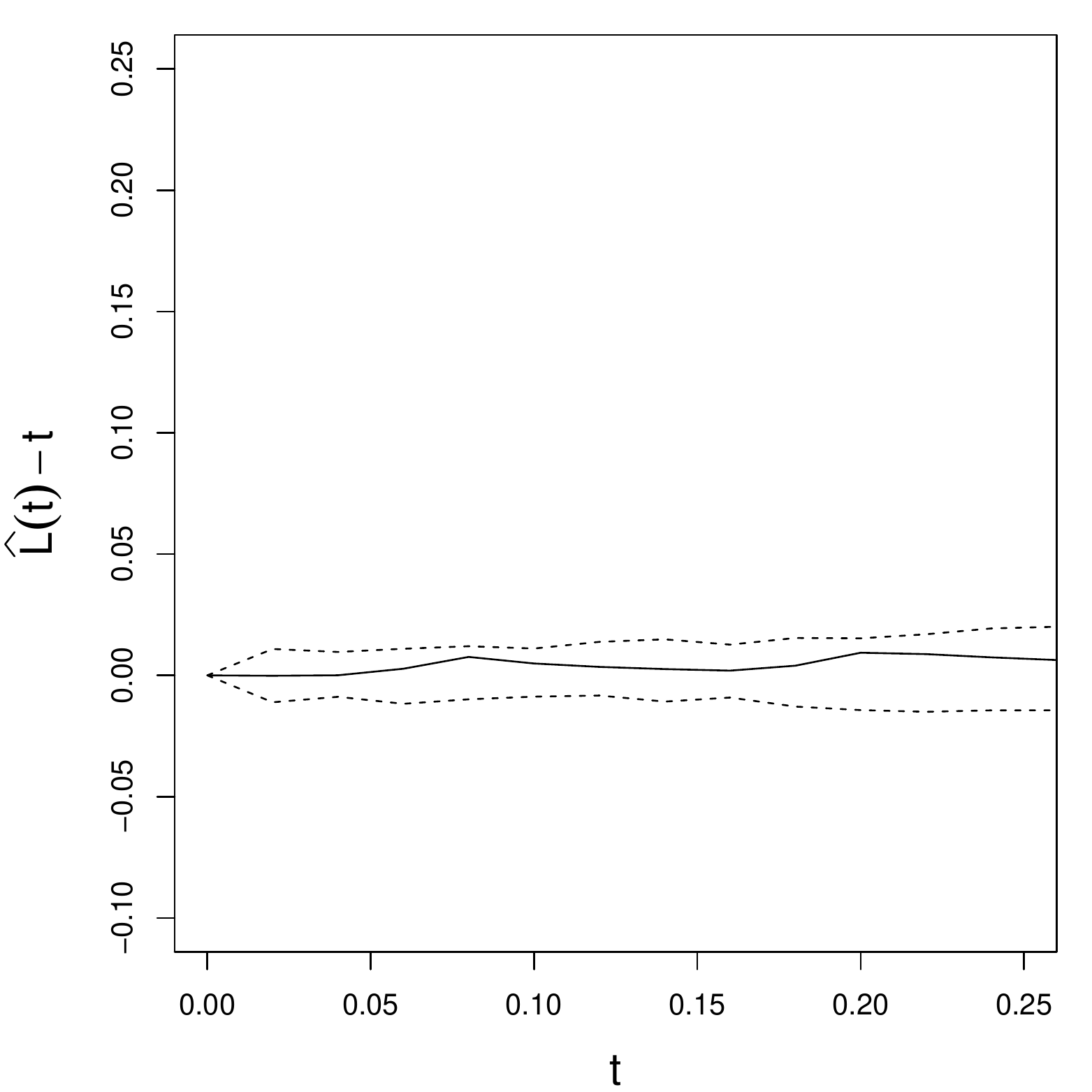} & \includegraphics[scale=0.33]{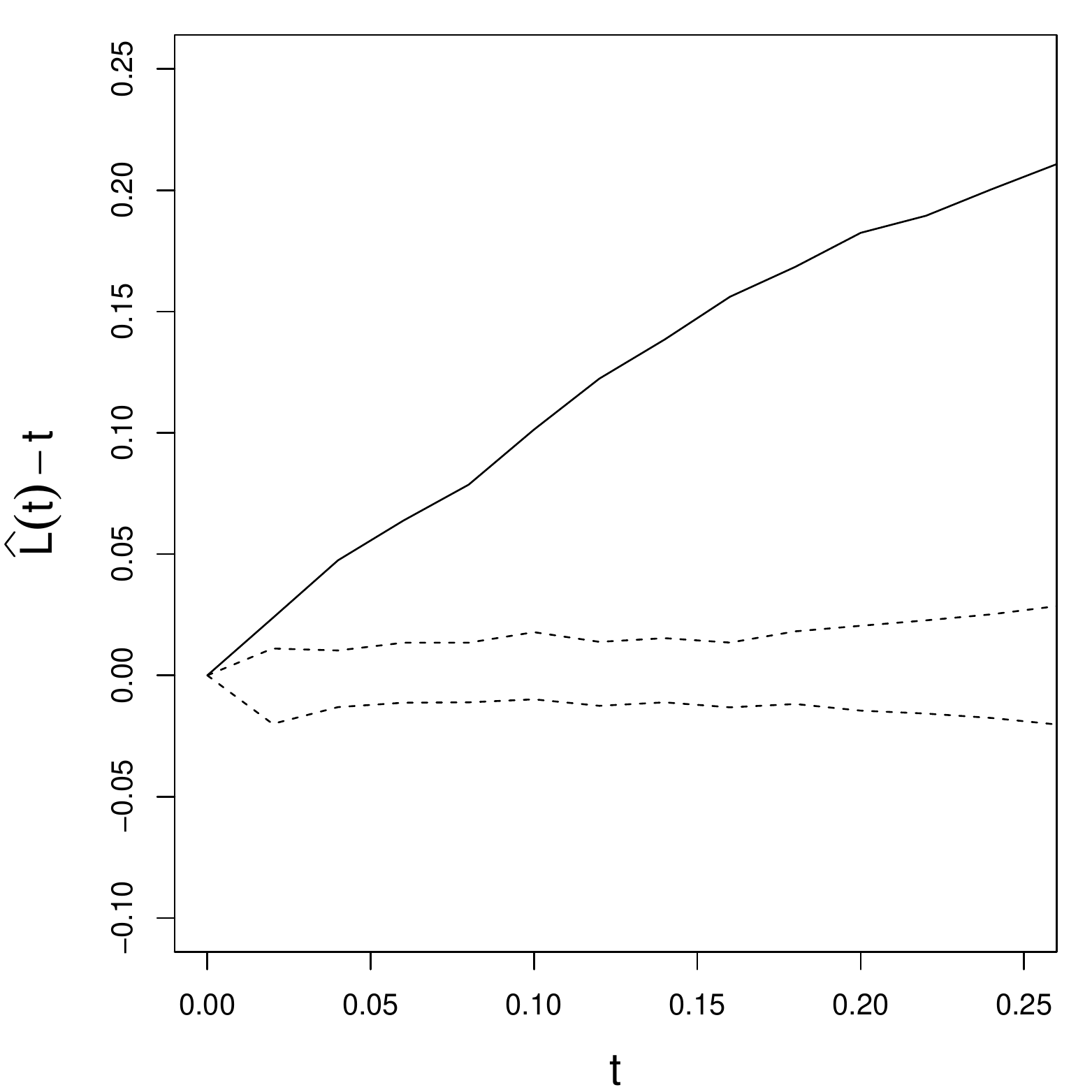} & \includegraphics[scale=0.33]{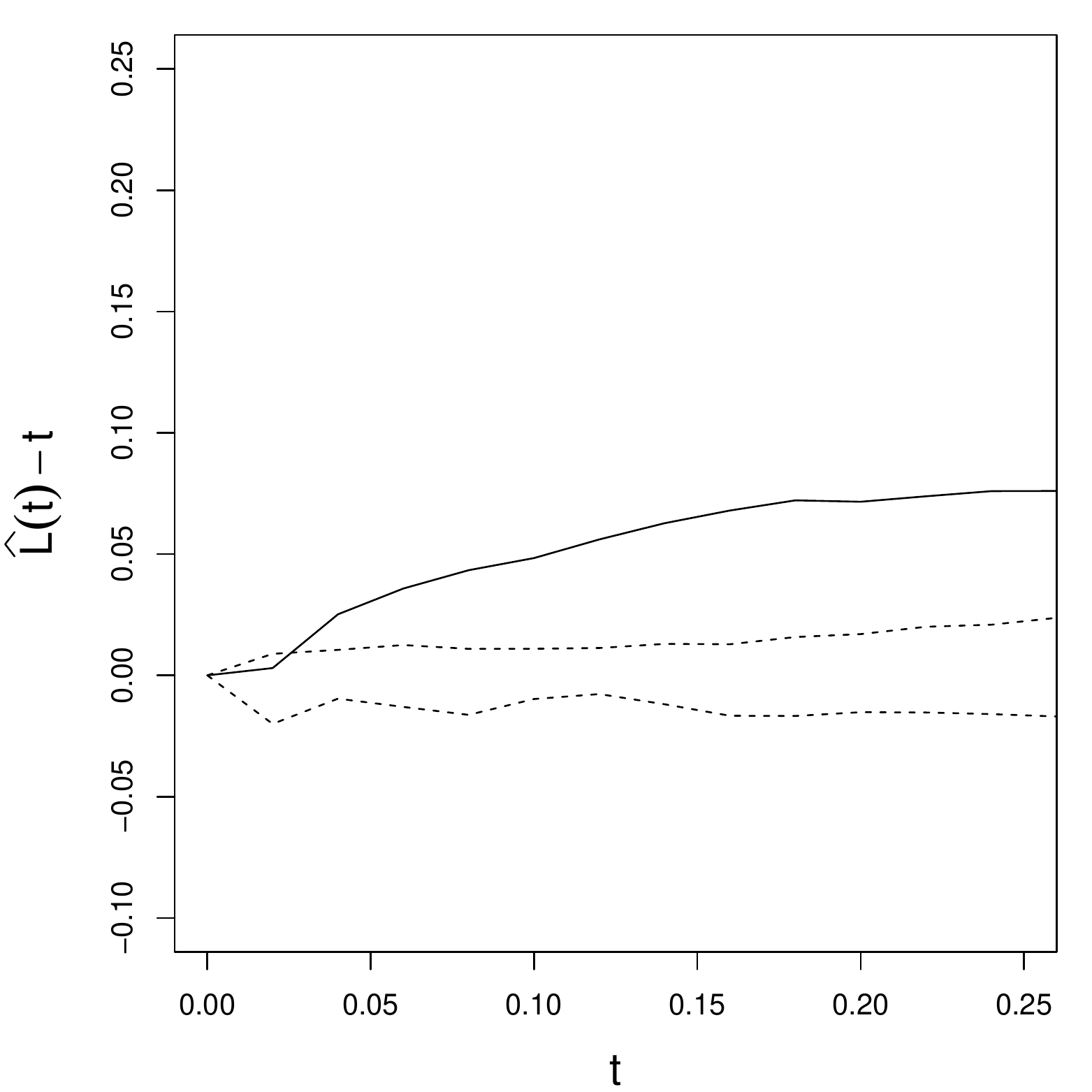} \\
(a)& (b) & (c) \\
\end{tabular}
\caption{Three examples of point patterns where
(a) a realization of homogeneous Poisson process,
(b) a clustered process with one cluster,
(c) and a clustered process with two clusters. The corresponding 99\% envelopes and the $\widehat{L}(t)-t$ are given at the bottom row.}
\label{fig:K_t_pois}
\end{figure}

\section{Clustering with CCDs} \label{sec:ccdclust}

Let $\X=\{X_1,X_2,\ldots,X_n\}$ be a set of $\R^d$-valued random variables with some distribution $F$ and the support $s(F)$, and assume $X_i$ is drawn from a finite mixture of distributions. The objective is to find the number of these components (i.e. clusters) of the mixture, as well as from which component each point $X_i$ is more likely drawn. We estimate the number of clusters $\K$, denoted as the estimand $\widehat{\K}$, and offer algorithms to partition the data set into estimated number of disjoint sets. To do so, we use the lower complexity covers of CCDs to reveal the hidden clusters provided by the covering balls associated with approximate minimum dominating sets whose elements are equivalent to the cluster centers.

Although the lower complexity cover gives an estimate of some hidden class supports, not all covering balls are equivalent to hidden clusters, since CCDs (similar to CCCDs) are vector quantization methods, i.e. the support is modeled by a set of prototypes \citep{marchette2004}. CCDs exploit the relationship between the set of covering balls and prototypes to estimate possibly existing clusters \citep{marchette2004,devinney2003}.
It is assumed that covering balls for the same cluster are more likely to intersect as, by construction, their centers would be closer to each other. Hence, MDSs, found by Algorithm~\ref{alg:dom_greedy_outdegree}, are more appealing,
since the resulting covering balls are more likely to intersect than ones found by Algorithm~\ref{alg:dom_greedy}. Two covering balls intersect if and only if the neighborhoods of vertices in CCDs share at least one vertex. Let $D=(\V,\A)$ be a digraph such that, for $v \in \V$, we define the neighborhood of $v$ as $N(v):=\{u \in \V: (v,u) \in \A\}$. Therefore, let the graph $G_{MD}=(\V_{MD},\E_{MD})$ be defined by $\V_{MD}=S_{MD}$ where,
for $v \in \V_{MD}$ and for $v,u \in \V_{MD}$, we have $\{v,u\} \in \E_{MD}$ if and only if $N(v) \cap N(u) \neq \emptyset$.
Hence, the graph $G_{MD}$ is referred to as an \emph{intersection graph} \citep{das1989}.
CCDs use $S(G_{MD})$ to denote the (approximate) MDS of $G_{MD}$.
We assume that the higher the number of points a ball covers, the closer the ball to the center.
Hence, we first order the elements of $S(G_{MD})$ from highest to lowest associated covering ball cardinality. Starting from the first two elements of $S(G_{MD})$, we incrementally add points of $S(G_{MD})$ to a set $S$ until the silhouette measure reaches to a maximum value.
Here, $S$ is the estimated set of true cluster centers with $|S| = \widehat{\K}$. If a maximum or substantially high silhouette measure is obtained, this suggests that there is no need to look for more clusters, and thus, the remaining clusters are viewed as potential noise clusters.

We illustrate the cover of an example simulated data set with two clusters in Figure~\ref{fig:ccd_dom}(a). Observe that, some covering balls associated with $S_{MD}$ intersect with each other, however there seem to be two sets of covering balls that do not intersect with one another. These two sets of covering balls may represent seperate estimates of two class supports that could be the clusters we are looking for. We illustrate the covering balls of $S(G_{MD})$ in Figure~\ref{fig:ccd_dom}(b). Although the cardinality of the set $S_{MD}$ is seven; that is, the MDS of CCD, $D$, has seven elements, two big covering balls are those with the highest cardinality. The centers of these two covering balls are elements of the MDS and they are closely located at the center of the clusters. Finally, we illustrate another simulated data set with its associated the covering balls of $S(G_{MD})$ and $S$ to demonstrate the use of silhouette measure in Figure~\ref{fig:ccd_dom_noise}. The data set has three bivariate normally distributed components (or clusters) such that some low cardinality covering balls may be realized as including unnecessary points or noise. The silhouette measure is maximized when the three most dense covering balls are added to the final set of MDSs that are the true clusters. In $k$-means algorithm, the silhouette measure should be computed again for the selection of the best $k$ in separating the data set. In CCDs, however, we use the silhouette measure to separate true clusters from the noisy clusters. Hence, we terminate Algorithm~\ref{alg:dom_greedy_score} when the $sil(\mathcal{P})$ is maximized, hence the set of remaining vertices are said to belong to the noise clusters (see Algorithms~\ref{alg:ccdks} and~\ref{alg:ccdripley}).

\begin{figure}[h]
\centering
\begin{tabular}{cc}
\includegraphics[scale=0.35]{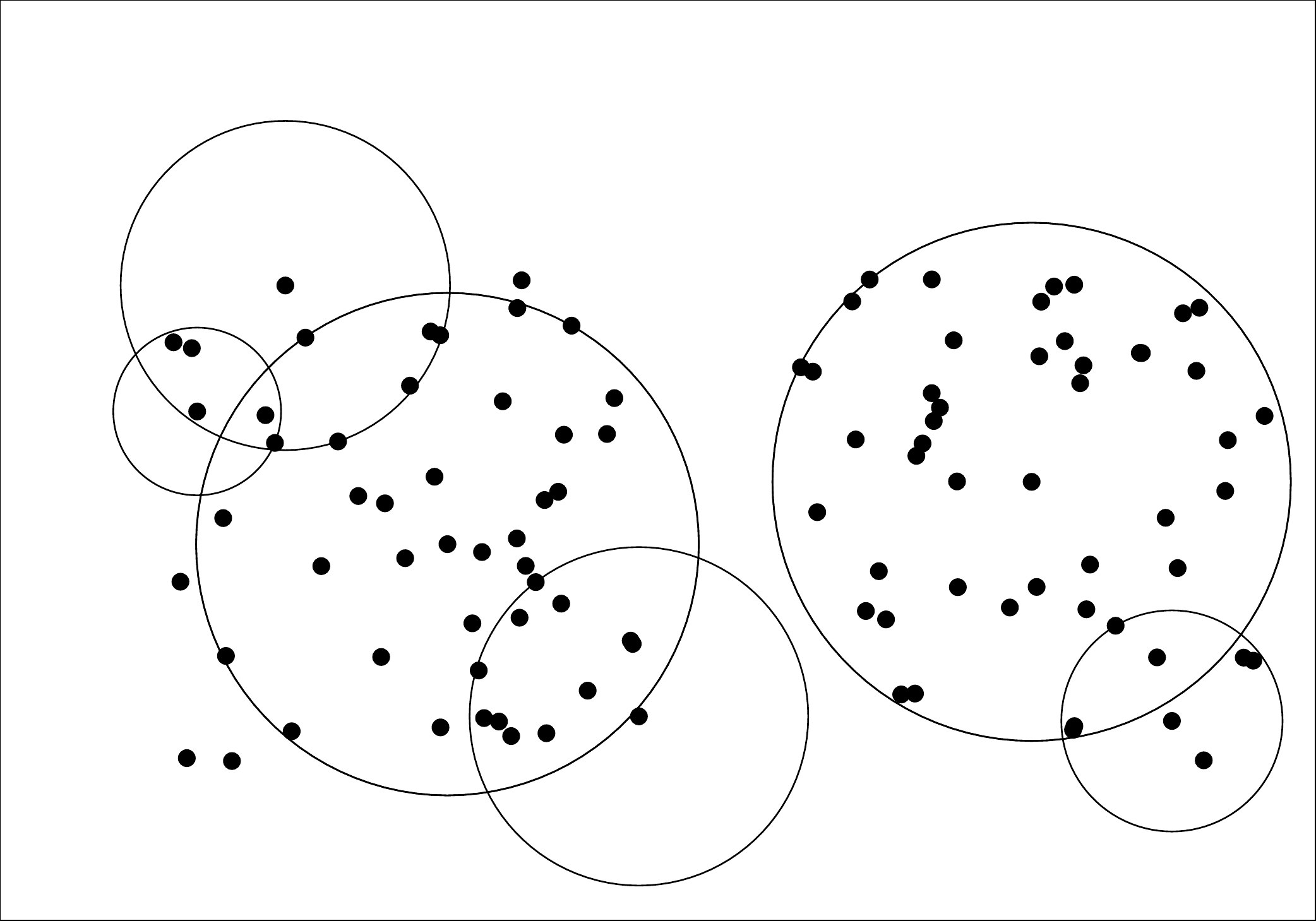} & \includegraphics[scale=0.35]{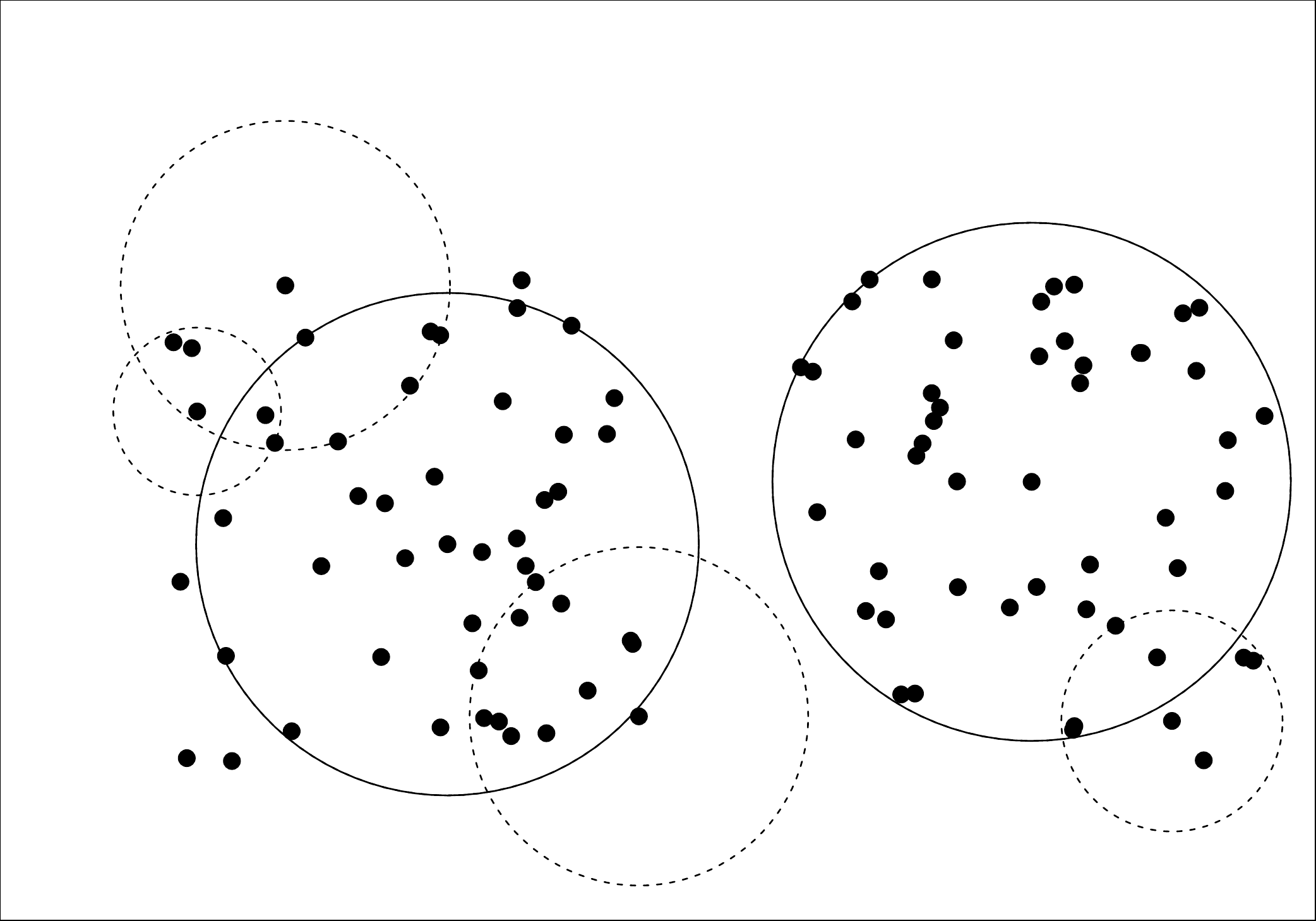} \\
(a) & (b) \\
\end{tabular}
\caption{(a) Covering balls based on the points of the MDS, $S_{MD}$.
The data set is composed of two sets of size $n=50$ randomly drawn from bivariate uniform distributions with $F_1=U([0,1]^2)$ and $F_2=U([2.5,3.5] \times [0,1])$.
Here, we take the intensity parameter to be $\delta=0.2$. (b) Covering balls of $S(G_{MD})$. Balls of dominating points of $S(G_{MD})$ are given with solid lines, and balls of points of $S_{MD} \setminus S(G_{MD})$ with dashed lines. }
\label{fig:ccd_dom}
\end{figure}

\begin{figure}
\centering
\begin{tabular}{cc}
\includegraphics[scale=0.35]{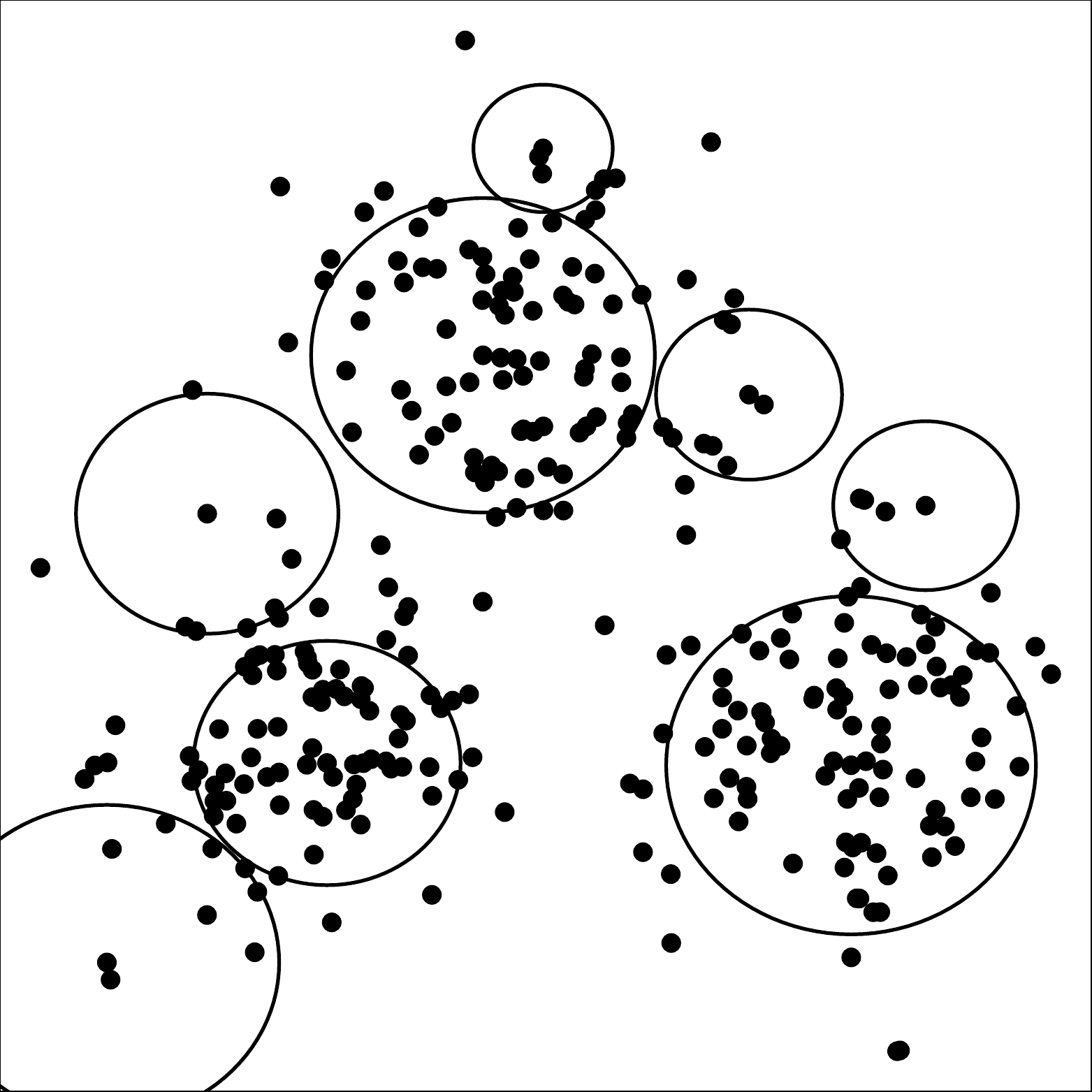} & \includegraphics[scale=0.35]{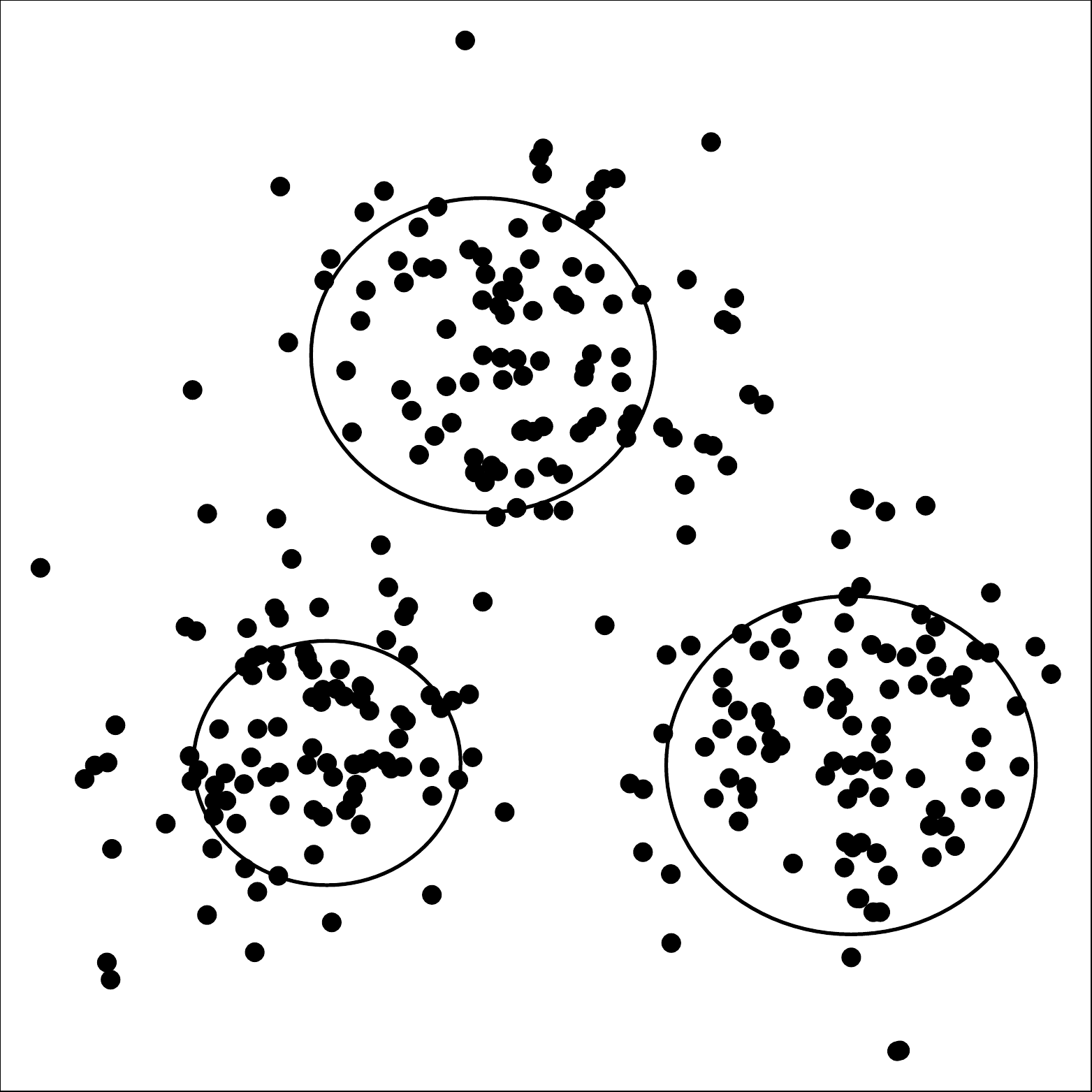} \\
(a) & (b) \\
\end{tabular}
\caption{Clustering of a data set with three bivariate normally distributed classes in $\R^2$.
(a) Centers of clusters that are either the true or the noise clusters.
(b) The true clusters which maximize the average silhouette index in the data set.}
\label{fig:ccd_dom_noise}
\end{figure}

Some points in the data set, however, are not covered by any ball since no significant clustering occurs around them,
or some of them are neglected,
since they were the members of the noise cluster(s). After finding $S$, we look for closest covering balls to decide which clusters these noise points belong to. Given a convex set $\mathcal{H} \subset \Omega$, for $x \in \mathcal{H}$, we use the distance function $\rho(x,z)$ of the form
\begin{equation}
\rho(z,\mathcal{H}):=\frac{d(z,x)}{d(y,x)},
\end{equation}
where, $y$ is the point of intersection of the line segment $L(x,z)$ and the boundary of the convex set  $\partial(\mathcal{H})$ in Figure~\ref{fig:convdist}(a) and the covering ball $\partial(B)$ in Figure~\ref{fig:convdist}(b) \citep{manukyan2017}.

\begin{figure}
\centering
\begin{tabular}{cc}
\includegraphics[scale=0.35]{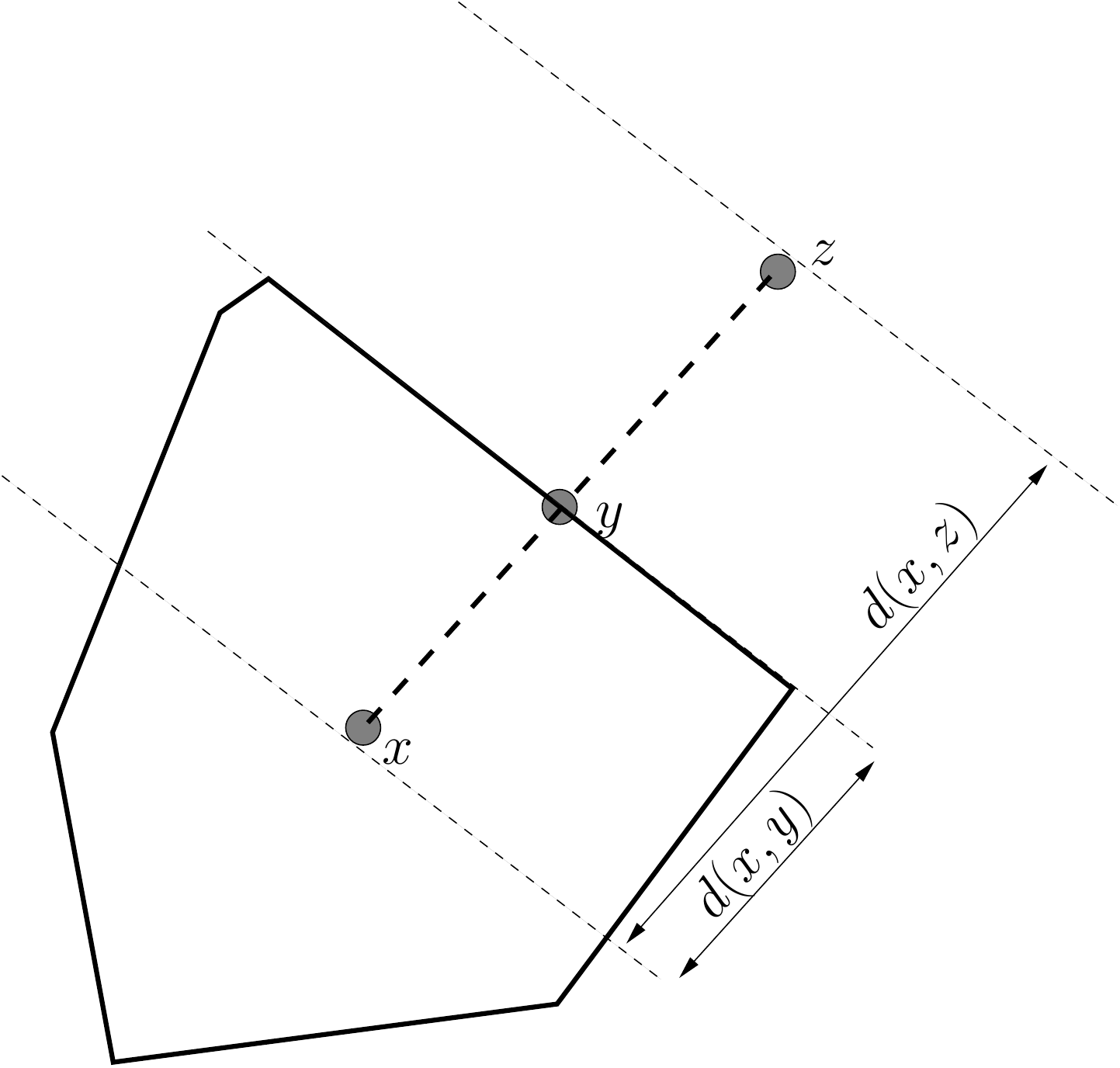} & \includegraphics[scale=0.32]{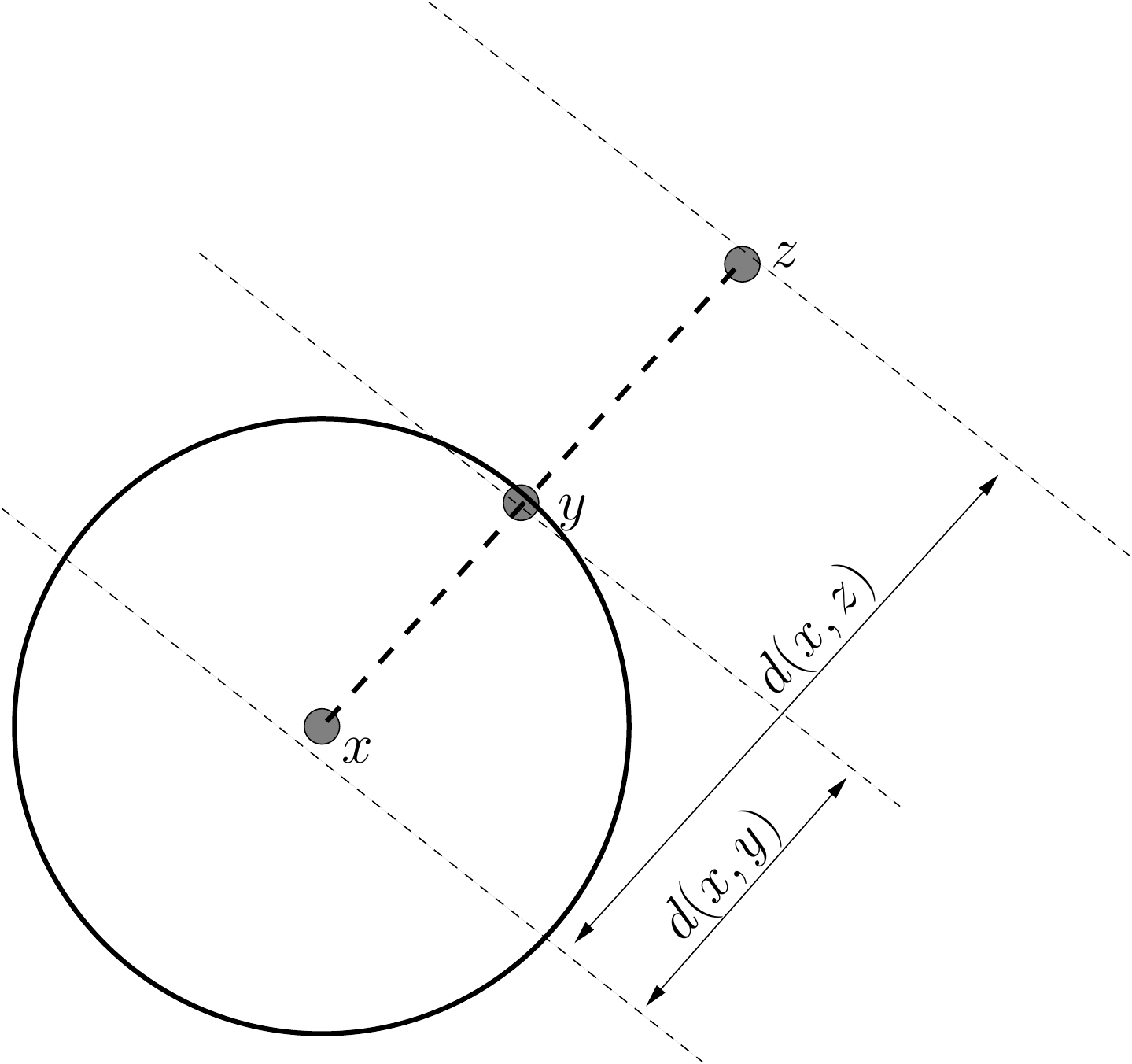} \\
(a) & (b) \\
\end{tabular}
\caption{Illustration of the convex distance from a point $z$ to an (a) arbitrary convex set $\mathcal{H}$ and (b) ball $B$ in $\R^2$.
Here, $x$ is the center of mass of both $\mathcal{H}$ and $B$.
The point $y$ is the point of intersection between the line from $x$ to $z$ and the boundary of the convex set $\mathcal{H}$ and $B$.}
\label{fig:convdist}
\end{figure}

We give the pseudo code of the CCD algorithm with K-S based statistics in Algorithm~\ref{alg:ccdks} \citep{devinney2003}. First, for all points of the data set, we choose the best radii $r(X)$ by maximizing the K-S statistics given in Equation (\ref{k-s}).
Below, we find the MDS, $S_{MD}$, which constitutes an estimate of the support of a possible hidden class.
Then, to reveal cluster centers and the number of hidden classes, we construct the intersection graph with the vertex set $S_{MD}$, and find its own dominating set $S(G_{MD})$. Finally, we incrementally add more points of the $S(G_{MD})$ to $S$ until the silhouette measure is maximized.
This algorithm runs in $\mathcal{O}(n^3)$ time for $d < n$.

\begin{algorithm}
\begin{algorithmic}[1]
 \REQUIRE The data set $\X$ and the intensity parameter $\delta \in \R_+$
 \ENSURE The set of centers, $S$, and their corresponding radii
 \FORALL{$x \in \X$}
 	\STATE $r(x) = \argmax_{r \in \{d(x,z): z \in \X\}} RW(x,r)-\delta r^d .$ 	
 \ENDFOR  		
 \STATE Construct $D=(\V,\A)$
 \STATE Find $S_{MD}$ with Algorithm~\ref{alg:dom_greedy_outdegree}
 \STATE Construct $G_{MD}=(\V_{MD},\E_{MD})$
 \STATE $sc(v) \leftarrow |\bar{N}(v)|$
 \STATE Find $S(G_{MD})$ with Algorithm~\ref{alg:dom_greedy_score} (stop when $sil(\mathcal{P})$ is maximized)
 \STATE $S \leftarrow S(G_{MD})$
\end{algorithmic}
\caption{CCD clustering algorithm with K-S based statistics}
\label{alg:ccdks}
\end{algorithm}

\begin{theorem} \label{thm:complexity}
Let $\X \subset \R^d$ be a data set with $n=|\X|$ observations. Algorithm~\ref{alg:ccdks} partitions the data set $\X$ in $\mathcal{O}(n^3 + n^2 (d+\log{n}))$ time.
\end{theorem}	

\noindent {\bf Proof:}
In Algorithm~\ref{alg:ccdks}, the matrix of distances between points of the set $\X$ should be computed which takes $\mathcal{O}(n^2d)$ time. For each iteration, the radius $r(x)$ that maximizes $RW(x,r)-\delta r^d$ could easily be computed by sorting the distances from each $x$ to all other points which takes $\mathcal{O}(n \log{n})$ time for each $x \in \X$. Then, $r(x)$ is found on linear time for each $x \in \X$. Hence, the digraph $D$ is established in a total of $\mathcal{O}(n^2\log{n})$ running time. Both Algorithm~\ref{alg:dom_greedy_outdegree} and Algorithm~\ref{alg:dom_greedy_score} run in $\mathcal{O}(n^2)$ time in the worst case. Finally, to maximize the silhouette, for each added element of $S(G_{MD})$, we partition the data set and update the silhouette, taking $\mathcal{O}(n^2)$ time in the worst case for each $s \in S(G_{MD})$. Hence, maximizing the silhouette takes $\mathcal{O}(n^3)$ time in the worst case. $\blacksquare$

Here, the choice of the intensity parameter $\delta$ is of utmost importance; that is, CCDs provide a variety of different results for different values of $\delta$. Figure~\ref{fig:inter_dom} illustrates covering balls of $S(G_{MD})$ for four different values of the intensity parameter $\delta$. Observe that the best result is obtained by $\delta=0.2$; however, just a single cluster (no clusters) is found with $\delta=0.05$, three clusters are found with $\delta=0.4$, and three small clusters have been found for $\delta=0.6$. Obviously, none of these $\delta$ values provide the true partitioning other than $\delta=0.2$. In the following section, we introduce new CCDs where we drop the necessity of choosing the parameter $\delta$.

\begin{figure}[t]
\centering
\setlength{\tabcolsep}{18pt}
\begin{tabular}{cc}
\includegraphics[scale=0.35]{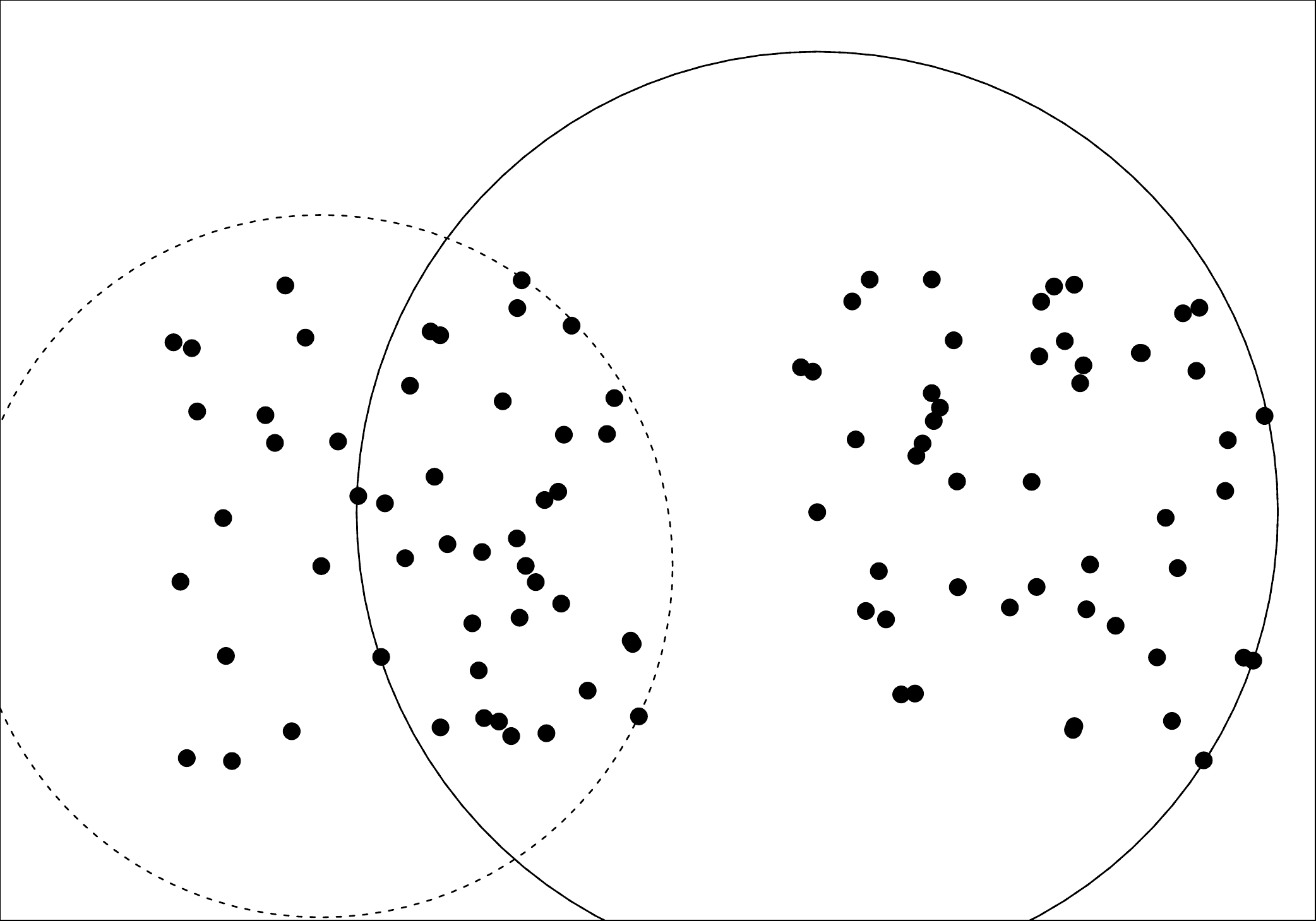} & \includegraphics[scale=0.35]{ccd_dom_m020-eps-converted-to.pdf} \\
(a) $\delta=0.05$ & (b) $\delta=0.2$ \\[15pt]
\includegraphics[scale=0.35]{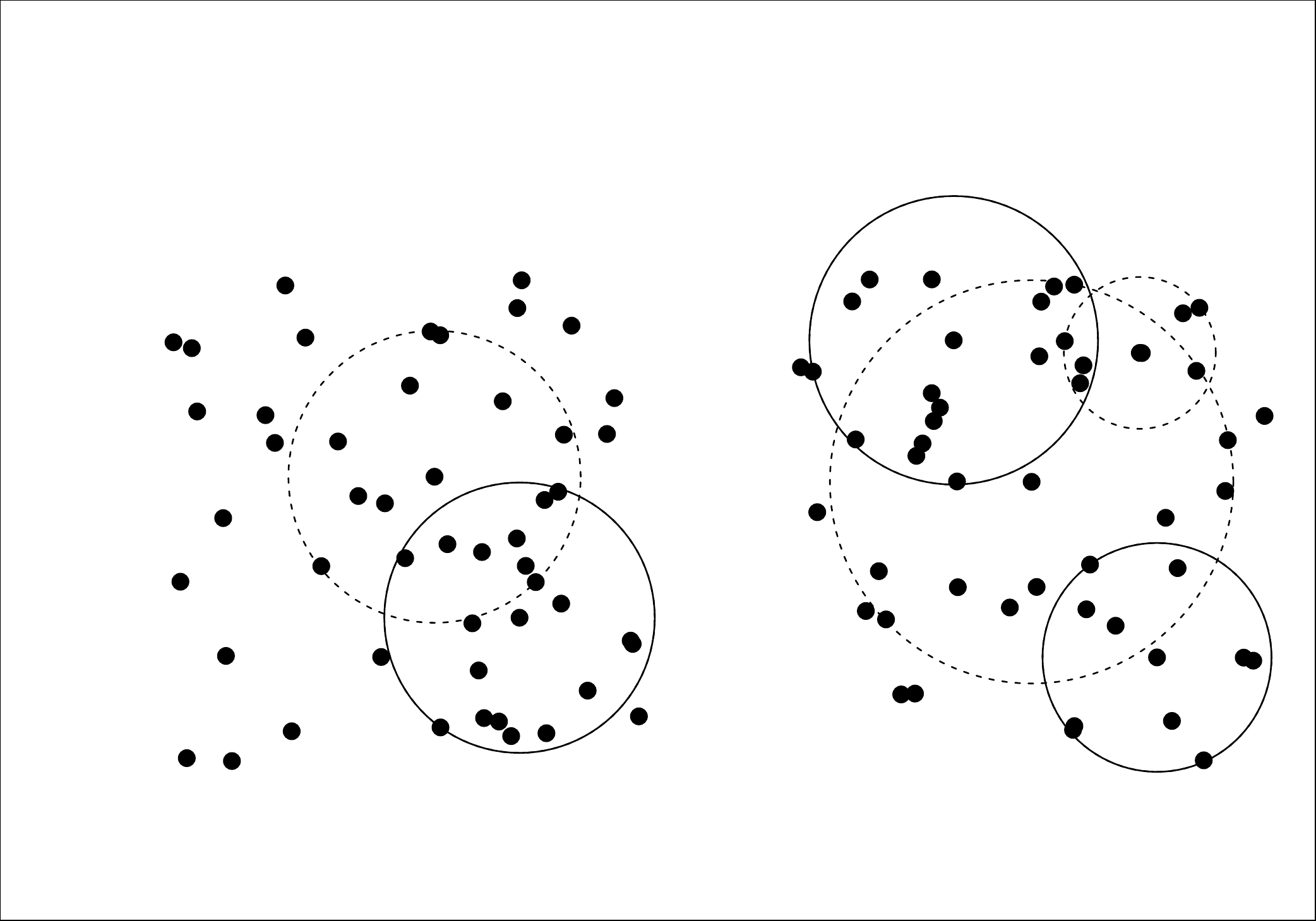} & \includegraphics[scale=0.35]{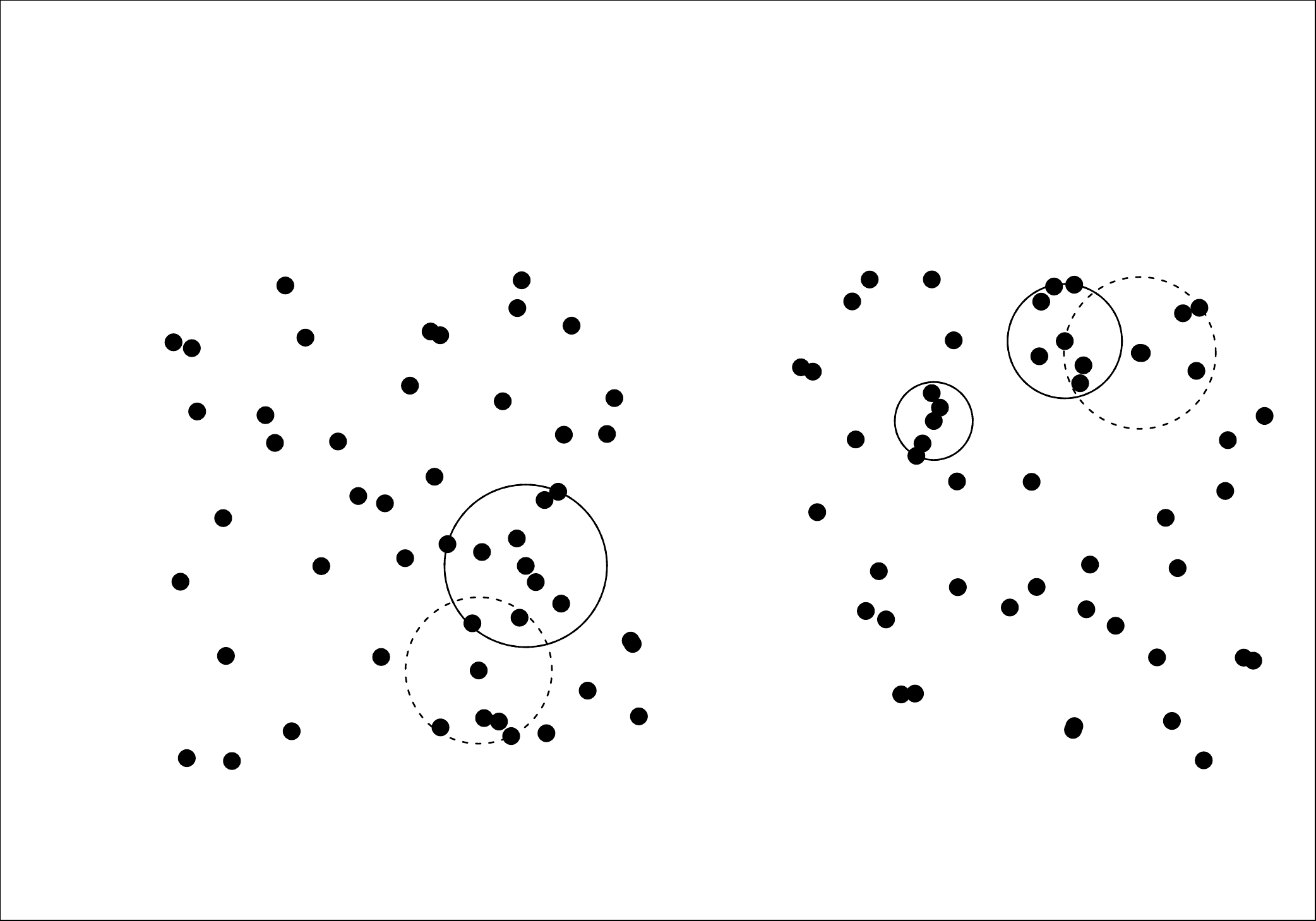} \\
(c) $\delta=0.4$ & (d) $\delta=0.6$ \\
\end{tabular}
\caption{Covering balls of the set $S(G_{MD})$ produced by Algorithm~\ref{alg:ccdks} with different choices of the intensity parameter $\delta$. Here, we consider $\delta=0.05,0.20,0.40,0.60$. Covering balls of the set $S(G_{MD})$ are given with solid lines, and balls of the set $S_{MD} \setminus S(G_{MD})$ with dashed lines.}
\label{fig:inter_dom}
\end{figure}

\section{CCDs with Ripley's $K$ function} \label{sec:ccdripley}

We introduce an adaptation of CCDs where the test based on K-S statistics replaced by those with Ripley's $K$ function.
In CCDs, the K-S statistic in Equation~(\ref{k-s}) is maximized over the radius $r(X)$ to find which radius achieves a covering ball with points inside clustered the most. We will refer to this family of digraphs as KS-CCDs throughout this article. K-S statistics do not test the relative distribution of data points inside the covering ball, and hence may misinterpret a collection of points from two clusters as a single cluster. On the other hand, tests based on the $K$ function exploit the second-order properties of a spatial distribution, and hence may potentially give a better description of clustering. We use the $K(t)$ function to test whether points inside a ball (or a window) are members of the same cluster. Our methods work in a similar fashion to the algorithms of \cite{streib2011}, however we use $\widehat{K}(t)$ to determine the radii of covering balls whereas \cite{streib2011} determine the bandwidth of Gaussian kernels to be used in spectral clustering.

Here, we think of each ball $B$ as a window that we test whether points inside the ball are drawn from a homogeneous Poisson process with $\widehat{\lambda}=n/\vol(B)$ (i.e. complete spatial randomness), where $n$ is the size of the data set and $\vol(B)$ is the area (or volume) of the ball $B$.
Hence, we employ a spherical window that we use the translation edge correction to mitigate the effects of bias resulting from the edges \citep{ohser1983}. Let the covering ball, or the window, $B$ has radius $R$, and let $SC(a,R)$ be the spherical cap with height $a$ of the $d$-sphere $B$ (a sphere of dimension $d$).
Hence, given the distance between the pair $z,z' \in \R^d$ denoted by $\rho(z,z')$, let the translation correction coefficient $\vartheta(z,z')$ for Equation~(\ref{K_biased}) be defined as:
\begin{equation}
\vartheta(z,z'):=\frac{\vol(B)}{2\vol(SC(a,R))}=\frac{\frac{\pi^{d/2}R^d}{\Gamma(\frac{d}{2}+1)}}{2\frac{\pi^{(d-1)/2}R^d}{\Gamma(\frac{d+1}{2})}\int_0^{\arccos(\frac{\rho(z,z')}{2R})} \sin^d(t) dt}.
\end{equation}
The test rejects the null hypothesis, if there are significantly empty regions or there is sufficient evidence that clusters exists inside the ball.
In the test, we do only check if the observed curve is above the upper envelope to test for possible clustering. If the curve of $\widehat{K}(t)$ is below the lower curve of the envelope, the test for CSR rejected and point pattern of the data is significantly regular. However, we are not interested with regular spatial patterns at this point, since we only test if the point pattern inside the ball constitute clusters \citep{ripley1977}.

The envelopes are empirically computed given a set of Monte Carlo experiments on a ball, or window, with some radius and center. However, to decrease the running time of the test, we rely on the invariance property of the $\widehat{K}$  function under linear transformation of random sequences whose support is a ball, or the window, itself. Let $\mathcal{G}$ be the set of all linear transformations $g \in \mathcal{G}$ on $\R^d$ of the form $g(x)=ax+b$ for $a \in \R \setminus \{0\}$, $b \in \R$ and $x \in \R^d$, then it can easily be seen that $\mathcal{G}$ is a \emph{transformation group} since the group is closed under composition and inversion, and the $\mathcal{G}$ also includes the identity transformation (i.e. $g(x)=x$). The following theorem shows how a single set of Monte Carlo experiments could be used for multiple covering balls given a set of linear transformations $\mathcal{G}$.

\begin{theorem} \label{thm:Kfunc}
Let $\Z=\{Z_1,Z_2,\ldots,Z_n\}$ be a set of $\R^d$-valued random variables whose support $B(\mathbf{0}_{d\times1},1) \subset \R^d$ is a hyperball with radius 1 and centered on $\mathbf{0}_{d\times1}$ (i.e. zero vector in $\R^d$). Then, the test for CSR with $\widehat{K}$ function of window $\W = B(\mathbf{0}_{d\times1},1)$ is invariant under the transformation group $\mathcal{G}$.
\end{theorem}


\noindent {\bf Proof:}
For $Z \in \Z$, define by $Z'=g(Z)=aZ+b$ for some $g \in \mathcal{G}$. Hence, $Z'$ has the support $B(\mathbf{b}_{d\times1},a)$ where $\mathbf{b}_{d\times1} = b \cdot \mathbf{1}_{d\times1}$ (i.e. one vector in $\R^d$). Now, let $\widehat{K}(\Z,B,t)$ denote the $\widehat{K}(t)$ associated with the set $\Z$ in window $B$, and let $\Z'=g(\Z):=\{g(Z):Z \in \Z\}$. Then, $$\widehat{K}(\Z,B(\mathbf{0}_{d\times1},1),t)=\widehat{K}(\Z',B(\mathbf{b}_{d\times1},a),at)$$ since $d(x,y) < t$ if and only if $d(g(x),g(y)) < at$ for all $x,y \in B(\mathbf{0}_{d\times1},1)$. $\blacksquare$

Theorem~\ref{thm:Kfunc} shows that, regardless of the center or the radius of the ball $B(X,r(X))$,
a single simulated data is sufficient for computing the upper and lower envelopes of the $\widehat{K}$ if $t$ is scaled accordingly; that is, for all $B(b,a)$ for some  $a \in \R$ and $b \in \R^d$ there exists an equivalent test for a ball $B(\mathbf{0}_{d\times1},1)$. Theorem~\ref{thm:Kfunc} decreases the computation time drastically such that the envelope is only computed once for each subset of $\X$ with cardinality $m=1,2,\ldots,n$. We provide the pseudo code of CCD algorithm with Ripley's $K$ function in Algorithm~\ref{alg:ccdripley}. We call these digraphs RK-CCDs. One difference of Algorithm~\ref{alg:ccdripley} from Algorithm~\ref{alg:ccdks} is how the radii of the covering balls are chosen. Instead of looking at each possible ball for all $X \in \X$, Algorithm~\ref{alg:ccdripley} checks the radii until the test for CSR is rejected.

\begin{algorithm}
\begin{algorithmic}[1]
 \REQUIRE The data set $\X$
 \ENSURE The set of centers $S$
 \STATE Calculate $\widehat{K}(t)$ for each of $N$ data sets simulated in $B(\mathbf{0}_{d\times1},1)$
 \FORALL{$x \in \X$}
 	\STATE $\mathbf{D}(x) := \{d(x,y):y \in \X \setminus \{x\}\}$
 	\FORALL{$r \in \mathbf{D}(x)$ in ascending order}
 		\STATE Let $B:=B(x,r)$ and $\X':= \X \cap B$
 		\STATE $r(x) \leftarrow r$
 		\IF{test is rejected for $\X'$} \STATE break \ENDIF
 	\ENDFOR
 \ENDFOR  		
 \STATE Construct $D=(\V,\A)$
 \STATE Find $S_{MD}$ with Algorithm~\ref{alg:dom_greedy_outdegree}
 \STATE Construct $G_{MD}=(\V_{MD},\E_{MD})$
 \STATE $sc(v) \leftarrow |N(v)|$
 \STATE Find $S(G_{MD})$ with Algorithm~\ref{alg:dom_greedy_score} (stop when $sil(\mathcal{P})$ is maximized)
 \STATE $S \leftarrow S(G_{MD})$
\end{algorithmic}
\caption{CCD clustering algorithm using Ripley's $K$ function (RK-CCD).}
\label{alg:ccdripley}
\end{algorithm}

Figure~\ref{fig:CCDseq} shows how Algorithm~\ref{alg:ccdripley} finds an optimum value of $r(X)$ for a given point $X \in \X$. Here, $t_{max}$ is set to half of radius of the covering ball $B(\mathbf{0}_{d\times1},1)$, i.e. $t_{max}=0.5$. The last panel on the far right illustrates a covering ball centered on a point from one class that encapsulates some points from the other class. The observed curve $\widehat{L}(t)-t$ is above the envelope for some values of $t$; that is, the distribution inside the covering ball significantly deviates from CSR. Hence, we stop increasing the radius, and we use the previous radius as the radius of the covering ball $B$. Although the clusters are considerably close to each other, the test based on the function $\widehat{K}(t)$ rejects the hypothesis, if the covering ball centered on some point $x$ begins to cover regions outside of its respective cluster. In Figure~\ref{fig:CCDseq}, the ball gets larger until it stumbles on a point from another cluster or if there are some empty or sparsely populated regions.


\begin{figure}[!h]
\centering
\includegraphics[scale=0.6]{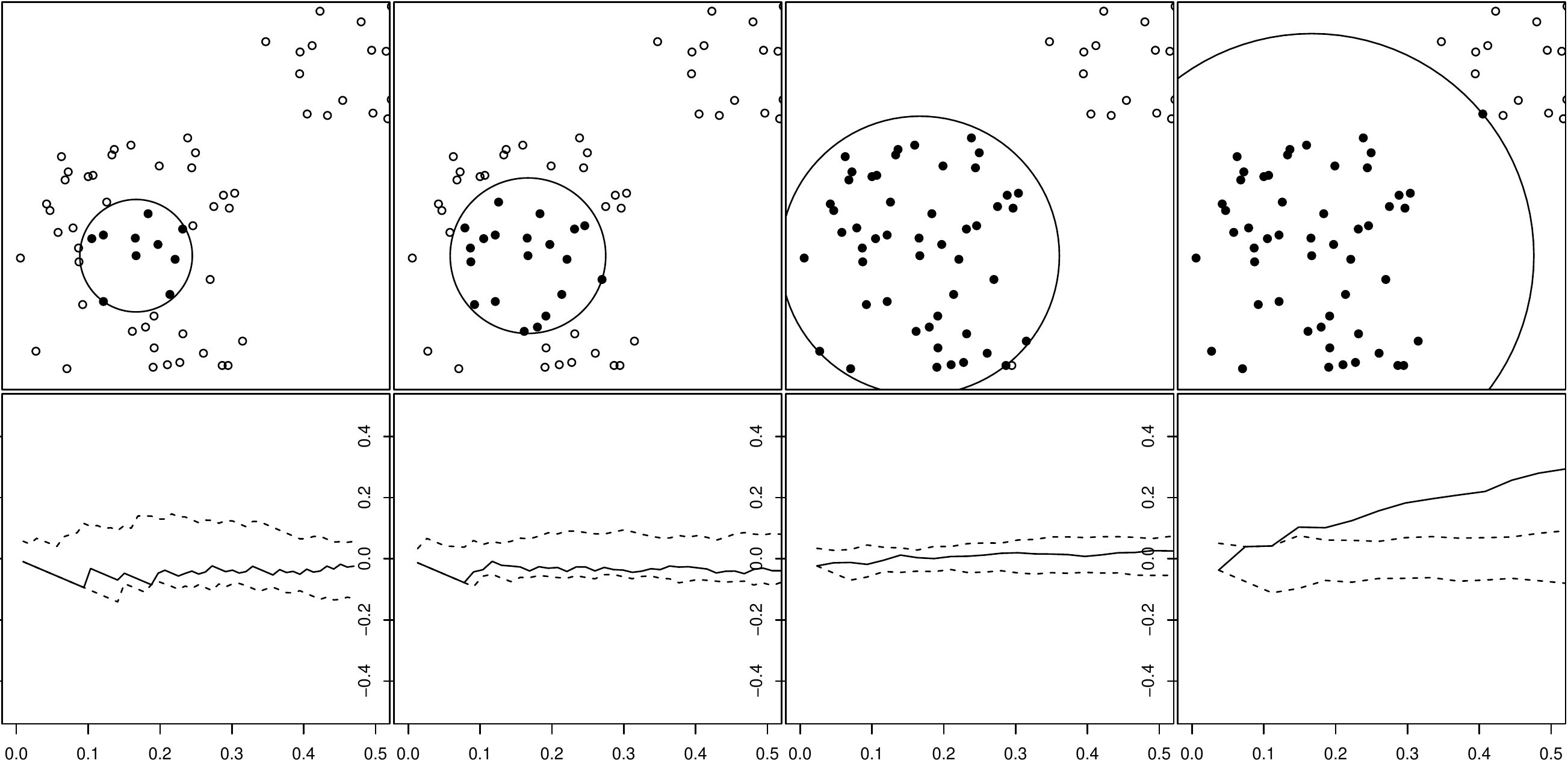}
\caption{Four snapshots (from left to right) of the growing balls. Solid points are those inside the covering ball (top). The envelopes of four snapshots plotting $\widehat{L}(t)-t$ (y-axis) against $t$ (x-axis) where solid lines are $\widehat{L}(t)-t$ curve of the data set and dashed lines are the upper and lower envelopes provided by the Monte Carlo Simulations (bottom).}
\label{fig:CCDseq}
\end{figure}


Algorithm~\ref{alg:ccdripley} of RK-CCDs is a parameter free version of the KS-CCD clustering algorithm given in Algorithm~\ref{alg:ccdks}; however, RK-CCDs have higher running time than KS-CCDs.

\begin{theorem} \label{thm:complexity2}
Let $\X \subset \R^d$ be a data set with $n=|\X|$ observations, and let $N$ be the number of Monte Carlo replications required
for the confidence envelopes of $\widehat{K}(t)$. Then, Algorithm~\ref{alg:ccdripley} partition the data set $\X \subset \R^d$ in $\mathcal{O}(n^3(\log{n}+N) + n^2d)$ time.
\end{theorem}	

\noindent {\bf Proof:}
Following Theorem~\ref{thm:complexity}, the only difference between Algorithm~\ref{alg:ccdks} and Algorithm~\ref{alg:ccdripley} is how $r(x)$ is computed for each point $x \in \X$. We first calculate the upper envelope before the main loop and that takes at most $\mathcal{O}(n^3N)$ for $N$ being the number of simulated data sets. Without loss of generality, let $\Y$ be a one of these simulated data sets. We take a random subset of $Y$ with cardinality $m = |\Y| < n$ and we calculate a single upper envelope of $\widehat{K}(t)$. We repeat for all $m=1,2,\ldots,n$ which takes $\mathcal{O}(n^3)$ time. The distance matrix of $\X$ is found in $\mathcal{O}(n^2d)$ time. For sorted distances from $x$ to all other points, we search for a window $B(x,r)$ that rejects the null hypothesis in a binary search fashion. Since the algorithm looks for the minimum $r$ that rejects the null hypothesis, we conduct a binary search on the sorted distances and pick the last $r$ that rejected the null hypothesis during the search. During the search, if the middle element rejects the null hypothesis for radius $r=r_0$, we repeat the search for the array of radii smaller than $r_0$, since all $r > r_0$ are assumed to reject the null hypothesis as well. Hence, the test is repeated only $\mathcal{O}(\log{n})$ times. For all $r \in \mathbf{D}(x)$, we compute $\widehat{K}(t)$ of points inside window $B(x,r)$. Hence, we compute $r(X)$ in $\mathcal{O}(n^2 \log{n})$ time, since $\widehat{K}(t)$ is calculated in $\mathcal{O}(n^2)$ time for each $r \in \mathbf{D}(x)$. Therefore, it takes $\mathcal{O}(n^3\log{n})$ time to compute $\widehat{K}(t)$ of the entire data set. Similar to Algorithm~\ref{alg:ccdks}, $S_{MD}$ and $S(G_{MD})$ is computed in $\mathcal{O}(n^3)$ in the worst case. $\blacksquare$

Algorithm~\ref{alg:ccdripley} works well for data sets with convex clusters. MDSs are equivalent to the possible cluster centers whose balls have considerably high number of points. On the other hand, arbitrarily shaped clusters are not characterized, by their centers, but by their shapes. Hence, we have to revise Algorithm~\ref{alg:ccdripley} slightly to locate the arbitrarily shaped clusters. Instead of finding the (approximate) MDS, $S(G_{MD})$, of the intersection graph $G_{MD}$, we search for connected components of $G_{MD}$ that are disconnected (i.e. vertices of different components are disconnected).


\section{Monte Carlo Simulations} \label{sec:montecarlo}

We conduct a series of Monte Carlo experiments with 100 replicates to assess the performance of RK-CCDs, and other clustering algorithms;  KS-CCDs, Affinity Propagation (AP) for $q=0.5$ and $q=0.0$, fuzzy $c$-means (FCmeans), and pdfCluster (pdfC); on data sets with convex clusters. In each trial, we record the relative frequency of finding the correct number of the clusters for each method, and we also record the rand index (RI) \citep{gan2007}. We first investigate the performance of clustering methods on three simulation settings wherein we partition data sets whose clusters have (i) fixed centers or (ii) random cluster centers (generated by a Strauss process) without noise, or (iii) random cluster centers with noise. For each simulation setting, we consider $\K=2,3,5$ as the number of clusters, $n=30,50,100,200$ as the number of observations for each cluster, and $d=2,3,5$ as the dimensionality of the data set.

RK-CCDs are parameter free clustering methods, and hence do not require any parameters to be specified. On the other hand, FCmeans methods require specifying the desired number of clusters $\widehat{\K}$, and KS-CCD and pdfC methods require specifying the intensity parameters $\delta$ and $h$, respectively. We incorporate a set of parameters for each method and report the parameters that perform the best. We report on the optimum parameters of KS-CCD and pdfC methods to investigate the relation between the intensity parameters and success rates of these methods. We take $k=2,3,\ldots,10$ for FCmeans and $h=0.50,0.55,0.60,\ldots,4$ for pdfC. However, we choose different sets of values of $\delta$ for each of the three simulation settings as will be described below. Moreover, only for FCmeans, we use the silhouette measure to choose a number of clusters that attain the best partitioning where we select the value $\widehat{\K}$ that achieves the maximum silhouette. AP algorithm requires specifying the input preferences, which are simplified by the parameter $q$, setting input preferences to the $q$'th quantile of the set of similarity values between points of the data set. Since each input preference is equal to the $q$'th quantile, each point is equally likely to be a candidate for being a cluster center. We run AP for both $q=0.5$ (the default value) and $q=0$ using the \emph{apcluster} package in R \citep{R2008}. We show that both $q=0$ and $q=0.5$ produce the true number of clusters in different cases.

We start with a simulation setting where clusters have fixed centers. These centers $\{\mu_k\}_{k=1,2,\ldots,\K}$, corresponding to each simulation setting, are given in Table~\ref{fig:CCDcenter}. We conduct two separate experiments where in first, members (or points) of each cluster are drawn uniformly from a box $[\mu_k-1,\mu_k+1]^d \subset \R^d$, and in second, points of each cluster are multivariate normally distributed in $\R^d$ with mean vector $\mu_k$ and covariance matrix $I_d$ which is an identity matrix of dimension $d=2,3,5$. We illustrate two realizations of this simulation setting in Figure~\ref{fig:ccdripley_fixed}.

\begin{table}[ht]
\centering
\caption{Centers of the clusters used in the first simulation setting.}
\resizebox{\textwidth}{!}{
\begin{tabular}{cccc|ccc}
& \multicolumn{3}{c}{Uniform} & \multicolumn{3}{c}{Normal}\\
\hline
& $d=2$ & $d=3$ & $d=5$ & $d=2$ & $d=3$ & $d=5$\\
\hline
\multirow{2}{*}{$\K=2$}  &  $\mu_1=(0,0)$  &  $\mu_1=(0,0,0)$ &  $\mu_1=(0,0,0,0,0)$  &  $\mu_1=(0,0)$  &  $\mu_1=(0,0,0)$ &  $\mu_1=(0,0,0,0,0)$  \\
                            &  $\mu_2=(3,0)$  &  $\mu_2=(3,0,0)$ &  $\mu_2=(3,0,0,0,0)$    &  $\mu_2=(5,0)$  &  $\mu_2=(5,0,0)$ &  $\mu_2=(5,0,0,0,0)$   \\
\hline
\multirow{3}{*}{$\K=3$}  &  $\mu_1=(0,0)$  &  $\mu_1=(0,0,0)$  &  $\mu_1=(0,0,0,0,0)$   &  $\mu_1=(0,0)$  &  $\mu_1=(0,0,0)$  &  $\mu_1=(0,0,0,0,0)$  \\
                            &  $\mu_2=(3,0)$  &  $\mu_2=(3,0,0)$  &  $\mu_2=(3,0,0,0,0)$       &  $\mu_2=(5,0)$  &  $\mu_2=(5,0,0)$  &  $\mu_2=(5,0,0,0,0)$  \\
                            &  $\mu_3=(1.5,2)$  &  $\mu_3=(1.5,2,2)$  &  $\mu_3=(1.5,2,2,0,0)$    &  $\mu_3=(2,4)$  &  $\mu_3=(2,3,3)$  &  $\mu_3=(2,3,3,0,0)$ \\
\hline
\multirow{5}{*}{$\K=5$}  &  $\mu_1=(0,0)$  &  $\mu_1=(0,0,0)$  &  $\mu_1=(0,0,0,0,0)$  &  $\mu_1=(0,0)$  &  $\mu_1=(0,0,0)$  &  $\mu_1=(0,0,0,0,0)$  \\
                            &  $\mu_2=(3,0)$  &  $\mu_2=(3,0,0)$  &  $\mu_2=(3,0,0,0,0)$   &  $\mu_2=(5,0)$  &  $\mu_2=(5,0,0)$  &  $\mu_2=(5,0,0,0,0)$  \\
                            &  $\mu_3=(0,3)$  &  $\mu_3=(0,3,0)$  &  $\mu_3=(0,3,0,0,0)$   &  $\mu_3=(0,5)$  &  $\mu_3=(0,5,0)$  &  $\mu_3=(0,5,0,0,0)$  \\
                            &  $\mu_4=(3,3)$  &  $\mu_4=(3,3,0)$  &  $\mu_4=(3,3,0,0,0)$   &  $\mu_4=(5,5)$  &  $\mu_4=(5,5,0)$  &  $\mu_4=(5,5,0,0,0)$  \\
                            &  $\mu_5=(1.5,6)$  &  $\mu_5=(1.5,1.5,3)$  &  $\mu_5=(1.5,1.5,3,0,0)$ & $\mu_5=(2.5,10)$  &  $\mu_5=(2.5,2.5,5)$  &  $\mu_5=(2.5,2.5,5,0,0)$ \\
\hline
\end{tabular}
}
\label{fig:CCDcenter}
\end{table}

In Table~\ref{fig:CCDauc} and Table~\ref{fig:CCDauc_auc}, we provide the relative frequency of success and the average RI measures (of 100 Monte Carlo replicates) on simulated data sets whose centers of clusters are given in Table~\ref{fig:CCDcenter}. We choose $\delta=0.01,0.02,0.03,\ldots,4$ as the set of intensity parameters for KS-CCDs. For uniformly distributed clusters in $d=2,3$, the RK-CCDs estimates the true partitioning with 100 \% success rate and 1.00 RI. For $d=5$, however, the lowest success rate is observed in the case where $(\K,n)=(5,30)$ which is 88 \%. With normally distributed settings, RK-CCDs perform better for clusters with higher number of observations. Even though the covering balls test for CSR, RK-CCDs estimates the locations of all clusters with a slight error where success rate goes up to 97 \% for some normally distributed settings with $n=200$. As a result of the high success rate, RIs of RK-CCDs are mostly higher than 0.90 in all settings. However, RI measures are still higher for some cases with low success rate. The reason for high RI is that; for example, in a simulation setting with $n=30$, RK-CCDs confuse two clusters  as a single one. This results in a failure of detecting the true number of clusters, although a considerable portion of the data set is well partitioned. However, the main reason for the drop in the relative frequency of success in cases with $n=30$ is the low number of observations in each cluster with moderately high number of dimensions which manifest low spatial intensity in $\R^d$. When $n=30$, the average inter-cluster distance is near the average intra-cluster distance which makes it harder to detect existing clusters, and it results in a deterioration of the performance of RK-CCDs.

\begin{figure}[!h]
\centering
\begin{tabular}{cc}
\includegraphics[scale=0.3]{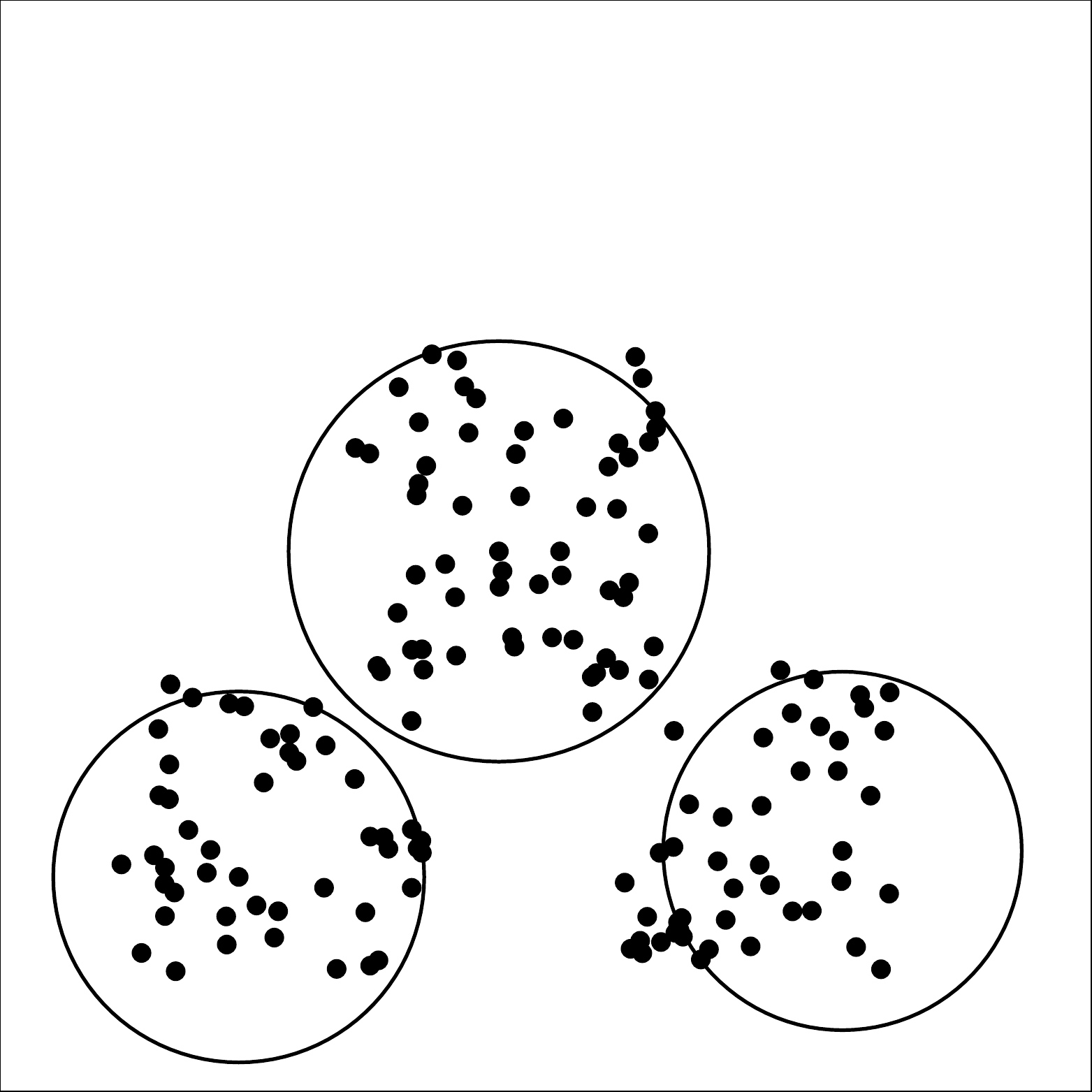} & \includegraphics[scale=0.3]{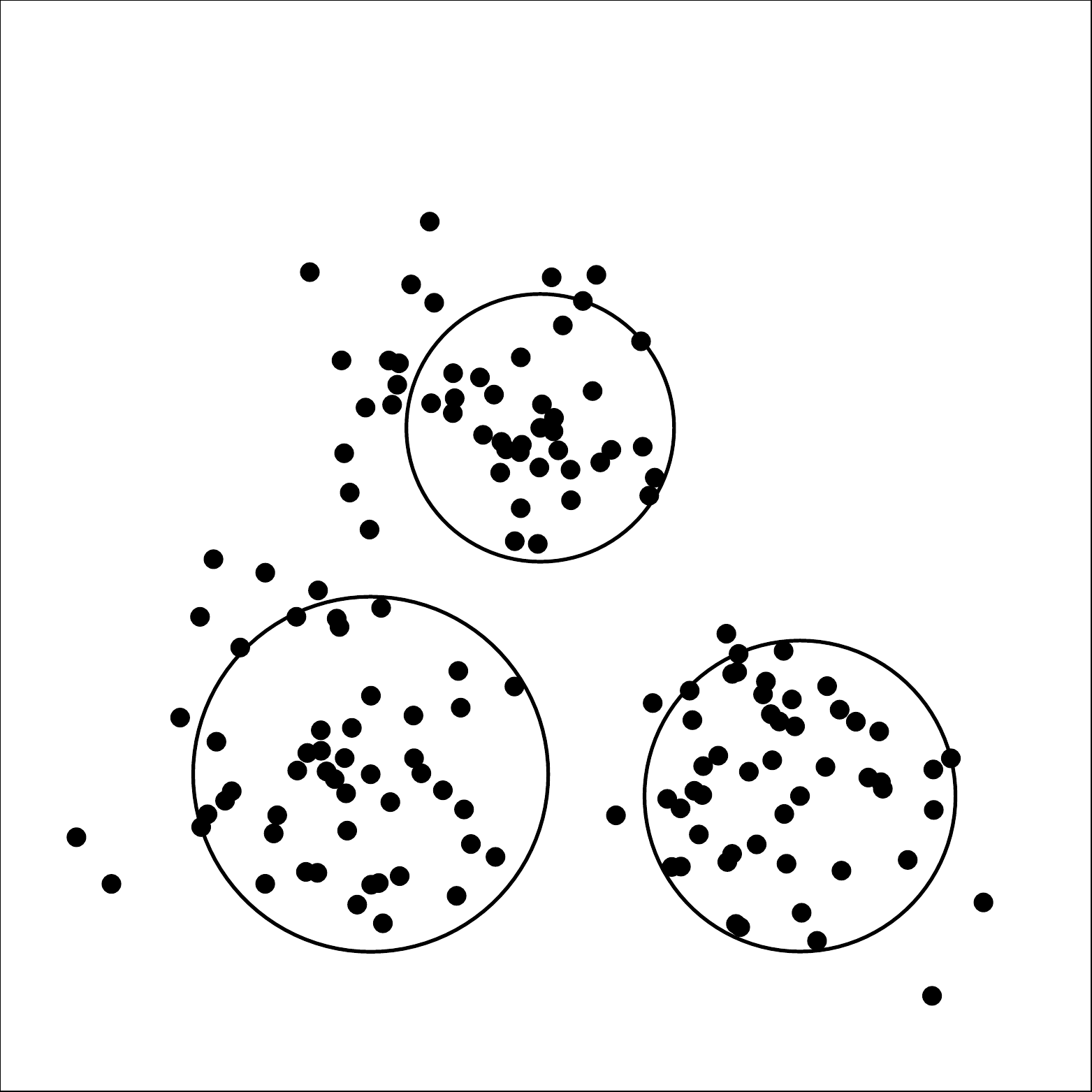} \\
(a) & (b) \\
\end{tabular}
\caption{Two depictions of RK-CCDs on the first simulation study setting with three clusters, and the covering balls of the elements of $S(G_{MD})$ provided by Algorithm~\ref{alg:ccdripley}. Each covering ball represents a cluster.
(a) Three bivariate uniformly distributed clusters $F_1=U([-1,1]^2)$, $F_2=U([2.5,3.5] \times [-1,1])$ and $F_3=U([1,2] \times [1.5,2.5])$. (b) Three bivariate normally distributed clusters $F_1=N(\mu_1=(0,0),I_2)$, $F_2=N(\mu_2=(5,0),I_2)$ and $F_3=N(\mu_3=(2,4),I_2)$ where $I_2$ is the identity matrix of size 2.}
\label{fig:ccdripley_fixed}
\end{figure}

In Table~\ref{fig:CCDauc} and Table~\ref{fig:CCDauc_auc}, observe that KS-CCDs perform better than or comparable to RK-CCDs except for some simulated settings with high $d$ and $n$. Similar to RK-CCDs, covering balls of KS-CCDs test whether points are randomly fall inside, hence the parameter $\delta$ plays an important role. However, the optimum value of $\delta$ parameter is usually around $0.05$ and $0.10$. KS-CCDs perform the worst in normally simulated settings with $d=5$, and with increasing number of $\K$, the relative success rate gets below \%10 in some cases. However, the high RI indicates that KS-CCDs still correctly partition some of the true clusters. FCmeans, on the other hand, achieve perfect success rates in all settings including uniformly and normally distributed ones. Here, the silhouette measure successfully decides which number of cluster is best to partition the data set. However, AP algorithm with $q=0.5$ performs the worst where the algorithm only achieves success in settings with low number of observations which gets even worse with increasing dimensionality. With $q=0$, on the other hand, the AP algorithm performs considerably better with exceptions in uniformly distributed settings of $n=200$. It appears that the lower the $q$ the better the AP algorithm for increasing number of observations where performance degrades with increasing dimensionality. For simulation settings with normally and uniformly distributed clusters, pdfC performs well except for $d=5$. \cite{Azzalini2014Clustering} have also indicated that fixing the bandwidth parameter of pdfC method may cause issues for high dimensional data sets. For settings with small $\K$ and $n$ however, the success rate of pdfC degrades slightly. Similar to KS-CCD, pdfC depends on a spatial intensity (or bandwidth) parameter $h$ whose choice is critical for capturing the true number of clusters. There seems to be no difference in the optimum values of $h$ between settings with uniformly and normally distributed clusters, but the optimum $h$ slightly decreases in $\K$ and $d$. Obviously, with increasing number of clusters and the dimensionality, it is harder to estimate the density of each cluster, and hence, it is preferable to choose lower values of bandwidth $h$ to distinguish clusters. But we observe an increase in the optimum value of $h$ with increasing number of observations since the clusters get denser for' increasing $n$.

RK-CCDs and FCmeans perform relatively well while the performance of KS-CCDs and pdfC degrade with increasing dimensionality. For $d=5$ and $\K=3,5$, since the average inter-cluster distance gets near the average intra-cluster distance, we observe a decrease in the success rate. AP algorithm, on the other hand, behaves oppositely where the performance of AP($q=0$) degrades for increasing $n$, but the success rate deteriorates slower for increasing number of clusters. In our experiments, increasing number of clusters decreases the intra-cluster distances which makes the observations of different clusters get closer. Other clustering methods show slightly better results than RK-CCDs in $d=2,3$ but these methods require a prior specification of parameters to achieve such high success rates. Nevertheless, RK-CCDs show relatively good performance despite the lack of any pre-specified parameters. In settings with higher dimensions and higher true number of clusters, the success rate of RK-CCD decreases slightly but it still shows comparable performance. When clusters being relatively close to each other, it is indeed harder to detect existing clusters \citep{tibshirani2001}.

\begin{table}
\centering
\caption{Success rates and optimum parameters of all clustering methods where centers of clusters listed as in Table~\ref{fig:CCDcenter}.}
\resizebox{\textwidth}{!}{
     \begin{tabular}{|ccc|cccccc|cccccc|cccccc|}
     \hline \rule{0pt}{3ex}
&  \multicolumn{2}{c|}{Dist.}    & \multicolumn{6}{c|}{$d=2$}   & \multicolumn{6}{c|}{$d=3$}   & \multicolumn{6}{c|}{$d=5$}  \\
          \hline \rule{0pt}{3ex}
     &     \multirow{2}{*}{$K$} & \multirow{2}{*}{$N$} & \multirow{2}{*}{ RK-CCD  } & \multirow{2}{*}{KS-CCD ($\delta$)} & AP & AP & \multirow{2}{*}{ FCmeans } & \multirow{2}{*}{ pdfC ($h$) } & \multirow{2}{*}{ RK-CCD  } & \multirow{2}{*}{ KS-CCD ($\delta$)} & AP  & AP  & \multirow{2}{*}{ FCmeans } & \multirow{2}{*}{ pdfC ($h$)} & \multirow{2}{*}{ RK-CCD  } & \multirow{2}{*}{ KS-CCD ($\delta$)} & AP & AP & \multirow{2}{*}{ FCmeans } & \multirow{2}{*}{ pdfC ($h$)} \\
&&&&&($q=0.5$)&($q=0$)&&&&&($q=0.5$)&($q=0$)&&&&&($q=0.5$)&($q=0$)&&\\
\multirow{12}[0]{*}{Unif.}
&          2 &         30 &        100 & 100 (0.05) &         59 &        100 &        100 &  82 (1.40) &        100 & 100 (0.05) &         36 &        100 &        100 &  59 (1.05) &         96 &  96 (0.05) &          5 &        100 &        100 &  38 (0.50) \\

&          2 &         50 &        100 & 100 (0.05) &          0 &        100 &        100 &  84 (1.50) &        100 & 100 (0.05) &          1 &        100 &        100 &  55 (1.20) &        100 & 100 (0.05) &          0 &        100 &        100 &  58 (0.50) \\

&          2 &        100 &        100 & 100 (0.05) &          0 &        100 &        100 &  99 (1.75) &        100 & 100 (0.05) &          0 &         98 &        100 &  66 (1.45) &         97 & 100 (0.05) &          0 &         53 &        100 &  60 (0.95) \\

&          2 &        200 &        100 & 100 (0.05) &          0 &          0 &        100 & 100 (2.10) &        100 & 100 (0.05) &          0 &          0 &        100 &  88 (1.65) &        100 & 100 (0.05) &          0 &          2 &        100 &  44 (1.15) \\

&          3 &         30 &        100 & 100 (0.10) &         99 &        100 &        100 & 100 (0.60) &        100 & 100 (0.05) &         97 &        100 &        100 &  99 (0.80) &         92 &  69 (0.05) &         53 &        100 &        100 &  14 (0.70) \\

&          3 &         50 &        100 & 100 (0.10) &         41 &        100 &        100 & 100 (0.60) &        100 & 100 (0.05) &         39 &        100 &        100 & 100 (0.90) &         98 &  76 (0.05) &          2 &        100 &        100 &  41 (0.65) \\

&          3 &        100 &        100 & 100 (0.10) &          0 &        100 &        100 & 100 (0.65) &        100 & 100 (0.05) &          0 &        100 &        100 & 100 (0.85) &         93 &  84 (0.05) &          0 &         77 &        100 &  60 (1.05) \\

&          3 &        200 &        100 & 100 (0.10) &          0 &          0 &        100 & 100 (0.65) &        100 & 100 (0.05) &          0 &         23 &        100 & 100 (0.90) &        100 &  96 (0.05) &          0 &          1 &        100 &  74 (1.45) \\

&          5 &         30 &        100 & 100 (0.05) &        100 &        100 &         99 & 100 (0.65) &        100 & 100 (0.05) &        100 &        100 &        100 & 100 (0.60) &         88 &  21 (0.10) &         98 &        100 &        100 &   1 (0.50) \\

&          5 &         50 &        100 & 100 (0.05) &        100 &        100 &         99 & 100 (0.65) &        100 & 100 (0.05) &         96 &        100 &        100 & 100 (0.60) &         96 &  24 (0.05) &         42 &        100 &        100 &  10 (0.50) \\

&          5 &        100 &        100 & 100 (0.05) &          2 &        100 &        100 & 100 (0.65) &        100 & 100 (0.05) &          0 &        100 &        100 & 100 (0.60) &         91 &  26 (0.05) &          0 &        100 &        100 &  35 (0.65) \\

&          5 &        200 &        100 & 100 (0.05) &          0 &        100 &        100 & 100 (0.70) &        100 & 100 (0.05) &          0 &        100 &        100 & 100 (0.60) &         99 &  24 (0.05) &          0 &         35 &        100 &  48 (0.95) \\

\hline \rule{0pt}{3ex}
     &     \multirow{2}{*}{$K$} & \multirow{2}{*}{$N$} & \multirow{2}{*}{ RK-CCD  } & \multirow{2}{*}{ KS-CCD ($\delta$)} & AP & AP & \multirow{2}{*}{ FCmeans } & \multirow{2}{*}{ pdfC ($h$) } & \multirow{2}{*}{ RK-CCD  } & \multirow{2}{*}{ KS-CCD ($\delta$)} & AP  & AP  & \multirow{2}{*}{ FCmeans } & \multirow{2}{*}{ pdfC ($h$)} & \multirow{2}{*}{ RK-CCD  } & \multirow{2}{*}{ KS-CCD ($\delta$)} & AP & AP & \multirow{2}{*}{ FCmeans } & \multirow{2}{*}{ pdfC ($h$)} \\
&&&&&($q=0.5$)&($q=0$)&&&&&($q=0.5$)&($q=0$)&&&&&($q=0.5$)&($q=0$)&&\\
\multirow{12}[0]{*}{Normal}
  &          2 &         30 &         86 &  99 (0.10) &          0 &        100 &        100 &  94 (1.15) &         98 & 100 (0.05) &          0 &        100 &        100 &  65 (0.85) &         91 &  62 (0.05) &          0 &        100 &        100 &  24 (0.50) \\

 &          2 &         50 &        100 & 100 (0.10) &          0 &        100 &        100 &  99 (1.25) &        100 & 100 (0.05) &          0 &        100 &        100 &  76 (1.05) &        100 &  60 (0.05) &          0 &         98 &        100 &  48 (0.60) \\

 &          2 &        100 &        100 & 100 (0.10) &          0 &         87 &        100 & 100 (1.35) &        100 & 100 (0.05) &          0 &         82 &        100 &  99 (1.20) &         99 &  62 (0.10) &          0 &         25 &        100 &  56 (0.55) \\

 &          2 &        200 &        100 & 100 (0.10) &          0 &          1 &        100 & 100 (1.30) &         99 & 100 (0.05) &          0 &          3 &        100 &  99 (1.20) &        100 &  61 (0.05) &          0 &          0 &        100 &  58 (0.85) \\

 &          3 &         30 &         42 &  98 (0.10) &          0 &        100 &        100 & 100 (0.80) &         70 &  94 (0.05) &          0 &        100 &        100 &  97 (0.80) &         71 &  12 (0.05) &          0 &        100 &        100 &  11 (0.70) \\

 &          3 &         50 &         74 & 100 (0.10) &          0 &        100 &        100 & 100 (0.75) &         85 &  94 (0.05) &          0 &        100 &        100 & 100 (0.95) &         92 &  30 (0.05) &          0 &         97 &        100 &  39 (0.75) \\

 &          3 &        100 &         87 & 100 (0.10) &          0 &         89 &        100 & 100 (0.75) &         93 & 100 (0.05) &          0 &         86 &        100 & 100 (0.90) &         96 &  27 (0.10) &          0 &         36 &        100 &  62 (0.90) \\

 &          3 &        200 &         94 & 100 (0.10) &          0 &          2 &        100 & 100 (0.70) &         98 & 100 (0.05) &          0 &          1 &        100 & 100 (0.75) &         94 &  33 (0.15) &          0 &          1 &        100 &  87 (0.95) \\

 &          5 &         30 &         61 &  98 (0.05) &         25 &        100 &        100 & 100 (0.65) &         68 &  28 (0.05) &         11 &        100 &        100 & 100 (1.00) &         64 &   5 (0.05) &          0 &        100 &        100 &   2 (0.50) \\

 &          5 &         50 &         84 & 100 (0.05) &          0 &        100 &        100 & 100 (0.70) &         90 &  36 (0.05) &          0 &        100 &        100 & 100 (0.75) &         85 &   7 (0.05) &          0 &        100 &        100 &  10 (0.50) \\

 &          5 &        100 &         96 & 100 (0.05) &          0 &        100 &        100 & 100 (0.70) &         95 &  53 (0.05) &          0 &        100 &        100 & 100 (0.75) &         84 &   9 (0.10) &          0 &         87 &        100 &  34 (0.55) \\

 &          5 &        200 &        100 & 100 (0.05) &          0 &        100 &        100 & 100 (0.70) &         97 &  58 (0.05) &          0 &         44 &        100 & 100 (0.70) &         86 &  10 (0.15) &          0 &          2 &        100 &  82 (0.90) \\
          \hline
    \end{tabular}
}
\label{fig:CCDauc}
\end{table}

\begin{table}[h]
\centering
\caption{Rand Index (RI) measures of all clustering methods given in Table~\ref{fig:CCDauc}.}
\resizebox{\textwidth}{!}{
     \begin{tabular}{|ccc|cccccc|cccccc|cccccc|}
          \hline \rule{0pt}{3ex}
&  \multicolumn{2}{c|}{Dist.}    & \multicolumn{6}{c|}{$d=2$}   & \multicolumn{6}{c|}{$d=3$}   & \multicolumn{6}{c|}{$d=5$}  \\
          \hline \rule{0pt}{3ex}
     &     \multirow{2}{*}{$K$} & \multirow{2}{*}{$N$} & \multirow{2}{*}{ RK-CCD  } & \multirow{2}{*}{ KS-CCD   } & AP & AP & \multirow{2}{*}{ FCmeans } & \multirow{2}{*}{ pdfC  } & \multirow{2}{*}{ RK-CCD  } & \multirow{2}{*}{ KS-CCD  } & AP  & AP  & \multirow{2}{*}{ FCmeans } & \multirow{2}{*}{ pdfC  } & \multirow{2}{*}{ RK-CCD  } & \multirow{2}{*}{ KS-CCD  } & AP & AP & \multirow{2}{*}{ FCmeans } & \multirow{2}{*}{ pdfC } \\
&&&&&($q=0.5$)&($q=0$)&&&&&($q=0.5$)&($q=0$)&&&&&($q=0.5$)&($q=0$)&&\\
 \multirow{12}[0]{*}{Unif}
           &          2 &         30 &      1.000 &      1.000 &      0.590 &      1.000 &      1.000 &      0.901 &      1.000 &      1.000 &      0.360 &      1.000 &      1.000 &      0.811 &      0.985 &      0.999 &      0.050 &      1.000 &      1.000 &      0.573 \\

           &          2 &         50 &      1.000 &      1.000 &      0.000 &      1.000 &      1.000 &      0.968 &      1.000 &      1.000 &      0.010 &      1.000 &      1.000 &      0.830 &      1.000 &      1.000 &      0.000 &      1.000 &      1.000 &      0.610 \\

           &          2 &        100 &      1.000 &      1.000 &      0.000 &      1.000 &      1.000 &      0.999 &      1.000 &      1.000 &      0.000 &      0.980 &      1.000 &      0.876 &      1.000 &      1.000 &      0.000 &      0.530 &      1.000 &      0.717 \\

           &          2 &        200 &      1.000 &      1.000 &      0.000 &      0.000 &      1.000 &      0.995 &      1.000 &      1.000 &      0.000 &      0.000 &      1.000 &      0.967 &      1.000 &      1.000 &      0.000 &      0.020 &      1.000 &      0.658 \\

           &          3 &         30 &      1.000 &      1.000 &      0.990 &      1.000 &      1.000 &      1.000 &      1.000 &      1.000 &      0.970 &      1.000 &      1.000 &      0.998 &      0.987 &      0.978 &      0.530 &      1.000 &      1.000 &      0.566 \\

           &          3 &         50 &      1.000 &      1.000 &      0.410 &      1.000 &      1.000 &      1.000 &      1.000 &      1.000 &      0.390 &      1.000 &      1.000 &      1.000 &      1.000 &      0.983 &      0.020 &      1.000 &      1.000 &      0.730 \\

           &          3 &        100 &      1.000 &      1.000 &      0.000 &      1.000 &      1.000 &      1.000 &      1.000 &      1.000 &      0.000 &      1.000 &      1.000 &      1.000 &      0.998 &      0.985 &      0.000 &      0.770 &      1.000 &      0.894 \\

           &          3 &        200 &      1.000 &      1.000 &      0.000 &      0.000 &      1.000 &      1.000 &      1.000 &      1.000 &      0.000 &      0.230 &      1.000 &      1.000 &      1.000 &      1.000 &      0.000 &      0.010 &      1.000 &      0.949 \\

           &          5 &         30 &      1.000 &      1.000 &      1.000 &      1.000 &      1.000 &      1.000 &      1.000 &      1.000 &      1.000 &      1.000 &      1.000 &      1.000 &      0.990 &      0.888 &      0.980 &      1.000 &      1.000 &      0.510 \\

           &          5 &         50 &      1.000 &      1.000 &      1.000 &      1.000 &      1.000 &      1.000 &      1.000 &      1.000 &      0.960 &      1.000 &      1.000 &      1.000 &      1.000 &      0.935 &      0.420 &      1.000 &      1.000 &      0.650 \\

           &          5 &        100 &      1.000 &      1.000 &      0.020 &      1.000 &      1.000 &      1.000 &      1.000 &      1.000 &      0.000 &      1.000 &      1.000 &      1.000 &      1.000 &      0.947 &      0.000 &      1.000 &      1.000 &      0.839 \\

           &          5 &        200 &      1.000 &      1.000 &      0.000 &      1.000 &      1.000 &      1.000 &      1.000 &      1.000 &      0.000 &      1.000 &      1.000 &      1.000 &      1.000 &      0.943 &      0.000 &      0.350 &      1.000 &            0.950 \\
\hline
     &     \multirow{2}{*}{$K$} & \multirow{2}{*}{$N$} & \multirow{2}{*}{ RK-CCD  } & \multirow{2}{*}{ KS-CCD   } & AP & AP & \multirow{2}{*}{ FCmeans } & \multirow{2}{*}{ pdfC  } & \multirow{2}{*}{ RK-CCD  } & \multirow{2}{*}{ KS-CCD  } & AP  & AP  & \multirow{2}{*}{ FCmeans } & \multirow{2}{*}{ pdfC  } & \multirow{2}{*}{ RK-CCD  } & \multirow{2}{*}{ KS-CCD  } & AP & AP & \multirow{2}{*}{ FCmeans } & \multirow{2}{*}{ pdfC } \\
&&&&&($q=0.5$)&($q=0$)&&&&&($q=0.5$)&($q=0$)&&&&&($q=0.5$)&($q=0$)&&\\
\multirow{12}[0]{*}{Normal}
           &          2 &         30 &      0.893 &      0.970 &      0.000 &      0.987 &      0.987 &      0.940 &      0.940 &      0.972 &      0.000 &      0.985 &      0.984 &      0.815 &      0.918 &      0.701 &      0.000 &      0.983 &      0.988 &      0.534 \\

           &          2 &         50 &      0.975 &      0.979 &      0.000 &      0.990 &      0.983 &      0.980 &      0.962 &      0.975 &      0.000 &      0.987 &      0.988 &      0.851 &      0.948 &      0.788 &      0.000 &      0.965 &      0.985 &      0.599 \\

           &          2 &        100 &      0.981 &      0.983 &      0.000 &      0.863 &      0.987 &      0.987 &      0.966 &      0.983 &      0.000 &      0.813 &      0.986 &      0.948 &      0.956 &      0.786 &      0.000 &      0.248 &      0.987 &      0.627 \\

           &          2 &        200 &      0.979 &      0.985 &      0.000 &      0.010 &      0.989 &      0.987 &      0.975 &      0.983 &      0.000 &      0.030 &      0.988 &      0.978 &      0.947 &      0.888 &      0.000 &      0.000 &      0.986 &      0.716 \\

           &          3 &         30 &      0.694 &      0.964 &      0.000 &      0.975 &      0.980 &      0.974 &      0.847 &      0.944 &      0.000 &      0.976 &      0.982 &      0.970 &      0.839 &      0.537 &      0.000 &      0.972 &      0.980 &      0.544 \\

           &          3 &         50 &      0.845 &      0.966 &      0.000 &      0.981 &      0.979 &      0.974 &      0.892 &      0.955 &      0.000 &      0.980 &      0.982 &      0.976 &      0.912 &      0.694 &      0.000 &      0.941 &      0.982 &      0.721 \\

           &          3 &        100 &      0.903 &      0.971 &      0.000 &      0.877 &      0.980 &      0.975 &      0.926 &      0.968 &      0.000 &      0.848 &      0.983 &      0.979 &      0.927 &      0.655 &      0.000 &      0.353 &      0.982 &      0.863 \\

           &          3 &        200 &      0.949 &      0.974 &      0.000 &      0.020 &      0.980 &      0.978 &      0.950 &      0.976 &      0.000 &      0.010 &      0.982 &      0.980 &      0.927 &      0.765 &      0.000 &      0.010 &      0.983 &      0.948 \\

           &          5 &         30 &      0.890 &      0.982 &      0.248 &      0.989 &      0.990 &      0.989 &      0.917 &      0.910 &      0.109 &      0.987 &      0.989 &      0.988 &      0.920 &      0.479 &      0.000 &      0.981 &      0.990 &      0.495 \\

           &          5 &         50 &      0.953 &      0.984 &      0.000 &      0.989 &      0.990 &      0.988 &      0.964 &      0.925 &      0.000 &      0.988 &      0.990 &      0.989 &      0.957 &      0.610 &      0.000 &      0.988 &      0.991 &      0.608 \\

           &          5 &        100 &      0.977 &      0.986 &      0.000 &      0.992 &      0.990 &      0.989 &      0.971 &      0.938 &      0.000 &      0.992 &      0.991 &      0.989 &      0.950 &      0.645 &      0.000 &      0.861 &      0.991 &      0.812 \\

           &          5 &        200 &      0.985 &      0.988 &      0.000 &      0.994 &      0.990 &      0.990 &      0.975 &      0.984 &      0.000 &      0.437 &      0.991 &      0.990 &      0.966 &      0.696 &      0.000 &      0.020 &      0.991 &      0.969 \\
\hline
\end{tabular}
}
\label{fig:CCDauc_auc}
\end{table}

The performance of a clustering method depends on both the intra-cluster and inter-cluster distances of several clusters in a data set. In some cases, although the data set is sampled from a mixture distribution, the samples from two or more clusters may be so close in proximity that two or more clusters are confused as a single cluster. Therefore, we aim to investigate the performance of RK-CCDs when clusters are forced to be slightly closer to each other. In this second experiment, we simulate data sets with centers of clusters drawn from a Strauss Process. We generate random cluster centers that are at least $t \in \R_+$ apart from each other (by fixing $\beta=0$ in the Strauss Process). Let $\mathbf{M} = \{M_1,M_2,\ldots,M_{\K}\} \subset U([0,1]^d) \subset \R^d$ be a random point pattern drawn from a Strauss Process with $\beta=0$. We have two separate types of settings in this experiment. In first, members of each cluster are drawn uniformly from a box $[M_k-r,M_k+r]^d \subset \R^d$ where $2r$ being the range of the uniform distribution in all dimensions and $t=(2+\theta)r$. In second, points of each cluster are multivariate normally distributed in $\R^d$ with mean vector $M_k$ and covariance $\Sigma=I_{d \times d}(r/5)$ where $I_{d \times d}(\sigma^2)$ is a diagonal covariance matrix with fixed variance $\sigma^2$ for $d=2,3,5$.
For both settings, we fix $\theta=0.3$ and $r=0.15$ so as to simulate data sets with well separated clusters and with small intra-cluster distances. We illustrate two depictions of this simulation setting in Figure~\ref{fig:ccdripley_strauss}.

\begin{figure}[!h]
\centering
\begin{tabular}{cc}
\includegraphics[scale=0.3]{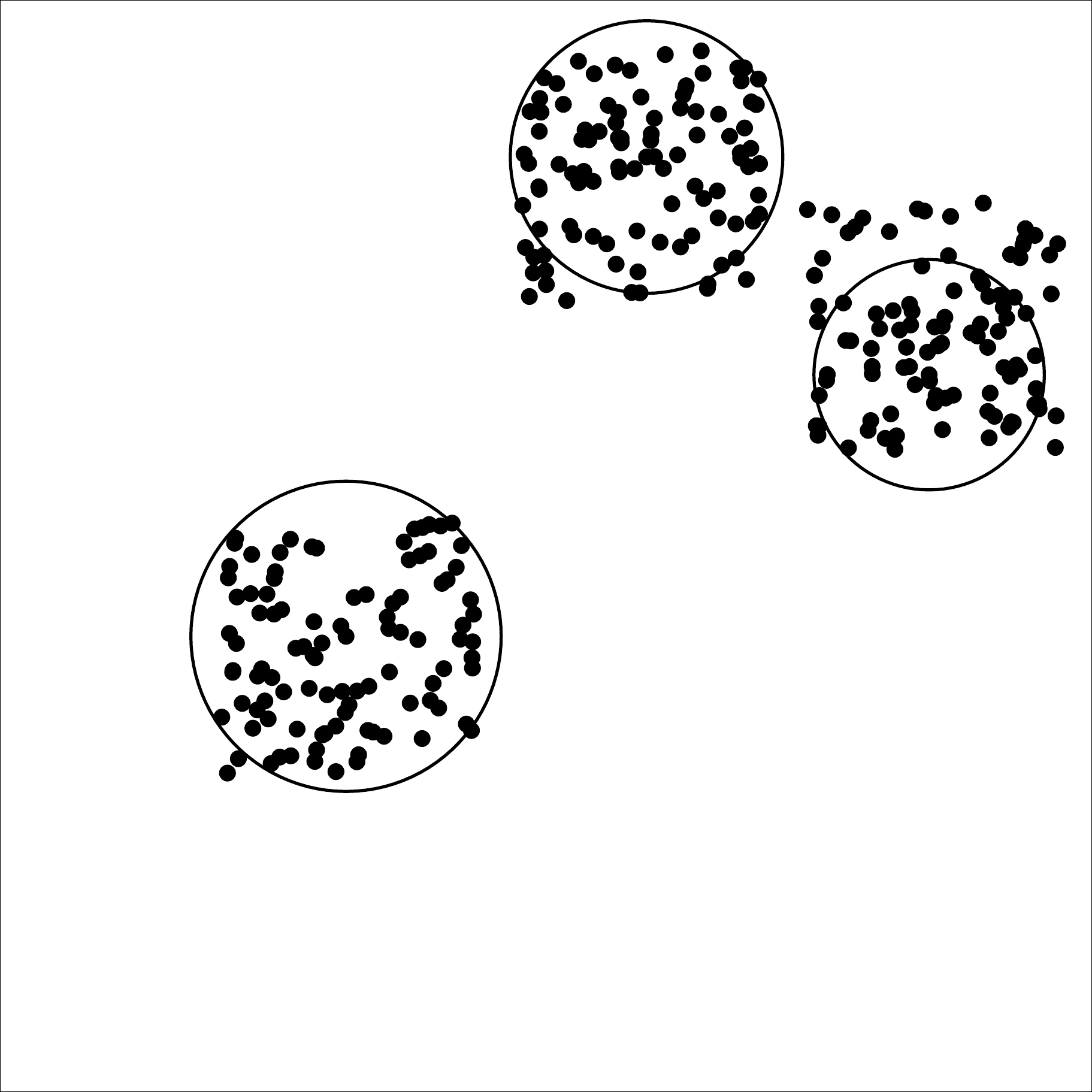} & \includegraphics[scale=0.3]{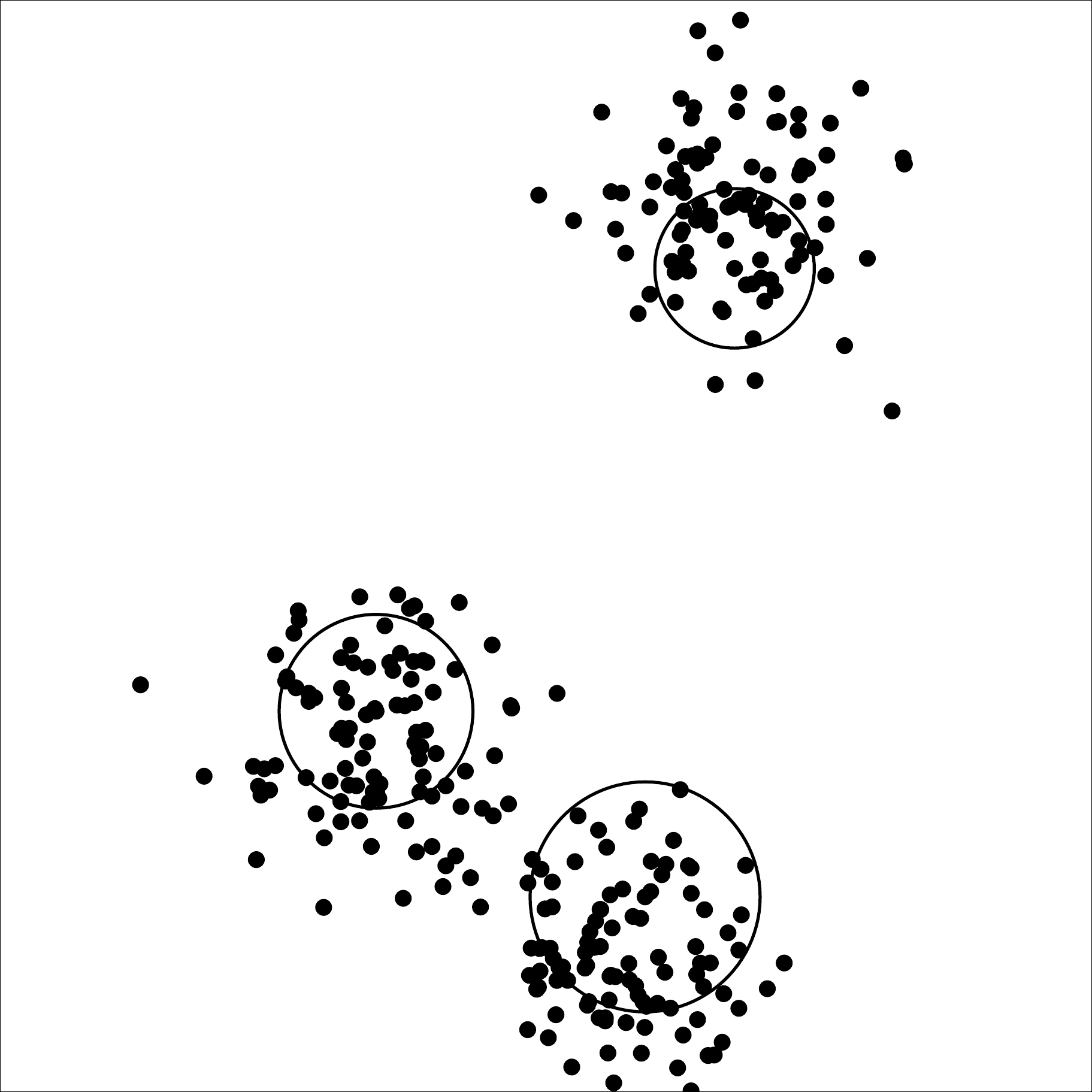} \\
(a) & (b) \\
\end{tabular}
\caption{Two depictions of RK-CCDs on the second simulation study setting with three clusters whose centers are drawn from a Strauss Process with $\beta=0$, and the covering balls of the elements of $S(G_{MD})$ provided by Algorithm~\ref{alg:ccdripley}. Each covering ball represents a cluster. (a) Bivariate uniformly distributed clusters (b) Bivariate normally distributed clusters.}
\label{fig:ccdripley_strauss}
\end{figure}


The relative frequencies of success and RI measures (of 100 Monte Carlo replicates) of the second experiment are given in Table~\ref{fig:CCDaucStrauss} and Table~\ref{fig:CCDaucstrauss_auc}. We choose $\delta=5.00,5.01,5.02,\ldots,15$ as the set of intensity parameters for KS-CCDs. Compared to the first simulation study (see Table~\ref{fig:CCDauc}), some clusters are now much closer to each other.
Hence, even in moderately high dimensions, RK-CCDs achieve slightly less success rate and RI; however, RK-CCDs show above 80\% success rate in finding the true number of clusters other than some cases with $n=30,50$. Although normally distributed clusters are detected much harder than uniformly distributed clusters (since covering balls are tested against null hypothesis of point patterns being CSR), RK-CCDs locate the true clusters with considerably high success rate. The performance of RK-CCDs drops,
if the number of observations is low and the dimensionality is high. The lower the $n$ and the higher the $d$, the closer the points of opposite clusters. Hence, it gets harder to distinguish two or more clusters from each other.
For the same reason, FCmeans does not successfully find the number of true clusters, and its performance is lower than or comparable to RK-CCDs in cases with low $n$ and high $d$. Although RK-CCDs perform slightly worse in partitioning data sets with normally distributed clusters, RK-CCDs are still comparable to all other clustering methods. KS-CCDs, on the other hand, seem to be better in $d=2$, but RK-CCDs outperform these digraphs in settings with higher $d$. The AP algorithm still performs better when $q=0$ and its success rate drops with increasing $n$.
The pdfC methods apparently show much higher success rates and RI than in first simulation studies given in Table~\ref{fig:CCDauc} but again its performance deteriorates with low $n$ and high $d$. Optimum parameters of pdfCs are similar to the ones from the first simulation study. Although, the optimum parameter of KS-CCDs, the optimum values of $\delta$, are drastically different than the ones in the first simulation study, there is not much of a change in the optimum $\delta$ for various values of $\K$, $d$ and $n$.

Compared to all other clustering methods, RK-CCDs show promising results. We test the point patterns in $\R^d$ inside the covering balls against CSR, but the method also works well with normally distributed clusters. With increasing number of observations, RK-CCDs achieve nearly 80\% success rate in finding the true number of normally distributed clusters for settings with $d=2$. The reason for that is, once a covering ball grows up to a point where it starts covering areas of lower density of the normally distributed cluster, it detects some locally homogeneous Poisson process, i.e. a region with less spatial intensity, which eventually make the test reject the null hypothesis. Then the greedy algorithm detects covering balls with high density which are more likely to be the centers of some normally distributed clusters. In summary, although RK-CCDs test against locally CSR regions to find possible clusters, it also works well for clusters with heterogeneous intensity which are unimodal and convex (clusters with points distributed around a center). Similar to the pdfC method, the performance of RK-CCDs decrease with lower $n$ and higher $d$. Once the clusters are close enough and the intensity of the data set is low, it is harder for density-based clustering methods to detect existing clusters.

\begin{table}[t]
\centering
\caption{Success rates and optimum parameters of all clustering methods where cluster centers are drawn from a Strauss process with $\beta=0$.}
\resizebox{\textwidth}{!}{
     \begin{tabular}{|ccc|cccccc|cccccc|cccccc|}
          \hline \rule{0pt}{3ex}
&  \multicolumn{2}{c|}{Dist.}    & \multicolumn{6}{c|}{$d=2$}   & \multicolumn{6}{c|}{$d=3$}   & \multicolumn{6}{c|}{$d=5$}  \\
          \hline \rule{0pt}{3ex}
     &     \multirow{2}{*}{$K$} & \multirow{2}{*}{$N$} & \multirow{2}{*}{ RK-CCD  } & \multirow{2}{*}{ KS-CCD ($\delta$)} & AP & AP & \multirow{2}{*}{ FCmeans } & \multirow{2}{*}{ pdfC ($h$) } & \multirow{2}{*}{ RK-CCD  } & \multirow{2}{*}{ KS-CCD ($\delta$)} & AP  & AP  & \multirow{2}{*}{ FCmeans } & \multirow{2}{*}{ pdfC ($h$)} & \multirow{2}{*}{ RK-CCD  } & \multirow{2}{*}{ KS-CCD ($\delta$)} & AP & AP & \multirow{2}{*}{ FCmeans } & \multirow{2}{*}{ pdfC ($h$)} \\
&&&&&($q=0.5$)&($q=0$)&&&&&($q=0.5$)&($q=0$)&&&&&($q=0.5$)&($q=0$)&&\\
\multirow{12}[0]{*}{Unif.}
 &          2 &         30 &         96 &         99 (6.69) &         21 &        100 &        100 &         94 (1.25) &        100 &         86 (6.98) &         42 &        100 &        100 &         88 (1.30) &         97 &         85 (6.36) &         54 &        100 &        100 &         68 (0.80)\\

 &          2 &         50 &        100 &         99 (6.47) &          8 &         97 &        100 &         98 (1.50) &        100 &         74 (6.93) &         21 &        100 &        100 &         95 (1.40) &         99 &         79 (6.14) &         18 &         98 &        100 &         86 (1.25) \\

 &          2 &        100 &         99 &         99 (6.62) &          1 &         61 &        100 &         99 (1.70) &        100 &         78 (6.79) &          2 &         83 &        100 &         98 (1.45) &        100 &         74 (6.35) &          2 &         85 &        100 &         93 (1.10)  \\

 &          2 &        200 &        100 &         99 (6.35) &          0 &         13 &        100 &        100 (1.85) &        100 &         84 (6.98) &          0 &         28 &        100 &       100 (1.85) &         99 &         76 (6.88) &          0 &         36 &        100 &        100 (1.60) \\

 &          3 &         30 &         78 &         89 (5.71) &         46 &        100 &         88 &         99 (0.95) &         88 &         73 (7.00) &         52 &        100 &         84 &        90 (0.80)&         93 &         57 (6.77) &         76 &        100 &         89 &         45 (0.75) \\

 &          3 &         50 &         83 &         90 (6.09) &         15 &        100 &         91 &         99 (0.85) &         89 &         73 (7.00) &         24 &        100 &         90 &        97 (1.10)&         91 &         64 (6.91) &         40 &        100 &         87 &         77 (0.85)  \\

 &          3 &        100 &         88 &         91 (5.88) &          0 &         97 &         88 &         98 (0.90) &         96 &         79 (7.00) &          2 &         99 &         85 &        99 (1.35) &         93 &         53 (6.99) &          3 &         96 &         90 &         98 (1.25)  \\

 &          3 &        200 &         92 &         90 (6.25) &          0 &         39 &         83 &         99 (1.20) &         87 &         79 (6.96) &          0 &         74 &         77 &        98 (1.40) &         90 &         50 (7.00)&          0 &         61 &         86 &        100 (1.40)  \\

 &          5 &         30 &         41 &         91 (5.25) &         59 &         87 &         92 &         92 (0.60) &         85 &         86 (6.98) &         83 &         98 &         85 &        87 (0.65) &         88 &         43 (6.95) &         86 &         99 &         88 &         19 (0.75)  \\

 &          5 &         50 &         74 &         92 (5.15) &         10 &        100 &         92 &         99 (0.80) &         88 &         80 (6.88) &         23 &        100 &         90 &        92 (0.75) &         84 &         25 (6.80) &         38 &        100 &         93 &         68 (0.60) \\

 &          5 &        100 &         91 &         96 (5.20) &          0 &        100 &         95 &        100 (0.80) &         96 &         85 (6.99) &          1 &        100 &         91 &       99 (0.85) &         91 &         30 (6.88) &          0 &        100 &         86 &         96 (0.90)  \\

 &          5 &        200 &         90 &         89 (5.05) &          0 &         91 &         92 &        100 (0.90) &         98 &         88 (6.97) &          0 &         97 &         85 &      100 (0.90) &         93 &         35 (6.90) &          0 &         85 &         83 &        100 (1.05)  \\
          \hline
     &     \multirow{2}{*}{$K$} & \multirow{2}{*}{$N$} & \multirow{2}{*}{ RK-CCD  } & \multirow{2}{*}{ KS-CCD ($\delta$)} & AP & AP & \multirow{2}{*}{ FCmeans } & \multirow{2}{*}{ pdfC ($h$) } & \multirow{2}{*}{ RK-CCD  } & \multirow{2}{*}{ KS-CCD ($\delta$)} & AP  & AP  & \multirow{2}{*}{ FCmeans } & \multirow{2}{*}{ pdfC ($h$)} & \multirow{2}{*}{ RK-CCD  } & \multirow{2}{*}{ KS-CCD ($\delta$)} & AP & AP & \multirow{2}{*}{ FCmeans } & \multirow{2}{*}{ pdfC ($h$)} \\
&&&&&($q=0.5$)&($q=0$)&&&&&($q=0.5$)&($q=0$)&&&&&($q=0.5$)&($q=0$)&&\\
\multirow{12}[0]{*}{Normal}
 &          2 &         30 &         95 &         99 (6.54) &         37 &        100 &        100 &         98 (1.00) &        100 &         77 (6.71) &         48 &        100 &        100 &         93 (1.15) &         90 &         66 (6.63) &         71 &        100 &        100 &         64 (0.85) \\

 &          2 &         50 &        100 &         99 (6.54) &         18 &        100 &        100 &         99 (1.00) &        100 &         65 (6.97) &         24 &        100 &        100 &         99 (1.10) &        100 &         81 (6.85) &         46 &        100 &        100 &         89 (0.75) \\

 &          2 &        100 &        100 &         99 (6.31) &          3 &         96 &        100 &        100 (1.20) &        100 &         76 (6.93) &          4 &         93 &        100 &      100 (1.05) &         97 &         79 (7.00) &          8 &         97 &        100 &         97 (1.00) \\

 &          2 &        200 &        100 &         99 (6.13) &          0 &         63 &        100 &        100 (1.15) &        100 &         68 (6.96) &          0 &         61 &        100 &      100 (1.00) &         99 &         68 (6.71) &          0 &         72 &        100 &         99 (0.85) \\

 &          3 &         30 &         76 &         86 (5.99) &         45 &         96 &         86 &         98 (0.75) &         85 &         58 (6.96) &         66 &         99 &         75 &        95 (0.90) &         73 &         55 (6.86) &         80 &        100 &         76 &         58 (0.95)  \\

 &          3 &         50 &         87 &         90 (6.24) &         17 &        100 &         83 &        100 (0.75) &         86 &         58 (6.96) &         24 &        100 &         74 &       99 (1.05) &         84 &         51 (6.99) &         57 &        100 &         74 &         85 (0.55)  \\

 &          3 &        100 &         87 &         95 (6.28) &          0 &        100 &         92 &        100 (0.90) &         85 &         52 (6.96) &          3 &        100 &         82 &      100 (1.15) &         81 &         49 (6.95) &          7 &        100 &         78 &         96 (0.90)\\

 &          3 &        200 &         81 &          84 (6.16) &          0 &         89 &         80 &        100 (0.70) &         85 &         59 (6.97) &          0 &         91 &         80 &      100 (1.10) &         88 &         55 (6.88) &          0 &         96 &         84 &        100 (0.75) \\

 &          5 &         30 &         42 &         94 (5.35) &         74 &         85 &         93 &         96 (0.65) &         68 &         63 (6.98)&         86 &         87 &         73 &         96 (0.65) &         67 &         27 (6.97) &         90 &         91 &         69 &         27 (0.65)  \\

 &          5 &         50 &         58 &         95 (5.42) &         22 &        100 &         92 &        100 (0.65) &         81 &         55 (6.99)&         46 &        100 &         74 &       100 (0.65) &         81 &         22 (6.96) &         65 &        100 &         71 &         77 (0.75) \\

 &          5 &        100 &         87 &         96 (5.45) &          0 &        100 &         95 &        100 (0.60) &         85 &         49 (6.99)&          1 &        100 &         77 &       100 (0.70) &         73 &         19 (6.91) &          9 &        100 &         71 &         97 (0.70)  \\

 &          5 &        200 &         89 &         89 (5.11)&          0 &        100 &         93 &        100 (0.70) &         86 &         43 (6.75) &          0 &        100 &         80 &      100 (0.65) &         81 &         20 (6.24)&          0 &        100 &         63 &        100 (0.60)  \\

          \hline
    \end{tabular}%
}
\label{fig:CCDaucStrauss}
\end{table}

\begin{table}
\centering
\caption{Rand Index (RI) measures of all clustering methods given in Table~\ref{fig:CCDaucStrauss}.}
\resizebox{\textwidth}{!}{
     \begin{tabular}{|ccc|cccccc|cccccc|cccccc|}
          \hline \rule{0pt}{3ex}
&  \multicolumn{2}{c|}{Dist.}    & \multicolumn{6}{c|}{$d=2$}   & \multicolumn{6}{c|}{$d=3$}   & \multicolumn{6}{c|}{$d=5$}  \\
          \hline \rule{0pt}{3ex}
     &     \multirow{2}{*}{$K$} & \multirow{2}{*}{$N$} & \multirow{2}{*}{ RK-CCD  } & \multirow{2}{*}{ KS-CCD   } & AP & AP & \multirow{2}{*}{ FCmeans } & \multirow{2}{*}{ pdfC  } & \multirow{2}{*}{ RK-CCD  } & \multirow{2}{*}{ KS-CCD  } & AP  & AP  & \multirow{2}{*}{ FCmeans } & \multirow{2}{*}{ pdfC  } & \multirow{2}{*}{ RK-CCD  } & \multirow{2}{*}{ KS-CCD  } & AP & AP & \multirow{2}{*}{ FCmeans } & \multirow{2}{*}{ pdfC } \\
&&&&&($q=0.5$)&($q=0$)&&&&&($q=0.5$)&($q=0$)&&&&&($q=0.5$)&($q=0$)&&\\
\multirow{12}[0]{*}{Unif.} &          2 &         30 &      0.978 &      0.973 &      0.210 &      1.000 &      1.000 &      0.968 &      0.998 &      0.925 &      0.420 &      1.000 &      1.000 &      0.937 &      0.980 &      0.920 &      0.540 &      1.000 &      1.000 &      0.824 \\

 &          2 &         50 &      0.997 &      0.985 &      0.080 &      0.970 &      1.000 &      0.985 &      0.999 &      0.865 &      0.210 &      1.000 &      1.000 &      0.977 &      1.000 &      0.890 &      0.180 &      0.980 &      1.000 &      0.929 \\

 &          2 &        100 &      0.992 &      0.986 &      0.010 &      0.610 &      1.000 &      0.993 &      0.998 &      0.886 &      0.020 &      0.830 &      1.000 &      0.997 &      0.999 &      0.866 &      0.020 &      0.850 &      1.000 &      0.970 \\

 &          2 &        200 &      0.990 &      0.983 &      0.000 &      0.130 &      1.000 &      0.999 &      0.999 &      0.916 &      0.000 &      0.280 &      1.000 &      1.000 &      0.999 &      0.876 &      0.000 &      0.360 &      1.000 &      1.000 \\

 &          3 &         30 &      0.945 &      0.966 &      0.460 &      0.999 &      0.973 &      0.996 &      0.971 &      0.932 &      0.520 &      0.997 &      0.964 &      0.977 &      0.983 &      0.870 &      0.760 &      0.999 &      0.975 &      0.802 \\

 &          3 &         50 &      0.960 &      0.969 &      0.150 &      0.999 &      0.979 &      0.995 &      0.974 &      0.923 &      0.240 &      0.999 &      0.977 &      0.992 &      0.979 &      0.886 &      0.400 &      1.000 &      0.971 &      0.936 \\

 &          3 &        100 &      0.971 &      0.972 &      0.000 &      0.970 &      0.973 &      0.998 &      0.990 &      0.946 &      0.020 &      0.990 &      0.966 &      0.998 &      0.984 &      0.875 &      0.030 &      0.960 &      0.978 &      0.997 \\

 &          3 &        200 &      0.960 &      0.969 &      0.000 &      0.390 &      0.962 &      0.998 &      0.970 &      0.937 &      0.000 &      0.740 &      0.949 &      0.997 &      0.978 &      0.858 &      0.000 &      0.610 &      0.969 &      0.998 \\

 &          5 &         30 &      0.930 &      0.978 &      0.589 &      0.867 &      0.988 &      0.994 &      0.984 &      0.974 &      0.829 &      0.978 &      0.977 &      0.990 &      0.989 &      0.917 &      0.859 &      0.989 &      0.986 &      0.861 \\

 &          5 &         50 &      0.967 &      0.983 &      0.100 &      0.999 &      0.987 &      0.999 &      0.985 &      0.970 &      0.230 &      0.998 &      0.985 &      0.993 &      0.989 &      0.899 &      0.380 &      1.000 &      0.994 &      0.949 \\

 &          5 &        100 &      0.985 &      0.981 &      0.000 &      0.999 &      0.993 &      1.000 &      0.995 &      0.976 &      0.010 &      0.999 &      0.991 &      0.999 &      0.992 &      0.890 &      0.000 &      1.000 &      0.979 &      0.997 \\

 &          5 &        200 &      0.994 &      0.889 &      0.000 &      0.910 &      0.990 &      1.000 &      0.996 &      0.973 &      0.000 &      0.969 &      0.974 &      1.000 &      0.992 &      0.899 &      0.000 &      0.850 &      0.982 &      1.000 \\
\hline
     &     \multirow{2}{*}{$K$} & \multirow{2}{*}{$N$} & \multirow{2}{*}{ RK-CCD  } & \multirow{2}{*}{ KS-CCD   } & AP & AP & \multirow{2}{*}{ FCmeans } & \multirow{2}{*}{ pdfC  } & \multirow{2}{*}{ RK-CCD  } & \multirow{2}{*}{ KS-CCD  } & AP  & AP  & \multirow{2}{*}{ FCmeans } & \multirow{2}{*}{ pdfC  } & \multirow{2}{*}{ RK-CCD  } & \multirow{2}{*}{ KS-CCD  } & AP & AP & \multirow{2}{*}{ FCmeans } & \multirow{2}{*}{ pdfC } \\
&&&&&($q=0.5$)&($q=0$)&&&&&($q=0.5$)&($q=0$)&&&&&($q=0.5$)&($q=0$)&&\\
\multirow{12}[0]{*}{Normal} &          2 &         30 &      0.967 &      0.975 &      0.370 &      0.994 &      0.991 &      0.991 &      0.988 &      0.877 &      0.480 &      0.997 &      0.997 &      0.959 &      0.943 &      0.824 &      0.710 &      0.999 &      0.999 &      0.813 \\

 &          2 &         50 &      0.974 &      0.969 &      0.180 &      0.995 &      0.993 &      0.992 &      0.985 &      0.817 &      0.240 &      0.998 &      0.997 &      0.986 &      0.997 &      0.900 &      0.460 &      0.999 &      0.998 &      0.933 \\

 &          2 &        100 &      0.987 &      0.934 &      0.030 &      0.957 &      0.995 &      0.993 &      0.990 &      0.875 &      0.040 &      0.929 &      0.997 &      0.996 &      0.995 &      0.891 &      0.080 &      0.970 &      0.999 &      0.983 \\

 &          2 &        200 &      0.993 &      0.940 &      0.000 &      0.630 &      0.994 &      0.996 &      0.993 &      0.831 &      0.000 &      0.610 &      0.996 &      0.996 &      0.995 &      0.836 &      0.000 &      0.720 &      0.999 &      0.995 \\

 &          3 &         30 &      0.939 &      0.952 &      0.449 &      0.951 &      0.962 &      0.987 &      0.959 &      0.881 &      0.657 &      0.984 &      0.942 &      0.983 &      0.922 &      0.856 &      0.799 &      0.998 &      0.946 &      0.853 \\

 &          3 &         50 &      0.960 &      0.959 &      0.170 &      0.993 &      0.956 &      0.989 &      0.962 &      0.881 &      0.240 &      0.994 &      0.939 &      0.991 &      0.962 &      0.857 &      0.570 &      0.998 &      0.941 &      0.957 \\

 &          3 &        100 &      0.961 &      0.973 &      0.000 &      0.996 &      0.976 &      0.992 &      0.960 &      0.874 &      0.030 &      0.995 &      0.957 &      0.996 &      0.960 &      0.838 &      0.070 &      0.999 &      0.951 &      0.989 \\

 &          3 &        200 &      0.962 &      0.000 &      0.000 &      0.887 &      0.950 &      0.991 &      0.958 &      0.876 &      0.000 &      0.908 &      0.952 &      0.995 &      0.972 &      0.867 &      0.000 &      0.959 &      0.964 &      0.996 \\

 &          5 &         30 &      0.916 &      0.967 &      0.728 &      0.836 &      0.978 &      0.986 &      0.945 &      0.941 &      0.854 &      0.863 &      0.946 &      0.987 &      0.947 &      0.878 &      0.898 &      0.908 &      0.953 &      0.891 \\

 &          5 &         50 &      0.933 &      0.972 &      0.218 &      0.987 &      0.980 &      0.989 &      0.959 &      0.931 &      0.458 &      0.991 &      0.955 &      0.992 &      0.970 &      0.863 &      0.648 &      0.996 &      0.954 &      0.976 \\

 &          5 &        100 &      0.963 &      0.966 &      0.000 &      0.990 &      0.982 &      0.989 &      0.966 &      0.920 &      0.010 &      0.993 &      0.963 &      0.993 &      0.961 &      0.860 &      0.090 &      0.997 &      0.961 &      0.995 \\

 &          5 &        200 &      0.979 &      0.874 &      0.000 &      0.993 &      0.981 &      0.990 &      0.974 &      0.892 &      0.000 &      0.995 &      0.972 &      0.994 &      0.975 &      0.839 &      0.000 &      0.998 &      0.954 &      0.998 \\
\hline
\end{tabular}
}
\label{fig:CCDaucstrauss_auc}
\end{table}

The simulation settings, of which we illustrated the performance in Table~\ref{fig:CCDaucStrauss}, were free of noise. Clustering in higher dimensions is often challenging, especially for density-based clustering methods. Now, we simulate data sets with clusters which are accompanied with some artificial noise.
That is, we repeat the experiments of Table~\ref{fig:CCDaucStrauss} with some added background noise. In this third experiment, we generate clusters which are multivariate normally distributed in $\R^d$ with mean vector $M_k$ and covariance matrix $\Sigma=I_{d \times d}(r/5)$ where $\theta=0.3$ and $r=0.15$. Here, cluster centers $\mathbf{M} = \{M_1,M_2,\ldots,M_{\K}\} \subset [0,1]^d$ are randomly drawn from a box in $\R^d$ that are at least $t \in \R_+$ apart from each other (a Strauss Process with $\beta=0$). We also introduce some noise to the data set by adding points of uniformly distributed points in the box $[-r,1+r]^d$. The noise is of size $m=0.2 \K n$ which is five times less of the total number of points in the original unnoisy data set. We illustrate two depictions of this simulation setting in Figure~\ref{fig:ccdripley_noise}.


\begin{figure}[!h]
\centering
\includegraphics[scale=0.3]{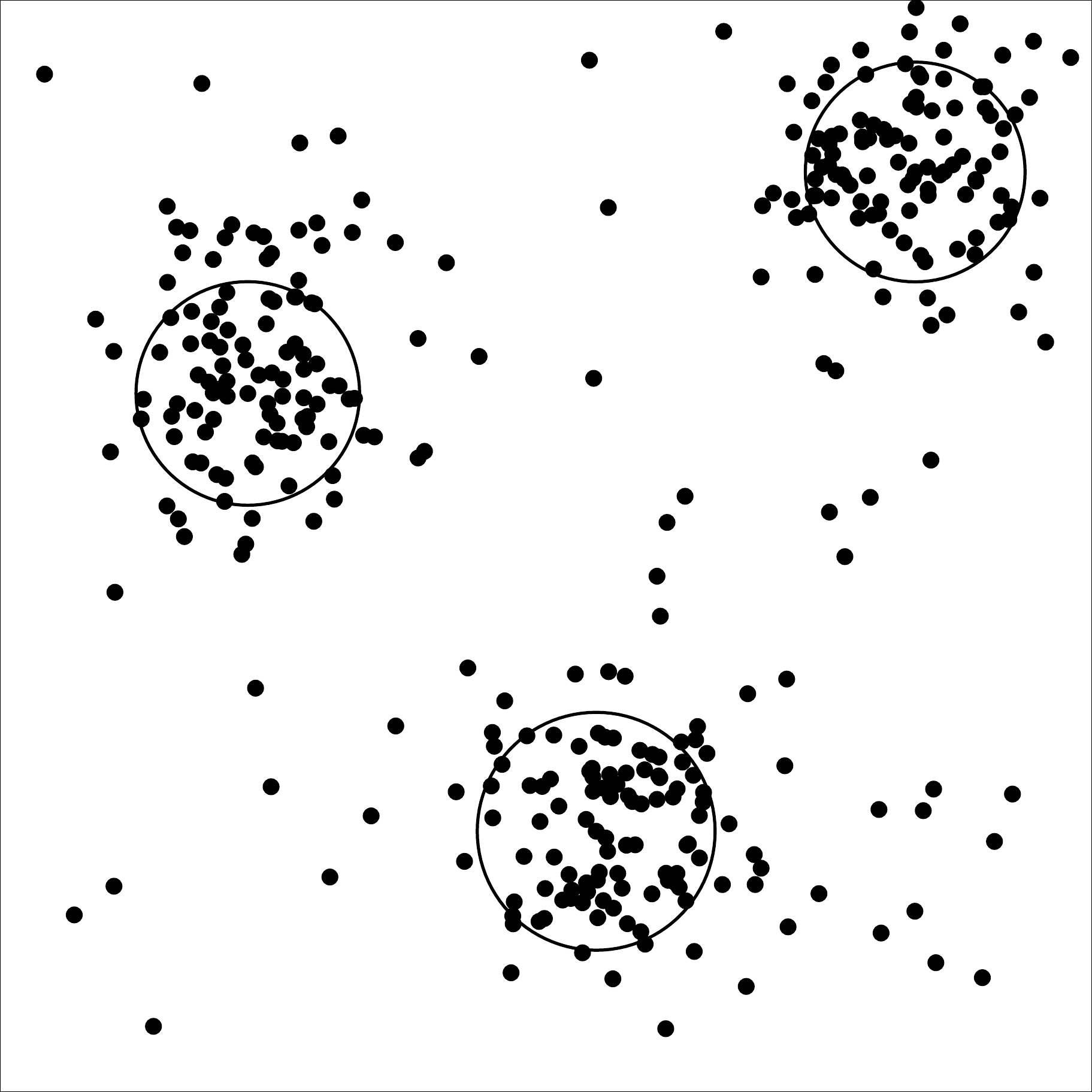}
\caption{A depiction of RK-CCDs (one with bivariate uniformly distributed clusters and the other with bivariate normally distributed) on the third simulation study setting with 3 clusters whose centers are drawn from a Strauss Process with $\beta=0$ and each cluster is bivariate normally distributed with $\Sigma=I_{d \times d}(r/5)$. A noisy set of samples have been added to the data set where the noisy points are uniformly distributed in the box $[-0.15,1.15]^d$. Here, the covering balls of the elements of $S(G_{MD})$ are provided by Algorithm~\ref{alg:ccdripley}, and each covering ball represents a cluster.}
\label{fig:ccdripley_noise}
\end{figure}


The relative frequency of success and RI measures (of 100 Monte Carlo replicates) of the third experiment have been given in Tables~\ref{fig:CCDaucNoise} and ~\ref{fig:CCDaucNoise_auc}. We again choose $\delta=5.00,5.01,5.02,\ldots,15$ as the set of intensity parameters for KS-CCDs.
Even in the presence of noise, RK-CCDs were able to achieve high success rates,
but perhaps the best performing method here is the pdfC. AP algorithm with $q=0.5$ is not successful in correctly partitioning the data sets at all while the methods perform relatively well when $q=0$. It is seen that, in all simulation studies, $q=0$ performed better than $q=0.5$. However, in the following sections, we will show that $q=0.5$ may achieve better partitioning than $q=0$ for some data sets.
But more importantly, the third experiment indicate that FCmeans is not as robust to noise as RK-CCD for low $n$ and high $d$ while KS-CCDs and pdfC methods achieving considerably good success rates.

Whether the data sets have noise or not, the performance of these clustering algorithms degrade with increasing number of clusters and dimensionality. Hence, similar to other clustering algorithms, our CCDs are also affected by the existence of clusters which are close in proximity and have close average inter-cluster and intra-cluster distances.
The success rates of all other clustering methods degrade with increasing number of clusters;
however, RK-CCDs perform relatively well compared to all these methods. Also, the optimum parameters $\delta$ and $h$ behave similar to those provided in Tables~\ref{fig:CCDauc} and~\ref{fig:CCDaucStrauss}.

\begin{table}[t]
\centering
\caption{Success rates and optimum parameters of all clustering methods where cluster centers are drawn from a Strauss process with background noise}
\resizebox{\textwidth}{!}{
     \begin{tabular}{|cc|cccccc|cccccc|cccccc|}
          \hline \rule{0pt}{3ex}
&  \multicolumn{1}{c|}{Dist.}    & \multicolumn{6}{c|}{$d=2$}   & \multicolumn{6}{c|}{$d=3$}   & \multicolumn{6}{c|}{$d=5$}  \\
          \hline \rule{0pt}{3ex}
        \multirow{2}{*}{$K$} & \multirow{2}{*}{$N$} & \multirow{2}{*}{ RK-CCD  } & \multirow{2}{*}{ KS-CCD ($\delta$)} & AP & AP & \multirow{2}{*}{ FCmeans } & \multirow{2}{*}{ pdfC ($h$) } & \multirow{2}{*}{ RK-CCD  } & \multirow{2}{*}{ KS-CCD ($\delta$)} & AP  & AP  & \multirow{2}{*}{ FCmeans } & \multirow{2}{*}{ pdfC ($h$)} & \multirow{2}{*}{ RK-CCD  } & \multirow{2}{*}{ KS-CCD ($\delta$)} & AP & AP & \multirow{2}{*}{ FCmeans } & \multirow{2}{*}{ pdfC ($h$)} \\
&&&&($q=0.5$)&($q=0$)&&&&&($q=0.5$)&($q=0$)&&&&&($q=0.5$)&($q=0$)&&\\
         2 &         30 &         97 &         99 (5.91) &          0 &         83 &         49 &         96 (1.10) &         99 &         90 (7.00)&          0 &         69 &         42 &        100 (0.90) &         91 &         81 (6.99) &          0 &         59 &         29 &         87 (0.65)  \\
         2 &         50 &         96 &         99 (6.23) &          0 &         50 &         59 &         90 (1.50) &         99 &         94 (7.00) &          0 &         31 &         40 &         99 (0.75) &         98 &         75 (6.70) &          0 &         27 &         34 &         99 (0.90)  \\
         2 &        100 &        100 &         99 (5.77) &          0 &         13 &         49 &         87 (1.60) &        100 &         93 (6.89) &          0 &          5 &         51 &         94 (1.25) &         99 &         75 (5.43) &          0 &          2 &         28 &        100 (0.55)  \\
         2 &        200 &        100 &         99 (5.20) &          0 &          0 &         54 &         94 (2.00)&        100 &         92 (5.62) &          0 &          0 &         46 &         91 (2.05) &         99 &         85 (5.33) &          0 &          0 &         33 &         99 (1.10)  \\
         3 &         30 &         66 &         84 (5.85)&          0 &         89 &         48 &         93 (0.60) &         62 &         54 (6.82) &          0 &         80 &         39 &         97 (0.60) &         75 &         53 (6.95) &          0 &         68 &         38 &         67 (0.60)  \\
         3 &         50 &         77 &         89 (5.90) &          0 &         75 &         58 &         92 (1.15) &         82 &         69 (6.98) &          0 &         47 &         41 &         95 (0.85) &         80 &         56 (6.88) &          0 &         43 &         32 &         95 (0.70)  \\
         3 &        100 &         82 &         91 (6.04) &          0 &         25 &         58 &         93 (1.35) &         74 &         63 (6.97) &          0 &          8 &         34 &         90 (1.15)	 &         80 &         53 (6.94) &          0 &          3 &         33 &        100 (0.75)  \\
         3 &        200 &         84 &         93 (5.67) &          0 &          1 &         51 &         93 (1.45) &         77 &         68 (6.85) &          0 &          0 &         41 &         83 (1.50) &         63 &         40 (6.77) &          0 &          0 &         38 &         96 (0.55)  \\
         5 &         30 &         27 &         85 (5.22) &          4 &         71 &         71 &         97 (0.70) &         62 &         68 (6.97)&          0 &         60 &         48 &         92 (0.50) &         61 &         27 (6.25) &          0 &         69 &         37 &         41 (0.60)  \\
         5 &         50 &         51 &         94 (5.50) &          0 &         94 &         86 &         93 (0.75) &         68 &         66 (7.00)&          0 &         66 &         58 &         95 (0.70) &         64 &         26 (6.96) &          0 &         49 &         50 &         84 (0.50)  \\
         5 &        100 &         75 &         97 (5.46) &          0 &         71 &         84 &         97 (1.10) &         78 &         61 (6.88) &          0 &         17 &         63 &         86 (0.90) &         49 &         20 (6.66) &          0 &          4 &         53 &         96 (0.65)  \\
         5 &        200 &         68 &         80 (5.00) &          0 &         18 &         87 &        100 (1.15) &         73 &         43 (6.15)&          0 &          0 &         60 &         83 (1.30) &         53 &         41 (5.80) &          0 &          0 &         51 &        100 (0.60) \\
\hline
\end{tabular}
}
\label{fig:CCDaucNoise}
\end{table}

\begin{table}
\centering
\caption{Rand Index (RI) measures of all clustering methods given in Table~\ref{fig:CCDaucNoise}.}
\resizebox{\textwidth}{!}{
     \begin{tabular}{|cc|cccccc|cccccc|cccccc|}
          \hline \rule{0pt}{3ex}
&  \multicolumn{1}{c|}{Dist.}    & \multicolumn{6}{c|}{$d=2$}   & \multicolumn{6}{c|}{$d=3$}   & \multicolumn{6}{c|}{$d=5$}  \\
          \hline \rule{0pt}{3ex}
        \multirow{2}{*}{$K$} & \multirow{2}{*}{$N$} & \multirow{2}{*}{ RK-CCD  } & \multirow{2}{*}{ KS-CCD   } & AP & AP & \multirow{2}{*}{ FCmeans } & \multirow{2}{*}{ pdfC  } & \multirow{2}{*}{ RK-CCD  } & \multirow{2}{*}{ KS-CCD  } & AP  & AP  & \multirow{2}{*}{ FCmeans } & \multirow{2}{*}{ pdfC  } & \multirow{2}{*}{ RK-CCD  } & \multirow{2}{*}{ KS-CCD  } & AP & AP & \multirow{2}{*}{ FCmeans } & \multirow{2}{*}{ pdfC } \\
&&&&($q=0.5$)&($q=0$)&&&&&($q=0.5$)&($q=0$)&&&&&($q=0.5$)&($q=0$)&&\\
         2 &         30 &      0.974 &      0.959 &      0.000 &      0.664 &      0.990 &      0.988 &      0.981 &      0.875 &      0.000 &      0.569 &      0.995 &      0.991 &      0.945 &      0.889 &      0.000 &      0.484 &      0.994 &      0.931 \\
         2 &         50 &      0.972 &      0.957 &      0.000 &      0.407 &      0.993 &      0.939 &      0.995 &      0.926 &      0.000 &      0.255 &      0.995 &      0.995 &      0.988 &      0.859 &      0.000 &      0.227 &      0.998 &      0.993 \\
         2 &        100 &      0.988 &      0.932 &      0.000 &      0.108 &      0.989 &      0.974 &      0.991 &      0.862 &      0.000 &      0.041 &      0.996 &      0.992 &      0.993 &      0.788 &      0.000 &      0.016 &      0.999 &      0.998 \\
         2 &        200 &      0.993 &      0.821 &      0.000 &      0.000 &      0.991 &      0.957 &      0.991 &      0.793 &      0.000 &      0.000 &      0.995 &      0.933 &      0.985 &      0.792 &      0.000 &      0.000 &      0.999 &      0.995 \\
         3 &         30 &      0.912 &      0.946 &      0.000 &      0.717 &      0.938 &      0.986 &      0.908 &      0.881 &      0.000 &      0.639 &      0.945 &      0.988 &      0.930 &      0.882 &      0.000 &      0.539 &      0.955 &      0.900 \\
         3 &         50 &      0.937 &      0.958 &      0.000 &      0.593 &      0.949 &      0.977 &      0.953 &      0.920 &      0.000 &      0.373 &      0.951 &      0.993 &      0.952 &      0.869 &      0.000 &      0.343 &      0.939 &      0.988 \\
         3 &        100 &      0.953 &      0.966 &      0.000 &      0.198 &      0.952 &      0.984 &      0.940 &      0.901 &      0.000 &      0.064 &      0.909 &      0.992 &      0.950 &      0.878 &      0.000 &      0.025 &      0.951 &      0.999 \\
         3 &        200 &      0.953 &      0.948 &      0.000 &      0.008 &      0.952 &      0.988 &      0.945 &      0.896 &      0.000 &      0.000 &      0.942 &      0.981 &      0.910 &      0.819 &      0.000 &      0.000 &      0.944 &      0.995 \\
         5 &         30 &      0.863 &      0.955 &      0.034 &      0.598 &      0.967 &      0.985 &      0.933 &      0.932 &      0.000 &      0.505 &      0.942 &      0.987 &      0.929 &      0.864 &      0.000 &      0.583 &      0.936 &      0.919 \\
         5 &         50 &      0.904 &      0.963 &      0.000 &      0.794 &      0.978 &      0.987 &      0.944 &      0.944 &      0.000 &      0.556 &      0.939 &      0.994 &      0.914 &      0.872 &      0.000 &      0.413 &      0.959 &      0.981 \\
         5 &        100 &      0.948 &      0.959 &      0.000 &      0.598 &      0.973 &      0.989 &      0.959 &      0.937 &      0.000 &      0.144 &      0.951 &      0.993 &      0.877 &      0.826 &      0.000 &      0.034 &      0.960 &      0.993 \\
         5 &        200 &      0.934 &      0.791 &      0.000 &      0.151 &      0.979 &      0.990 &      0.942 &      0.802 &      0.000 &      0.000 &      0.954 &      0.989 &      0.904 &      0.746 &      0.000 &      0.000 &      0.959 &      0.998 \\	
\hline
\end{tabular}
}
\label{fig:CCDaucNoise_auc}
\end{table}

\section{Real Data Examples and Arbitrarily Shaped Clusters} \label{sec:realdata}

In this section, we assess the performance of RK-CCDs in real life data sets with convex and arbitrarily shaped clusters \citep{BacheLichman,fdez2011}.
We first illustrate the estimated number of clusters found by all methods and the corresponding RI measure of all real data sets we consider in Table~\ref{fig:realdata_auc}.
In Figure~\ref{fig:realdata}, we illustrate some of these data sets and the covering balls of RK-CCDs associated with the clusters which ultimately indicate where the cluster centers are.
We also report on the optimum parameters of KS-CCDs and pdfC methods that achieve a partitioning with the true number of clusters and with maximum RI. For those data sets with no indication of the true clusters (or labels), we use the silhouette measure to assess the performance of the partitioning. Prior to clustering, we normalize some of the features and use principal component analysis (PCA) to extract the principal components of some of these real data sets to mitigate the effect of dimensionality.

RK-CCDs successfully locate three clusters of the dimensionally reduced wine data set with 0.85 RI, respectively. The ecoli2 data set have two clusters but visually constitute three convex clusters. RK-CCDs and most of the other clustering methods were able to find these three clusters with similar RI values. On some occasions where clusters of a data set are weakly separated, we have observed that silhouette measure suggests partitioning the data into incorrect number of clusters. Here, FCmeans partitions the Iris data set into two clusters instead of three. Iris data set is indeed one of those data sets that two of its clusters are considerably close' to each other. However, RK-CCD partitions the Iris data set into three clusters since Algorithm~\ref{alg:ccdripley} locates the clusters first and then partitions the data once the true number of clusters are found. KS-CCD has only achieved a RI measure of 0.72 where RK-CCD achieved 0.89 RI, outperforming KS-CCD while still finding the exact number of clusters. Moreover, the optimum values of $\delta$ for KS-CCD are drastically different for wine and Iris data sets. For $h$ value of pdfC, however, there seems to be no considerable difference. AP algorithm with $q=0$ produces the true number of clusters in Iris, wine and seeds data sets with over 0.80 RI. For data sets with more clusters like R15 and D31, RK-CCDs locates all of their clusters with high accuracy. Other than AP($q=0$), all methods were able to find 15 clusters in R15 data set. AP algorithm, contrary to those data sets with small number of clusters, only achieves the true number of clusters in R15 and D31 for $q=0.5$. It appears that AP algorithm performs much better for data sets with many of clusters where, as $q$ decreases, the algorithm produces smaller number of partitions \citep{frey2007,bodenhofer2011}. Both OFGD and BirthDeathRates data sets visually comprise of two clusters although there exists no information on the true labels of their points, hence we decided to use the silhouette measure for performance assessment. Comparing the accuracies (or silhouette measures) of CCDs in Table~\ref{fig:realdata_auc}, we observe that RK-CCDs outperform both AP($q=0$) and AP($q=0.5$). FCmeans achieves the highest silhouette measure among all other methods (including RK-CCDs), although the accuracy of RK-CCD is quite close to the accuracy of FCmeans.

\begin{table}[t]
\caption{Success rates and optimum parameters of all clustering methods for real data sets with convex clusters. PC$d$ provides the number of principal components extracted for the associated data set. (*) Inherent number of convex clusters are 3 in ecoli2 data.
The Accuracy (abbreviated as Acc.) stands for the Rand Index for real data sets with known true clustering labels, and Silhouette index for those with no labels.}
\resizebox{\textwidth}{!}{
    \begin{tabular}{c|cccc|cc|ccc|cc|cc|cc|ccc|}
  &   \multicolumn{4}{c|}{Data}      & \multicolumn{2}{c|}{RK-CCD} & \multicolumn{3}{c|}{KS-CCD} & \multicolumn{2}{c|}{AP($q=0.5$)} & \multicolumn{2}{c|}{AP($q=0$)} &\multicolumn{2}{c|}{FCmeans} & \multicolumn{3}{c|}{pdfC} \\
          \hline \rule{0pt}{3ex}
    & $N$   & $d$     & PC$d$   & $\K$     & $\widehat{\K}$ & Acc.   & $\widehat{\K}$ & Acc.   & $\delta$ & $\widehat{\K}$ & Acc. & $\widehat{\K}$ & Acc. & $\widehat{\K}$ & Acc.   & $\widehat{\K}$ & Acc.   & $h$ \\
          \hline
    BirthDeathRates & 70 & 2 & . & 2 & 2 & 0.43 & 2 & 0.39 & 16.00 & 6 & 0.17 & 2 & 0.38 & 2 & 0.58 & 2 & 0.38 & 0.57 \\
    wine  &178 & 13    & 4     & 3     & 3     & 0.85  & 3  & 0.93  & 0.02 & 13 & 0.73 & 3 & 0.93 & 3  & 0.93  & 3     & 0.86  & 0.75 \\
    iris  &150 & 4     & .     & 3     & 3     & 0.87  & 3  & 0.72  & 4.44  & 6 & 0.66 & 3 & 0.88 & 2     & 0.76  & 3     & 0.88  & 0.59 \\
    seeds & 210 & 7 & 4 & 3 & 3 & 0.89 & 3 & 0.86 & 14.50 & 11 & 0.73 & 3 & 0.86 & 2 & 0.74 & 3 & 0.93 & 1.11 \\
    OFGD & 272 & 2 & . & 2 & 2 & 0.72 & 2 & 0.75 & 11.00 & 11 & 0.50 & 2 & 0.62 & 2 & 0.75 & 2 & 0.63 & 0.85 \\ 	
    ecoli2 &336 & 7     & 2     & 2*   & 3     & 0.50     & 3  & 0.54  & 0.30  & 15 & 0.33 & 3 & 0.54 & 3     & 0.53     & 3 & 0.54  & 0.85 \\
    R15   &600 & 2     & .     & 15    & 15    & 0.99  & 15 & 0.99  & 20.00 & 15 & 0.99 & 10 & 0.93 & 15    & 0.99  & 15    & 0.99  & 0.50 \\
    D31   &3100 & 2     & .     & 31    &   31  &  0.99    &   31    &  0.99     &  15.00     & 31 & 0.99 & 17 & 0.96 & 31    & 0.99  & 31    & 0.99  & 0.50 \\
    \hline
    \end{tabular}
}
\label{fig:realdata_auc}
\end{table}

\begin{figure}[!h]
\centering
\begin{tabular}{cc}
\includegraphics[scale=0.4]{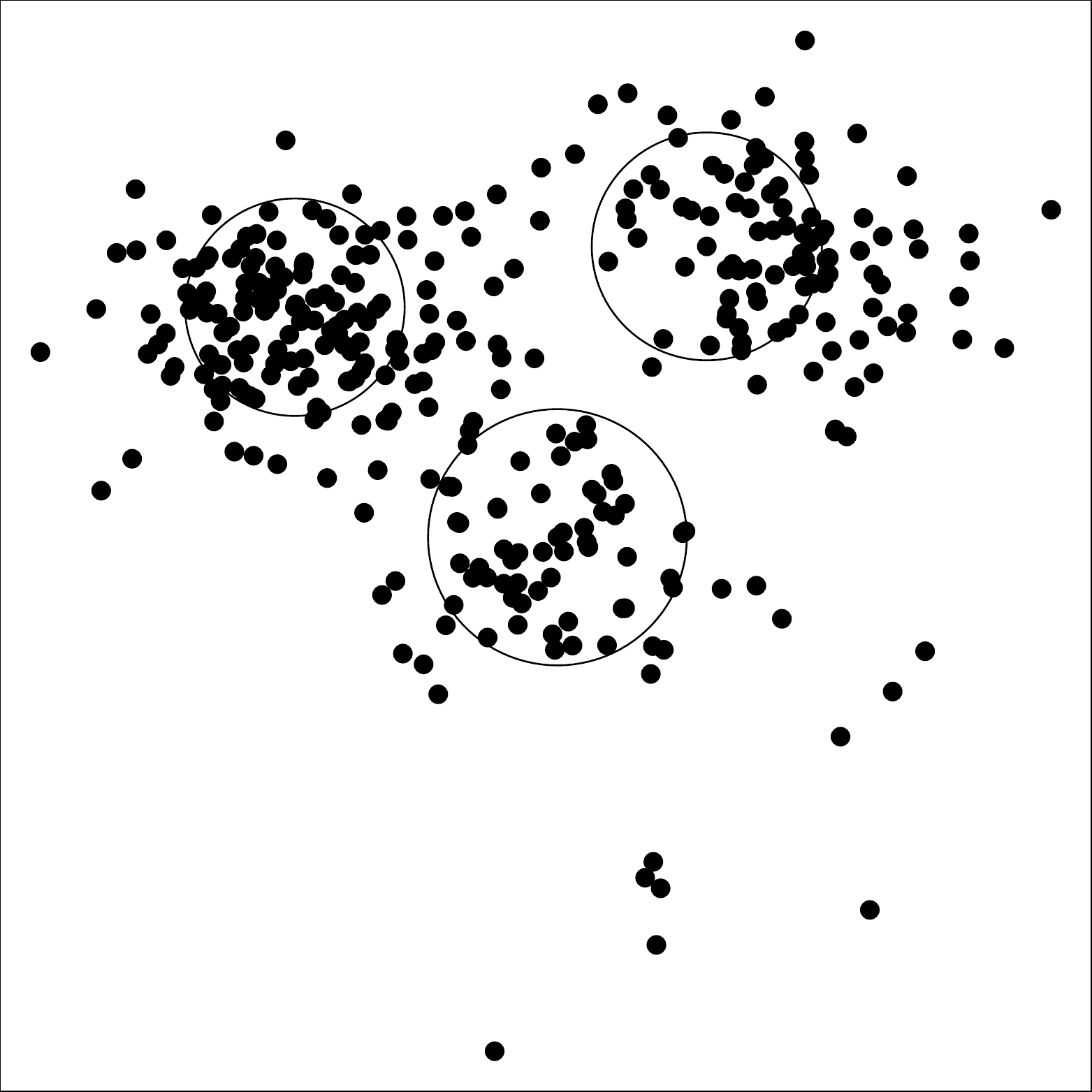} & \includegraphics[scale=0.4]{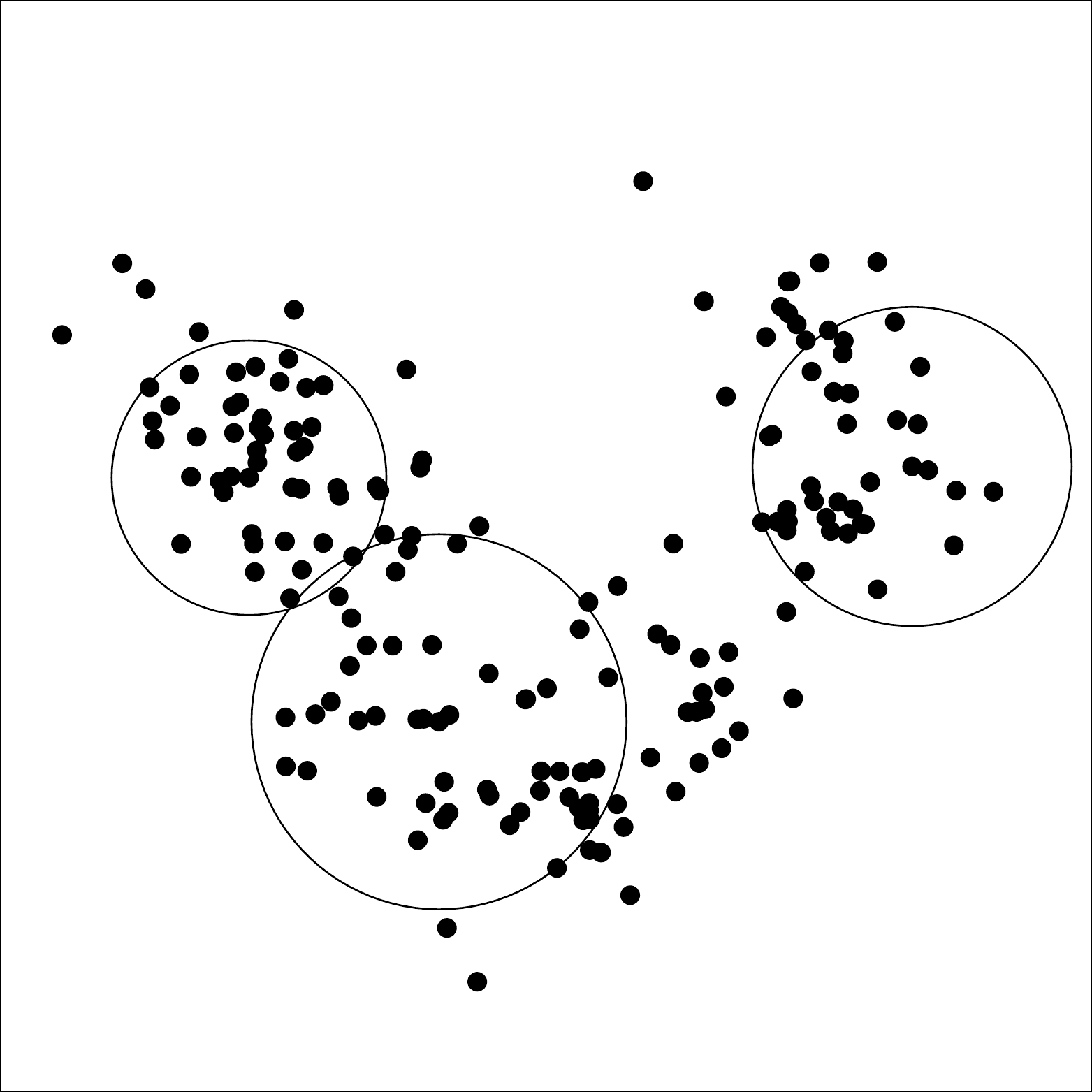} \\
(a) & (b) \\
\includegraphics[scale=0.4]{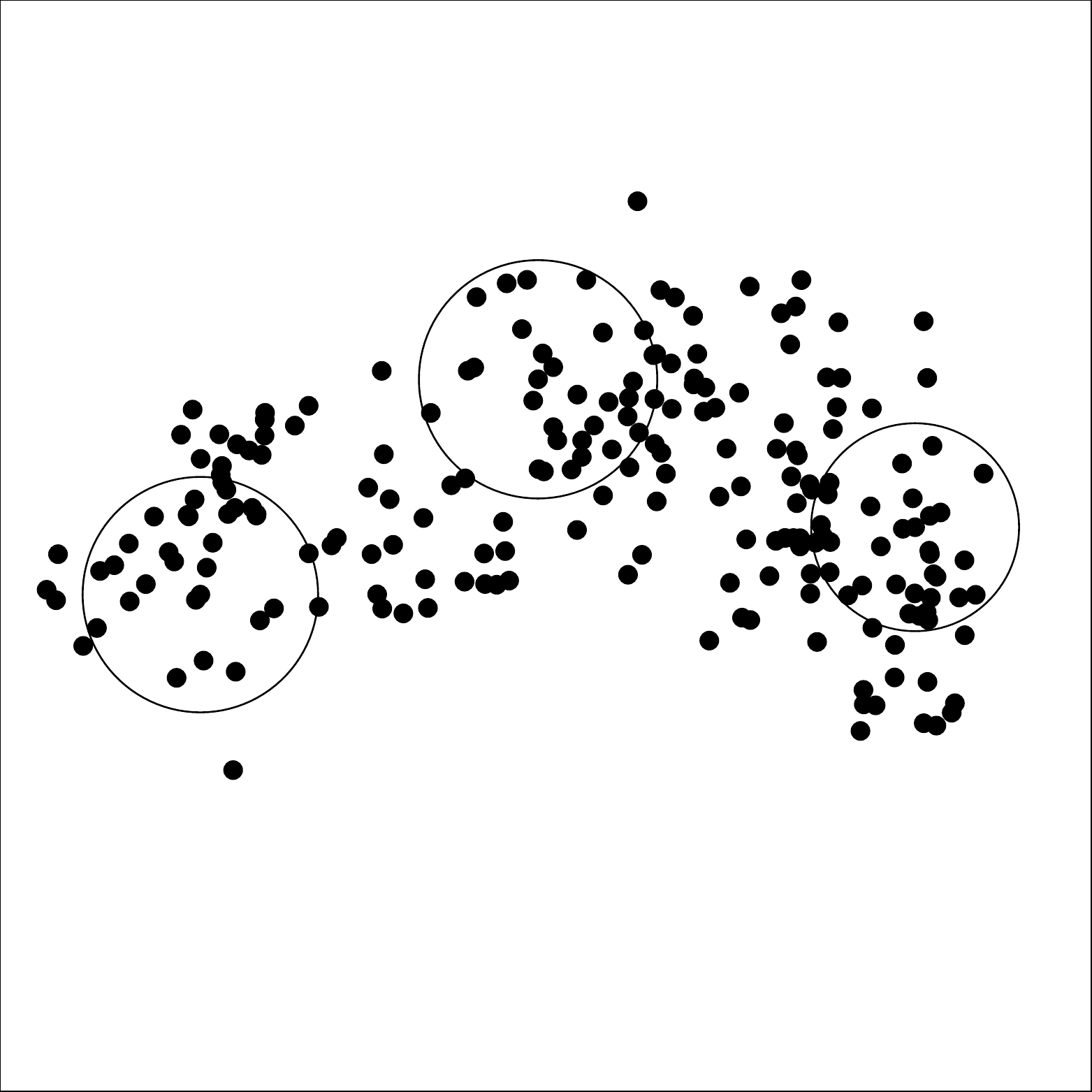} & \includegraphics[scale=0.4]{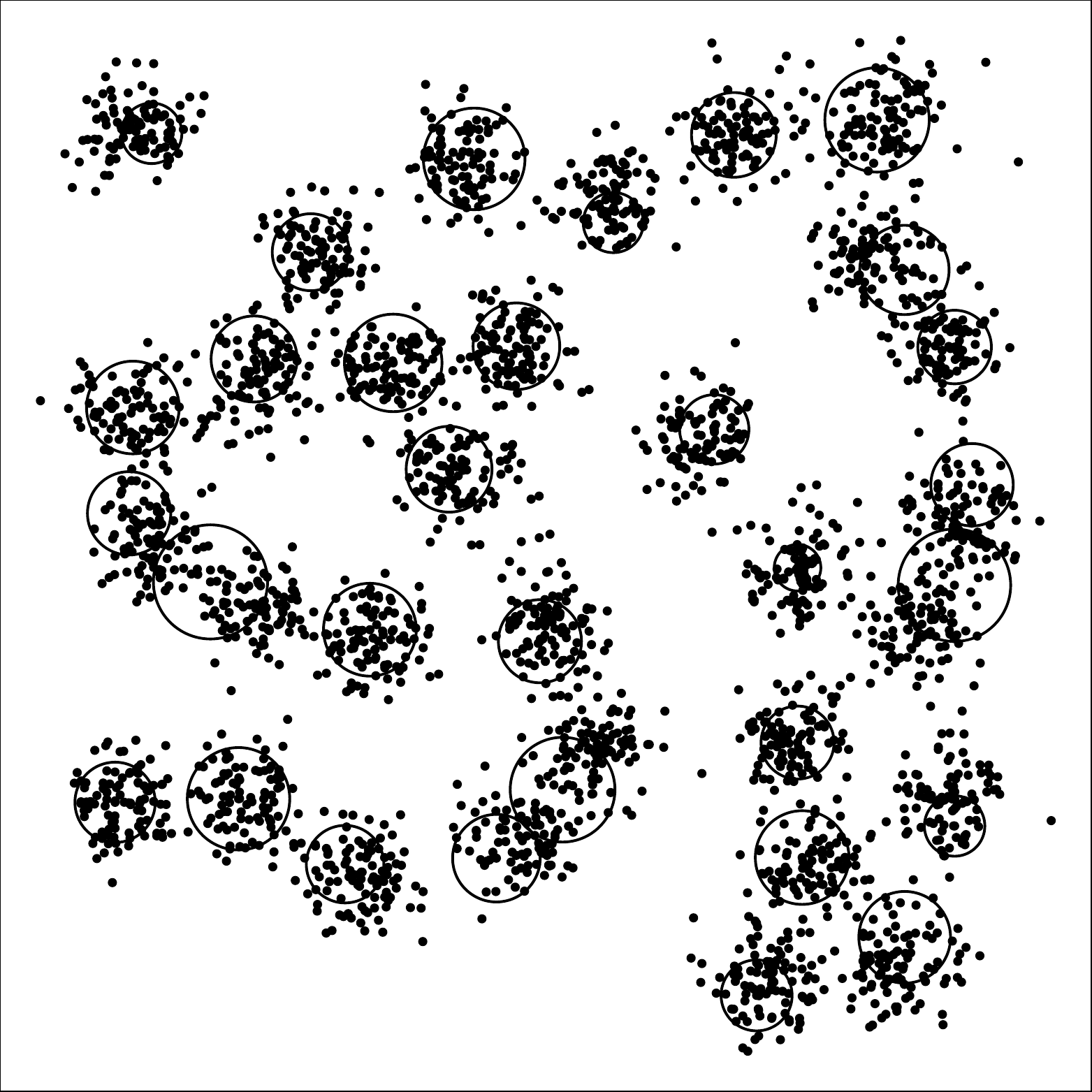} \\
(c) & (d) \\
\end{tabular}
\caption{Illustrations of the covering balls of RK-CCD associated with $S(G_{MD})$ on real data sets with convex clusters. (a) ecoli2 data set has two clusters but visually constitute 3 convex clusters (b) wine data set (c) seeds data set (d) RK-CCDs locate 30 clusters where RK-CCDs only fail to recognize two clusters. }
\label{fig:realdata}
\end{figure}

The data sets circles, moons, multishapes, in particular, aggregation are data sets with arbitrarily shaped clusters \citep{chang2008,bhissy2014,saxena2015,kassambara2017} that we use Algorithm~\ref{alg:ccdripley} of RK-CCDs to do the partitioning. However, we search for connected components instead of the minimum dominating set $S(G_{MD})$. We assess the performance of RK-CCDs and KS-CCDs against the DBSCAN algorithm. In Figure~\ref{fig:aggregationdata}, we illustrate the results of our RK-CCD algorithm on three spatial data sets, and we give the clustering results and optimum parameters of RK-CCDs, KS-CCDs and DBSCAN in Table~\ref{fig:realdata_auc2}. In all data sets, unions of covering balls do not extend far from the supports of the data sets, even though their supports are not convex clusters. All algorithms were able to find the true number of clusters. In particular, multishapes data set has clusters with different spatial intensities, but RK-CCDs are able to locate these clusters since RK-CCDs are able to detect clusters with varying inter-cluster distances. Aggregation data set has seven convex clusters but has five arbitrarily shaped clusters with two pairs of clusters weakly connected to each other. Algorithm~\ref{alg:ccdripley} successfully locates all these clusters despite the weak connection.


KS-CCD and DBSCAN algorithms were able to find the exact number of clusters in all data sets with arbitrarily shaped clusters. Optimum parameters are generally between [1,5] and [0.05,0.15] for KS-CCDs and DBSCAN algorithms, respectively. These parameters need to be initialized before the execution, and hence their selection is critical. However, RK-CCDs were able to find the true number of clusters with no given parameters.

\begin{table}[t]
\caption{Success rates and optimum parameters of all clustering methods for real data sets with arbitrarily shaped clusters. Rand index (RI) measures are all 1.00,
hence omitted.}
\centering
\begin{tabular}{c|cc|c|cc|cc|}
  &   \multicolumn{2}{c|}{Data}      & \multicolumn{1}{c|}{RK-CCD} & \multicolumn{2}{c|}{KS-CCD} & \multicolumn{2}{c|}{DBSCAN} \\
          \hline \rule{0pt}{3ex}
    & $N$  & $\K$  & $\widehat{\K}$ & $\widehat{\K}$ & $\delta$ & $\widehat{\K}$ & $\epsilon$  \\
          \hline
    multishapes       & 1000 & 4 & 4 & 4 & 1 	& 4 & 0.14 \\
    circles         & 500 & 2 & 2 & 2 & 4.5 & 2 & 0.09 \\
    moons         & 500 & 2 & 2 & 2 & 5 & 2 & 0.14 \\
    \hline
\end{tabular}
\label{fig:realdata_auc2}
\end{table}

\begin{figure}[!h]
\centering
\begin{tabular}{ccc}
\includegraphics[scale=0.18]{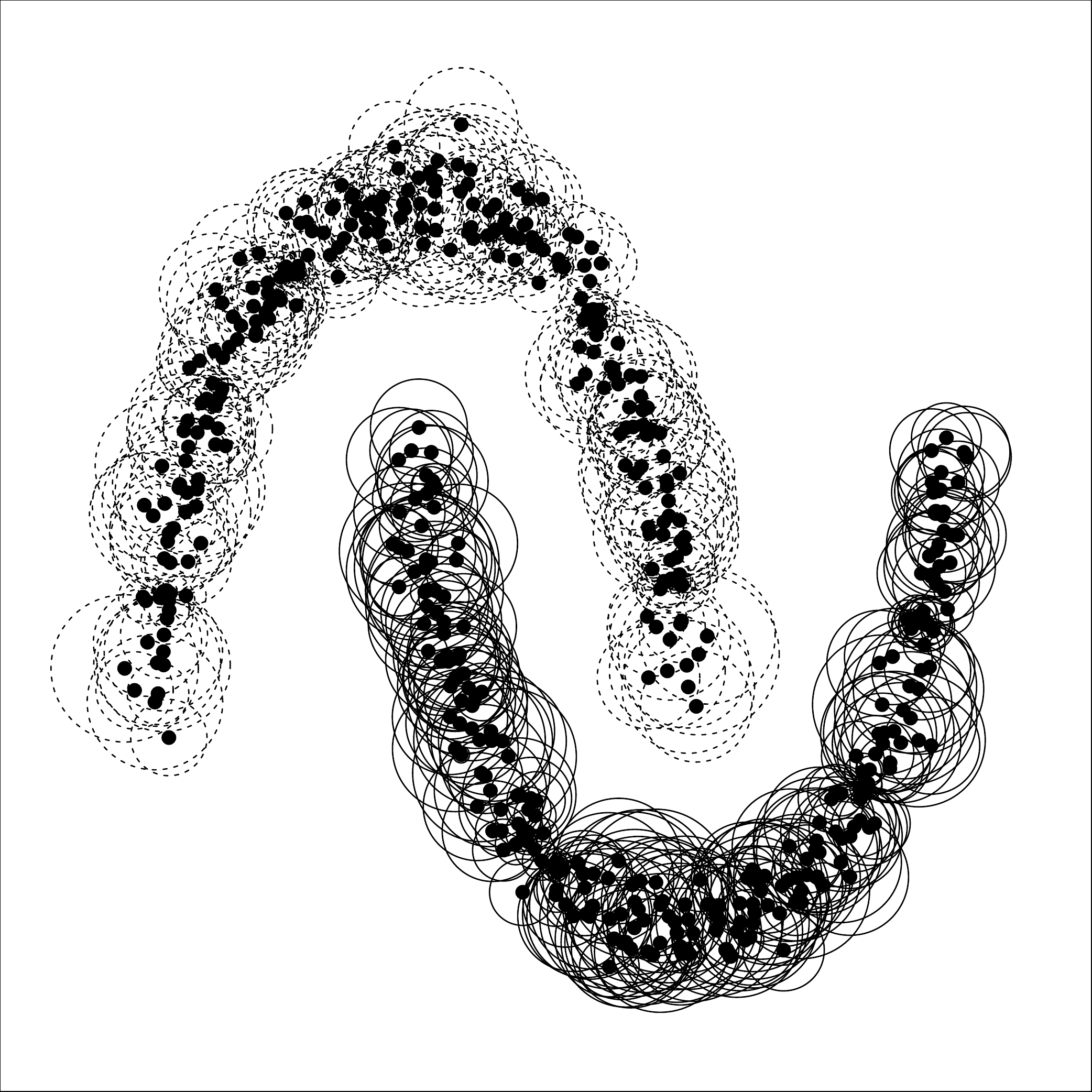} & \includegraphics[scale=0.18]{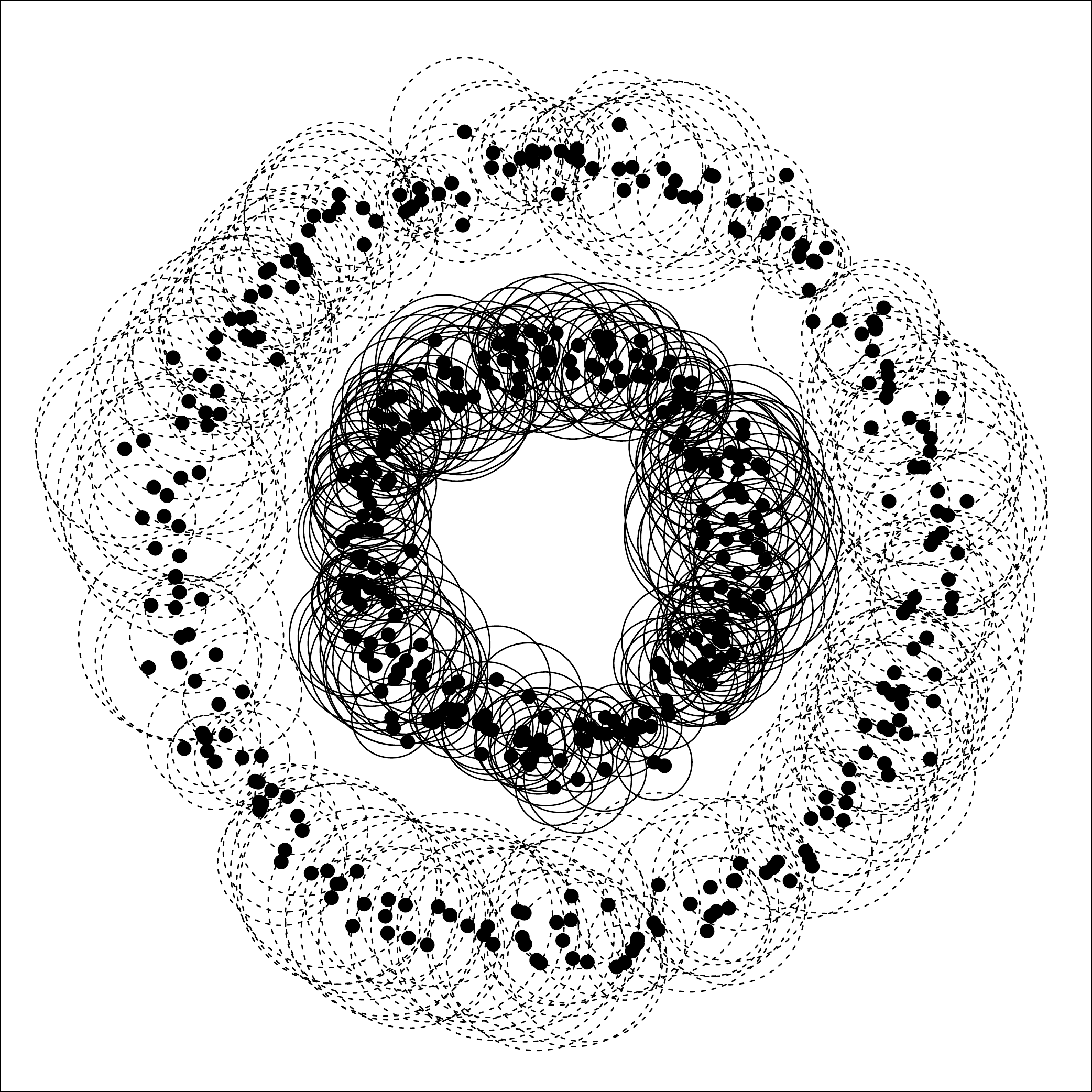} & \includegraphics[scale=0.18]{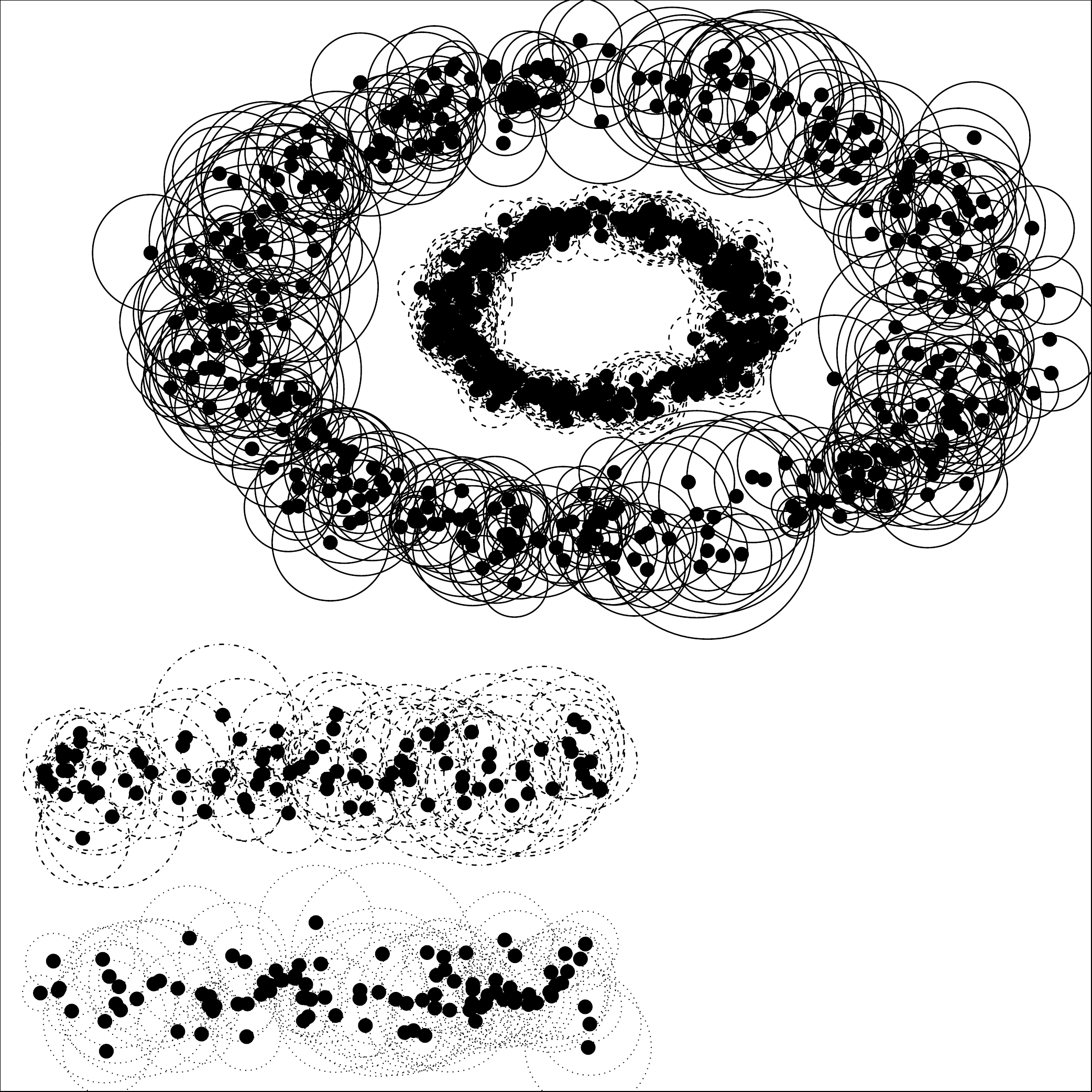} \\
\includegraphics[scale=0.18]{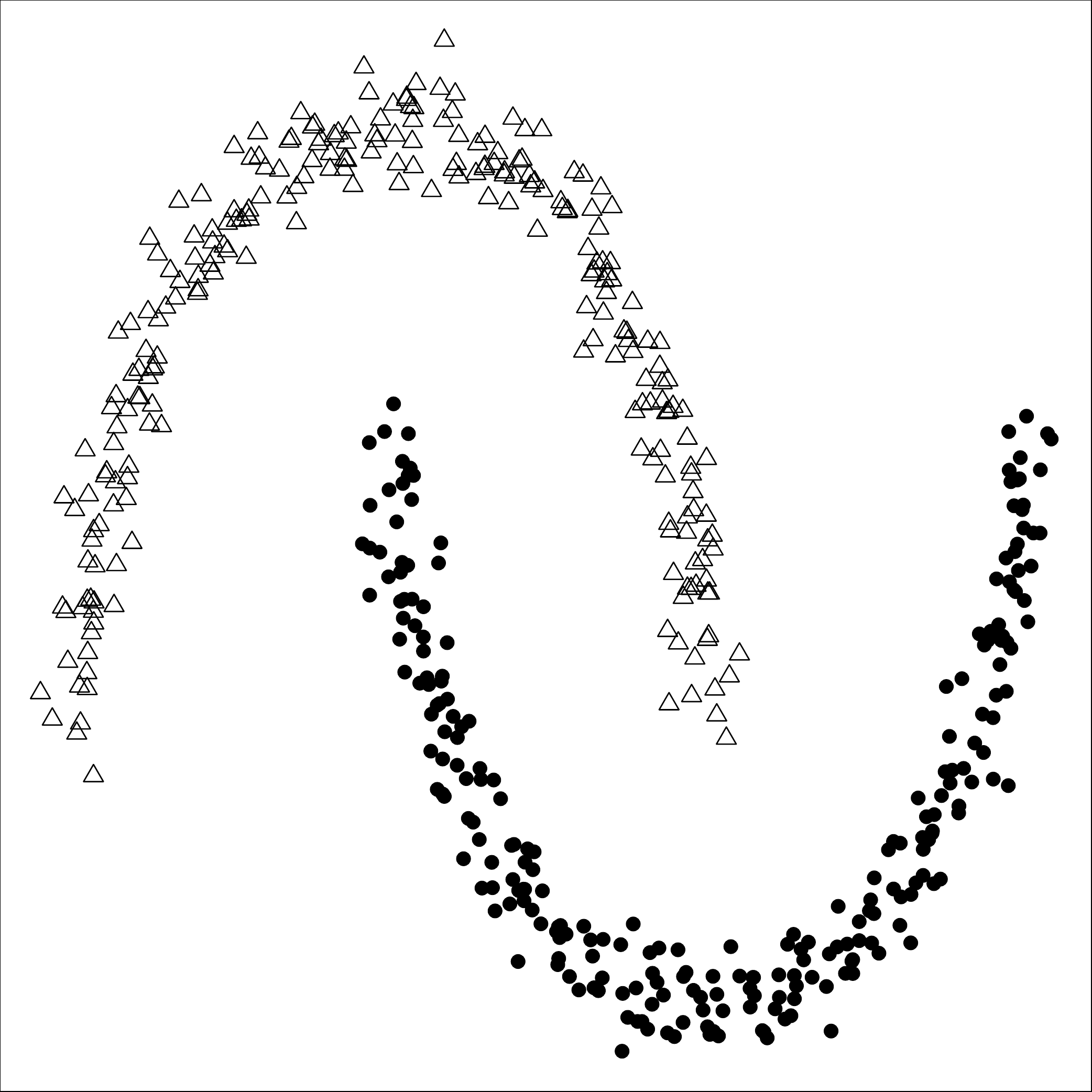} & \includegraphics[scale=0.18]{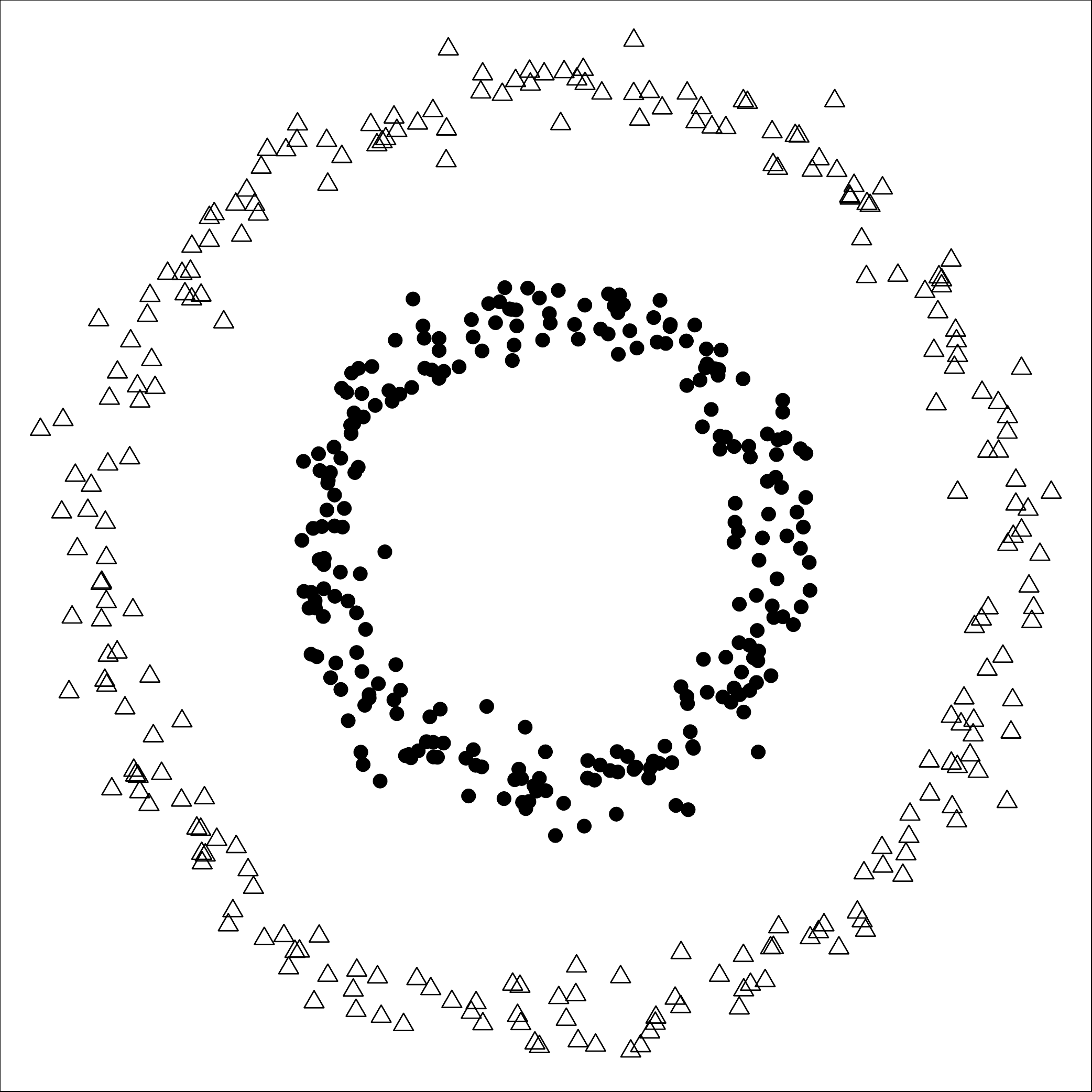} &
\includegraphics[scale=0.18]{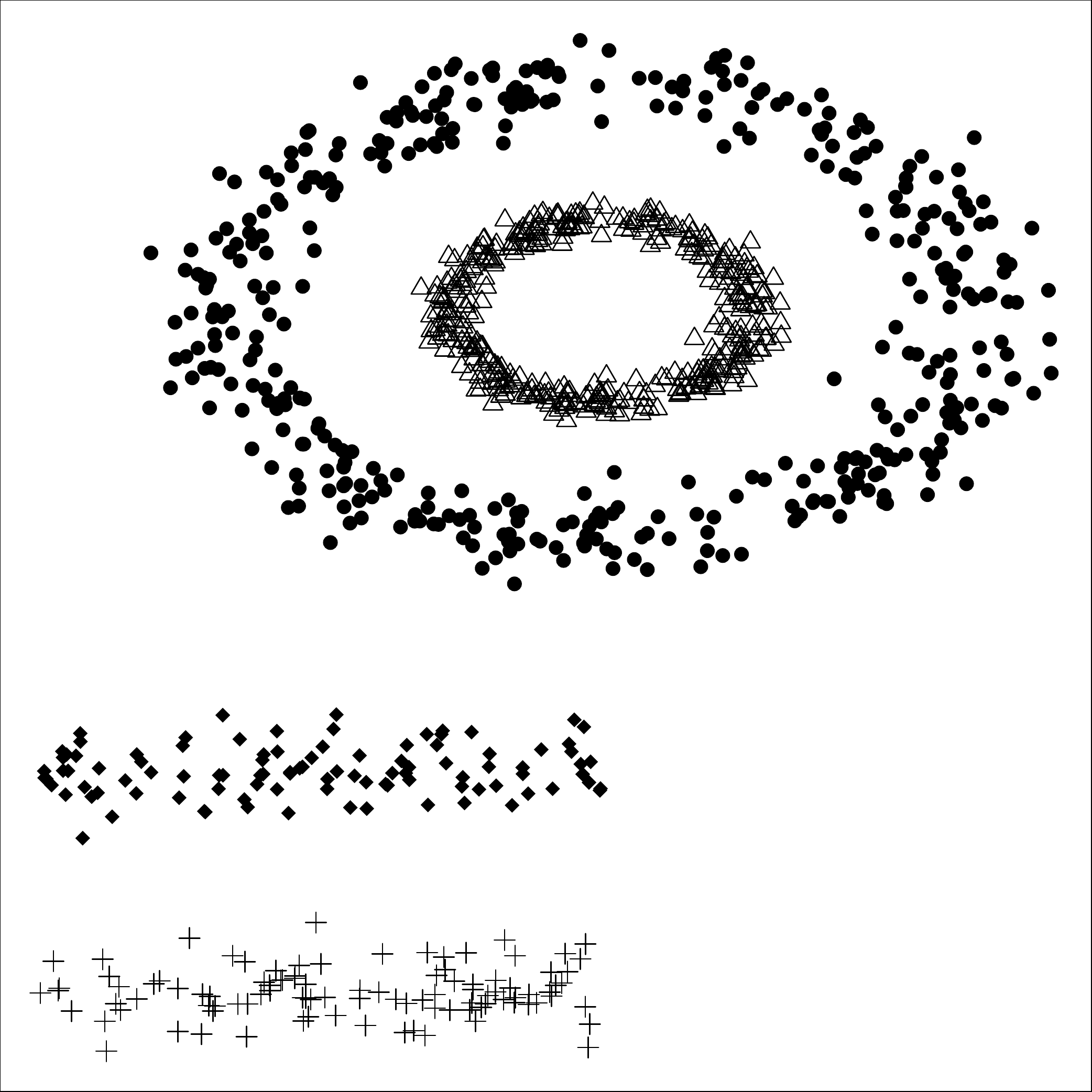} \\
(a) & (b) & (c) \\
\end{tabular}
\caption{Partitioning as a result of applying RK-CCD algorithm on data sets with arbitrarily shaped clusters.
(a) The moons data set with two moons,
(b) the circles data set with two circles,
(c) the multishapes data set. the RK-CCD covering balls of each cluster given with circles of different line types (top) and members of each cluster (found by RK-CCD algorithm) given with different point shapes (bottom).}
\label{fig:aggregationdata}
\end{figure}

\section{Conclusions and Discussion} \label{sec:conc}

We use cluster catch digraphs (CCDs) for locating clusters and for finding optimal partitionings of data sets. CCDs are unsupervised adaptations of class cover catch digraphs (CCCDs) introduced to provide graph-theoretic solutions to the class cover problem (CCP) \citep{priebe:2003b,manukyan2016}. We use these covers to estimate the support of the distribution of the data set, presumably a mixture distribution with a finite number of components (or clusters).
We develop methods to increase the performance of recently introduced clustering methods based on CCDs, namely KS-CCDs \citep{devinney2003,marchette2004}, and use them as a framework for novel clustering algorithms. These methods are based on spatial pattern tests that incorporate Ripley's $K$ function, hence we refer to them as RK-CCDs. One important advantage of RK-CCDs is that they find and estimate the number of clusters in the data set without the
necessity of specifying any parameters a priori. Density-based methods often estimate the number of clusters via specifying a set of parameters representing the assumed density, or spatial intensity, around points of the data sets. DBSCAN and pdfCluster are examples of such methods. Both KS-CCDs and RK-CCDs are hybrids of density- and graph-based methods, and in particular, RK-CCDs estimate the spatial intensity parameter using Ripley's $K$ function.

Our estimates of the support are provided by a collection of balls that encapsulates the data set. These estimates constitute a \emph{cover} given by the minimum dominating sets (MDSs) of CCDs. In KS-CCDs, the radius of each covering ball is the distance that maximizes a K-S based statistic. However, it requires an intensity parameter of the spatial distribution of the null hypothesis which assumes the points are drawn from a homogeneous Poisson process. In RK-CCDs, however, the radius of each covering ball is chosen by a test based on the estimated Ripley's $K$ function $\widehat{K}(t)$; that is, the radius is the smallest radius that rejects the null hypothesis, indicating that the spatial distribution of the point pattern inside the covering ball significantly deviates from complete spatial randomness (CSR).
The rejection implies that the ball covers a region of the domain outside of the class support.
More importantly, RK-CCDs find an optimal partitioning without any
a priori parameter specification,
since $\widehat{K}(t)$ is calculated by estimates of the spatial intensity parameter. We separately investigate data sets exhibiting convex and arbitrarily shaped clusters, and provide two variants of algorithms for both KS-CCDs and RK-CCDs.
We show that convex clusters are identified by MDSs of the intersection graphs, whereas we look for connected components of the same types of graphs to find arbitrarily shaped clusters. Current version of RK-CCDs partition only those data sets with arbitrarily shaped clusters where only small amount of noise exist, but such RK-CCDs that are not prone to noise within arbitrarily shaped clusters is still a subject of ongoing research by the authors.

By incorporating both the $\widehat{K}(t)$ function and the silhouette measure, we develop clustering algorithms that are robust to noise, and work well against any data set with unimodal convex clusters.
We demonstrate that CCD based methods successfully locate data sets with both uniformly and normally distributed clusters which may indicate that CCDs are able to detect the clusters of any data set whose clusters are convex and have a single mode. Each covering ball is a representation of the local density; that is, the covering balls which contain a high number of data points are more likely cover the center of a cluster. The greedy algorithm detects centers of clusters while the silhouette measure separates the low intensity regions (which are regions more likely to be containing noise) from the high intensity regions.Moreover, we show that CCDs are also robust to noise. CCD based clustering methods separate the true and noise clusters without assuming the number of noise clusters. We choose a minimum subset of the minimum dominating sets maximizing the silhouette measure, and we assume that the rest are just noise clusters. These collections of noise points do not substantially change the partitioning of the data set, and hence the average silhouette of the data set does not increase.

We conduct three extensive Monte Carlo experiments wherein simulated data sets have convex clusters. We assess the performance of both KS-CCDs and RK-CCDs on these simulated data sets, and also on real data sets. We compare these methods with some prototype-based and density-based clustering methods; such as fuzzy $c$-means and pdfCluster. First, we incorporate fixed centers for the clusters, then, we generate cluster centers from a Strauss Process. On both simulation settings, around each cluster center, we either simulate clusters of multivariate uniformly distributed points (drawn in a box in $\R^d$) or multivariate normally distributed points with a given covariance matrix. The results show that our RK-CCDs perform comparable to KS-CCDs and other methods. RK-CCDs achieve similar rates of correctly estimating the number of clusters in the data sets. However, the performance of RK-CCDs degrade on some cases when there are a large number of clusters, and the data set has high dimensionality. In these cases, average inter-cluster and intra-cluster distances are quite close, and hence, the clustering performance naturally decreases. Moreover, similar to other density-based methods, the performance also decreases with low number of points in each cluster. It is expected of density-based methods that they are inefficient in locating clusters in data sets with low intensity when clusters are considerably close to each other. We also assess the performance of CCD clustering methods on real data sets with arbitrarily shaped clusters, and compare them with DBSCAN. RK-CCDs successfully locate the clusters of data sets with a high number of clusters as well as those with arbitrarily shaped clusters.

RK-CCDs are appealing clustering algorithms with perhaps one drawback, they are computationally expensive. We report on the computational complexity of both CCD methods in Theorems~\ref{thm:complexity} and ~\ref{thm:complexity2}, and show that KS-CCDs are of cubic time and, in particular, RK-CCDs are over cubic time algorithms which makes them slower compared to some other density-based methods. RK-CCD based clustering algorithms are much slower in comparison to KS-CCDs, and it is due to the fact that RK-CCDs are parameter free algorithms whereas KS-CCDs and other methods are not. Although the computation of envelopes before searching for the elements of the MDS, $S(G_{MD})$,  substantially decreases the overall running time (see Theorem~\ref{thm:Kfunc}), the observed $\widehat{K}(t)$ value of the data set should be computed for each point of the data set which makes Algorithm~\ref{alg:ccdripley} still running relatively slower than Algorithm~\ref{alg:ccdks}. In that case, alternative spatial data analysis tools could be incorporated to decrease the computation time. However, our work proves the idea that parameter free CCD algorithms can be devised by estimating the spatial intensity. In our Monte Carlo experiments, the methods assuming the spatial intensity or the number of clusters may perform slightly better than RK-CCDs. However, it may not always be easy or possible to assume, or guess, the values of such parameters.


\section*{Acknowledgments}
Most of the Monte Carlo simulations presented in this article were executed at Ko\c{c} University High Performance Computing Laboratory, and the remaining numerical calculations on real data sets were performed at TUBITAK ULAKBIM, High Performance and Grid Computing Center (TRUBA resources).

\bibliography{ccdred}
\bibliographystyle{apalike}

\end{document}